\documentclass[authoryear,preprint,review,12pt]{elsarticle}

\usepackage[normalem]{ulem}
\usepackage{soul}

\usepackage{amssymb}

\usepackage[table]{xcolor}
\usepackage{setspace}
\usepackage{pdflscape}
\makeatletter
\@namedef{ver@everyshi.sty}{}
\makeatother
\usepackage{amsthm}
\usepackage{makecell}

\definecolor{LightGray}{gray}{0.95}

\usepackage{listings}

\theoremstyle{definition}

\usepackage{url}

\usepackage{makecell}
\usepackage{amsmath}
\usepackage{multicol}
\usepackage{multirow}

\usepackage{longtable}

\journal{Engineering Applications of Artificial Intelligence}

\begin{document}

\begin{frontmatter}



\title{Low-Cost Language Models: Survey and Performance Evaluation on Python Code Generation}

\author[1]{{Jessica} {López Espejel}}

\author[1]{{Mahaman Sanoussi} {Yahaya Alassan}}

\author[1]{Merieme Bouhandi}

\author[1]{Walid Dahhane}

\author[1]{El Hassane Ettifouri}

\affiliation[1]{organization={Novelis
Research and Innovation Lab},
            addressline={40 Av. des Terroirs de France}, 
            city={Paris},
            postcode={75012}, 
            state={},
            country={France}}

\begin{abstract}

\textcolor{black}{Large Language Models (LLMs) have become a popular choice for many Natural Language Processing (NLP) tasks due to their versatility and ability to produce high-quality results. Specifically, they are increasingly used for automatic code generation to help developers tackle repetitive coding tasks. However, LLMs' substantial computational and memory requirements often make them inaccessible to users with limited resources. This paper focuses on very low-cost models which offer a more accessible alternative to resource-intensive LLMs. We notably: (1) propose a thorough semi-manual evaluation of their performance in generating Python code, (2) introduce a Chain-of-Thought (CoT) prompting strategy to improve model reasoning and code quality, and (3) propose a new dataset of 60 programming problems, with varied difficulty levels, designed to extend existing benchmarks like HumanEval and EvalPlus. Our findings show that some low-cost compatible models achieve competitive results compared to larger models like ChatGPT despite using significantly fewer resources. We will make our dataset and prompts publicly available to support further research.}
\end{abstract} 

\begin{graphicalabstract}
\begin{figure}[!ht]
\centering
\makebox[\textwidth][c]{\includegraphics[width=16cm]{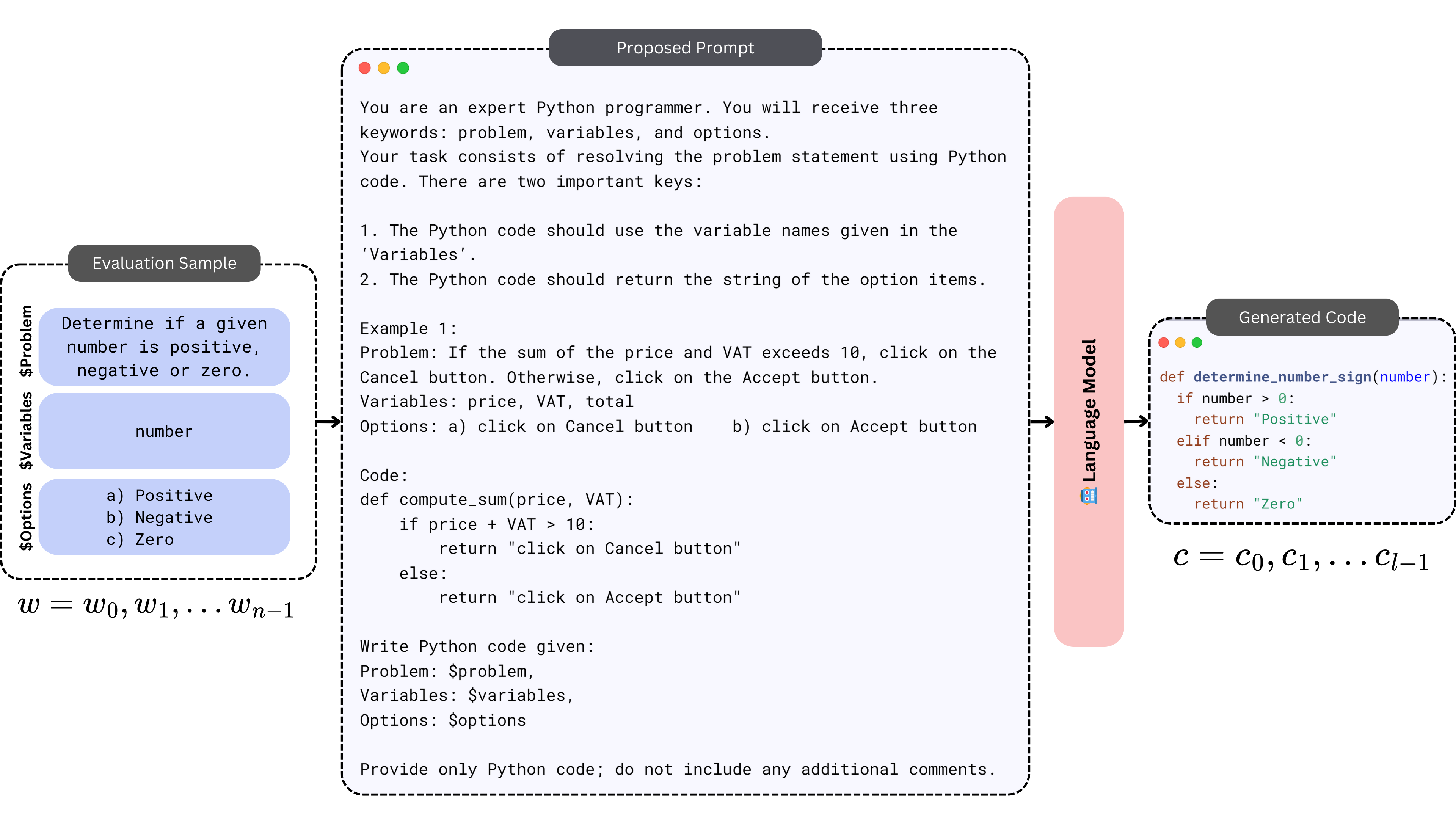}}
\end{figure}
\end{graphicalabstract}

\begin{highlights}
\item   \textcolor{black}{Low-cost} models achieve competitive scores compared to advanced chatbot models like ChatGPT-3.5 \textcolor{black}{\citep{chatGPT3.5}}, ChatGPT-4 \textcolor{black}{\citep{chatGPT4}}, and both Gemini 1.0 and 1.5 Pro \textcolor{black}{\citep{geminiteam2023_gemini}}, which require significant  \textcolor{black}{computational} resources.

\item Some models, such as dolphin-2.6-mistral-7b \textcolor{black}{\citep{ma2023_dolphins}}, excel in generating Python code but lag in meeting the required output formats. Often,  \textcolor{black}{when they generated correct code}, these models  \textcolor{black}{do not return} the wanted options from the input prompt.

\item In contrast, models like  \textcolor{black}{Llama-3.1-8B-Instruct~\citep{dubey2024_llama3herdmodels}} demonstrate consistent performance in understanding prompts and generating code, leading to superior results on our dataset. Similarly, they excel on HumanEval \textcolor{black}{\citep{Chen2021_EvaluatingLL}}, and EvalPlus \textcolor{black}{\citep{liu2023_evalGPTcode}} datasets.

\item Remarkably, llama.cpp \textcolor{black}{\citep{llamaCPP}} project has achieved competitive outcomes using standard computers, a development that seemed very unlikely just a few months ago. This progress highlights the feasibility of running sophisticated language models on \textcolor{black}{Central Processing Units} (CPUs) today.

\end{highlights}

\begin{keyword}
Python Code Generation \sep Natural Language Processing \sep Large Language Models \sep Low-Cost models \sep Chain-of-Thought Prompting


\end{keyword}

\end{frontmatter}


\section{Introduction}
\label{sec: introduction}

    In recent years, automatic code generation has gained importance in the Natural Language Processing (NLP) domain because it automates the programming tasks, especially the most repetitive ones~\citep{ahmad-etal-2021-PLBART}. Particularly, Python code generation has garnered extensive attention in programming language studies  \textcolor{black}{, being ranked among the top three most utilized languages globally} ~\citep{coursera2023_pl, Vailshery2024_pl}. In parallel, Large Language Models (LLMs) have made significant strides in NLP tasks~\textcolor{black}{\citep{chatGPT3.5, chatGPT4, manyika2023_BARD, heidarieverything}}, making them the go-to solution for many application domains including Python code generation. \textcolor{black}{However, despite these advancements, the task of generating programming language code  \textcolor{black}{continues to pose greater complexity compared to} standard natural language generation~\citep{Lopez2023_comprehensive}. This complexity arises from the need for neural networks to accurately interpret natural language instructions to produce correct source code while also adhering to lexical, grammatical, and semantic constraints~\citep{wang-etal-2021-codet5, Scholak2021_PICARD,Liu2023_ChatGPTCode, wong2023natural}. Even a minor mistake, like missing a period or colon, can drastically change the meaning of the code, leading to incorrect model outputs.}

     \textcolor{black}{While existing LLMs deliver exceptional performance, many} are closed-source or \textcolor{black}{face significant limitations due to their reliance on powerful Graphical Processing Units (GPUs) for inference. This reliance presents two main challenges: accessibility and cost,}  particularly impacting individuals and smaller institutions. \textcolor{black}{The reliance on high-performance GPUs, which are expensive and not readily available in standard computing setups, limits who can run these models efficiently. The financial burden of acquiring and maintaining such specialized hardware can be prohibitive. This diverts critical resources from other essential activities, leading to disparities in access to many Artificial Intelligence (AI) technologies. As a result, individuals and smaller organizations face significant barriers to fully participating in cutting-edge research and development, ultimately limiting the broader adoption and innovation of these powerful AI tools.} In addition, even when LLMs are freely accessible, there is a risk of privacy leakage~\citep{Wu2023_security}.

    To address these issues, the research community has  \textcolor{black}{shifted its focus to} models with fewer parameters. For  \textcolor{black}{instance}, while large-scale models like GPT-3 \textcolor{black}{(Generative Pre-trained Transformer)}~\citep{Brown2020_LLMs-ZeroShot_GPT3}, PALM \textcolor{black}{(Pathways Language Model)}~\citep{chowdhery2022_palm}, and ChatGPT-4~\citep{chatGPT4} boast 175B, 540B, and 1.7T parameters respectively, smaller models such as Mistral~\citep{jiang2023_mistral} and LLaMa \textcolor{black}{(Large Language Model Meta AI)}~\citep{touvron2023_llama1, touvron2023_llama2, dubey2024_llama3herdmodels}  demonstrate competitive or superior performance with only 7B parameters.  \textcolor{black}{Furthermore}, there is growing interest in quantization techniques to  \textcolor{black}{speed up} model inference. Popular techniques like GPTQ \textcolor{black}{(Generative Pre-trained Transformers Quantization)} \citep{frantar2023_GPTQ} and AWQ \textcolor{black}{(Activation-aware Weight Quantization)} \citep{lin2023awq} effectively compress LLMs while preserving much of their accuracy on downstream tasks. To leverage these techniques, libraries such as LlamaCPP \citep{llamaCPP} enable the execution of quantized LLMs on CPUs. LlamaCPP focuses on GGUF (GPT-Generated Unified Format) files \citep{Gerganov2023_GGUF}, an advanced binary format optimized for efficient storage and inference.

  \textcolor{black}{Motivated by the objective of demonstrating the potential and reliability of CPU-friendly models in natural language to Python code generation,}  this paper  \textcolor{black}{evaluates} the performance of 
     \textcolor{black}{these open-source models. We then compare their performance to that of closed-source models to assess their capabilities and limitations.}  \textcolor{black}{To do} this, we propose a new dataset for Python code generation, made of $60$ programming problems crafted using GPT-3.5-Turbo API \textcolor{black}{(Application Programming Interface)}.  \textcolor{black}{The input text is structured around} three keywords: $problem$, $variables$, and $options$. The $problem$ describes the desired code function, the $variables$ specify names of variables to use, and the $options$ define return values for conditional statements. There are three levels of difficulty in the dataset (easy, intermediate, challenging). Problems range from basic tasks (e.g., even/odd number check) to complex ones requiring problem understanding, background knowledge (e.g., math), and code structure creation. For \textcolor{black}{the} sake of completeness, we also provide results on two state-of-the-art datasets: HumanEval \citep{Chen2021_EvaluatingLL}, and EvalPlus~\citep{liu2023_evalGPTcode}.

     \textcolor{black}{In addition to that}, we improve the performance of these models by proposing a Chain-of-Thought (CoT)~\citep{Wei2022_COT} prompt that specifically guides the model into the problem solving. The prompt specifies  the model\textcolor{black}{'s}  role, the keywords it needs to focus on, and describes the target task. Then, a single example is provided to validate the process (one-shot inference). Finally, the prompt mentions that the goal is to have only one solving function, without any additional explanations or comments. 
    
    Our main contributions are:

    \begin{itemize}

        \item 
        We propose a Python code generation dataset composed of 60 programming problems with three levels of difficulty. Each programming sample in our dataset is expressed as a $problem$, $variables$, and $options$ (Section \ref{sec:dataset}).

        \item 
        We introduce a carefully\textcolor{black}{-designed} engineered prompt  to enhance the performance of the evaluated models, drawing on insights from state-of-the-art practices (Section \ref{sec:prompt_eng}).

        \item 
        We conducted an extensive semi-manual evaluation of the Python coding abilities of multiple CPU-friendly models from the state of the art. We used for evaluation our proposed dataset, HumanEval \citep{Chen2021_EvaluatingLL}, and EvalPlus~\citep{liu2023_evalGPTcode}. Here, we propose a quality-assessment prompt that is used to guide a GPT-$3.5$-Turbo~\citep{chatGPT3.5} to evaluate the CPU-friendly models, then we manually correct the mistakes done by the GPT model (Sections \ref{sec:results} and \ref{sec:results_human_eval}). 

        \item 
        \textcolor{black}{We will publicly share} our dataset to facilitate future research.
        
    \end{itemize}
 
    The rest of the paper is organized as follows: in Section~\ref{sec:related-work}, we present the related work. Section \ref{sec:pythonCodeGeneration} describes the problem formulation and the models used within the llama.cpp project. Section \ref{sec:dataset} presents the datasets used for the evaluation. Section \ref{sec:prompt_eng} describes our proposed engineered prompts to enhance CPU models performance. Section \ref{sec:evaluation_methodology} describes our evaluation methodology. We discuss the results in Sections \ref{sec:results} and \ref{sec:results_human_eval}, and finally conclude and provide some perspectives in Section \ref{sec:Conclusions}.

\section{Related Work}
\label{sec:related-work}

    \subsection{Python code generation} 

    Python code generation has become a popular and challenging task in recent years.  \citet{Chen2021_EvaluatingLL} \textcolor{black}{led a significant development in this area}  \textcolor{black}{by introducing} the Codex model and the HumanEval dataset. On the one hand, Codex is a fine-tuned GPT model containing up to 12 billion parameters. On the other hand, the HumanEval dataset was proposed as a benchmark for evaluating the proficiency of Python code generation models~\citep{chowdhery2022_palm, li2023_starcoder, roziere2024_codeLLaMA}. Codex has sparked the emergence of other powerful code-generation Language Models (LMs).  The study of these models \textcolor{black}{can be divided} into two groups: closed-source and open-source.

    \paragraph{\textbf{Closed-source models}}  \textcolor{black}{One} prominent example of closed-source models is Alphacode~\citep{Li2022_AlphaCode}, a model built upon the encoder-decoder transformer architecture~\citep{Vaswani2017_transformers}, boasting a staggering $41$ billion parameters. \textcolor{black}{During training,} the authors used the standard cross-entropy loss for next-token prediction in the decoder,  \textcolor{black}{as well as} a Masked Language Modeling (MLM) loss~\citep{devlin2019_bert} for the encoder . In contrast to Alphacode, LaMDA \citep{Thoppilan2022_LaMDA} relies on a decoder-only Transformer architecture~\citep{Vaswani2017_transformers}, comprising $137$ billion parameters  \textcolor{black}{that} was pre-trained to predict the next token in a text corpus.
    
     Gopher \citep{Jack2021_Gopher} and PALM (Pathways Language Model)~\citep{chowdhery2022_palm} share LaMDA's architecture but with $280$ billion and $540$ billion parameters, respectively. Although Gopher used autoregressive Transformer architecture~\citep{Radford2018_LLMMultitask}, the authors  \textcolor{black}{made two significant} changes: \textcolor{black}{they used} RMSNorm~\citep{zhang2019_rms} instead of LayerNorm~\citep{ba2016_layernormalization}, and \textcolor{black}{they opted for} relative positional encodings~\citep{dai2019_transformerxl} instead of absolute ones.  \textcolor{black}{On the other hand}, PALM incorporates modifications such as SwiGLU activations~\citep{shazeer2020_glu} over Rectified Linear Units~\citep{agarap2019_relu}, GeLU (Gated Linear Units)~\citep{Dauphin2017_glu}, or Swish activations~\citep{Ramachandran2017_SwishAS}, and parallel layers~\citep{Wang2021_gptj} within each Transformer block rather than the traditional serialized approach.

    \citet{Hoffmann2022_chinchilla}  found that for compute-optimal training, the model size and the number of training tokens should be scaled equally: for every doubling of model size, the number of training tokens should also be doubled. The study \textcolor{black}{also} revealed that for compute-efficient training, the model size should scale proportionally with the number of training tokens—doubling the model size necessitates doubling the training tokens.  \textcolor{black}{As a result}, Chinchilla used the same compute budget as Gopher but with only $70$B parameters, and four times more data. Chinchilla surpassed its predecessors like Gopher with $280$ billion parameters, GPT-3~\citep{Brown2020_LLMs-ZeroShot_GPT3} with $175$ billion parameters, and Megatron-Turing NLG~\citep{smith2022_megatron} with 530 billion parameters, across various downstream evaluation tasks.

     Chatbot models have exhibited outstanding performance in tasks involving Python code generation. Particularly, ChatGPT-$3.5$~\citep{chatGPT3.5} is built upon the GPT-$3.5$ model \citep{radford2018improving, Radford2018_LLMMultitask, Ouyang2022TrainingLM} and marks a watershed moment in the evolution of AI, excelling in NLP tasks like code generation~\citep{dong2023selfcollaboration, liu2023improving, Liu2023_ChatGPTCode}, and text generation~\citep{Mulia2023_text-generation-chatgpt, lancaster2023artificial}. Afterward, OpenAI introduced ChatGPT-$4$~\citep{chatGPT4}, an upgraded version of the chatbot built upon the GPT-$4$ model.  \textcolor{black}{Although the exact} number of parameters in ChatGPT-$3.5$ remains unknown,  \textcolor{black}{it was recently revealed} that ChatGPT-$4$ contains a total of 1.7 trillion parameters~\citep{Leswing2024_paramsGPT4}.

    Following the release of ChatGPT-$3.5$~\citep{chatGPT3.5} and before ChatGPT-$4$~\citep{chatGPT4}, Google introduced BARD~\citep{manyika2023_BARD} as an enhanced iteration of \textcolor{black}{the} LaMDA \textcolor{black}{model} ~\citep{Thoppilan2022_LaMDA}. BARD enhances predictive abilities through the use of Reinforcement Learning from Human Feedback (RLHF) \citep{Christiano2017_RL}. According to ~\cite{ahmed2023_chatgptBard} and \cite{lopez2023_reasoningGPT_bard}, ChatGPT-$4$ demonstrates \textcolor{black}{superior} reasoning capabilities  \textcolor{black}{compared to} BARD.  \textcolor{black}{More recently}, Google unveiled Gemini~\citep{geminiteam2023_gemini}, an advanced series of models that \textcolor{black}{further} enhances the capabilities of BARD. Built upon Transformer decoders~\citep{Vaswani2017_transformers}, Gemini models support a context length of $32,000$. Notably, Gemini Ultra has  \textcolor{black}{become} the first model to surpass human experts in the MMLU (Massive Multitask Language Understanding) benchmark~\citep{hendrycks2021_MMLU}.  \textcolor{black}{Lastly, the Claude series represents a notable group of performance models. The latest version, Claude $3.5$ Sonner \citep{ClaudeSonner_2024}, surpasses Claude $3$ Opus and other state-of-the-art models in reasoning, coding, question answering, and vision tasks.}

    \paragraph{\textbf{Open-source models}} Concurrently, open-source models emerged  \textcolor{black}{as} an alternative to the private models. These open-source models have also showcased remarkable performance in Python code generation tasks. For instance, InCoder~\citep{fried2023_incoder} and Starcoder~\citep{li2023_starcoder} are models specifically trained on programming code. Diverging from other coding models~\citep{Brown2020_LLMs-ZeroShot_GPT3, Chen2021_EvaluatingLL}, InCoder employs a causal masking objective during its training phase~\citep{aghajanyan2022cm3}. This approach combines the strengths of both causal and masked language models~\citep{devlin2019_bert}, enhancing its learning capabilities. Conversely, StarCoder adopts the FlashAttention mechanism~\citep{dao2022_flashattention} in its training process, which accelerates attention computations and reduces memory usage,  \textcolor{black}{leading to optimized} performance.

    Similar to Incoder, Code Llama-Python~\citep{roziere2024_codeLLaMA} employs the causal masking technique to train infilling models. This model builds upon the LLaMA-2 architecture~\citep{touvron2023_llama2}, incorporating fine-tuning adjustments. \textcolor{black}{The introduced a Long Context Fine-Tuning (LCFT) stage} \citep{roziere2024_codeLLaMA}   \textcolor{black}{allows} models \textcolor{black}{to} handle sequences up to $16,384$ tokens,  \textcolor{black}{which is a significant improvement from} the typical $4,096$ tokens in earlier LLaMA-$2$ stages. Despite having only 7B parameters, Code Llama-Python surpasses the 70B parameter Llama-$2$ in performance on the HumanEval \citep{Chen2021_EvaluatingLL} and MBPP~\citep{austin2021_program} datasets, showcasing its efficiency and effectiveness in handling large-scale coding challenges. \textcolor{black}{Nevertheless, the performance of LLaMA-$2$ has been surpassed by both LLaMA-$3$~\citep{llama3modelcard} and the more recent LLaMA-$3.1$~\citep{dubey2024_llama3herdmodels}.  Notably, LLaMA-$3.1$ is even more powerful than LLaMA-$3$ due to several enhancements: it is a multilingual model, benefits from improved data quality and quantity during both pre-training and post-training, and boasts a model size of $405$B parameters compared to LLaMA-$3$'s maximum of $70$B parameters.}

    In addition to \textcolor{black}{the} efforts to \textcolor{black}{move towards} open source language models, the introduction of the Phi model family~\citep{gunasekar2023_phi1, li2023_phi1.5} highlights a growing interest in creating models that compete in scale as well. The initial model, Phi-1~\citep{gunasekar2023_phi1}, is a $1.3$ billion parameter Transformer-based model designed for basic Python coding tasks,  \textcolor{black}{with an emphasis on} textbook-quality training data. An enhanced version, Phi-1.5~\citep{li2023_phi1.5}, builds upon Phi-1 by incorporating a next-word prediction objective and utilizing a dataset of $30$ billion tokens and training on $150$ billion tokens. This version  \textcolor{black}{is an extension of} Phi-2, a $2.7$ billion parameter model trained with the combined data from Phi-1.5 and a new source of synthetic NLP texts and filtered websites. Therefore, in the next subsection, we will delve exclusively into small models.

    \subsection{Small Language Models}

    Since the  \textcolor{black}{creation} of ChatGPT~\citep{chatGPT3.5}, numerous chatbots have emerged aiming to  \textcolor{black}{handle} similar tasks while prioritizing efficient resource usage and delivering competitive performance. In this subsection, we will \textcolor{black}{only} discuss models with up to 13 billion parameters. One model enabling the development of \textcolor{black}{small yet powerful} models is \textcolor{black}{LLaMA (versions 1 through 3.1)~\citep{touvron2023_llama1, touvron2023_llama2, llama3modelcard, dubey2024_llama3herdmodels}. Across its iterations, the LLaMA family includes variations of 7B, 8B, and 13B parameters, demonstrating comparable or superior performance to} GPT-3 (175B parameters)~\citep{Brown2020_LLMs-ZeroShot_GPT3}  on most  benchmarks. Inspired by LLaMA-1 and the success of instruction-following models like ChatGPT~\footnote{https://chat.openai.com/}, Claude~\footnote{https://claude.ai}, and Microsoft  BingChat~\footnote{https://www.bing.com/chat}, Stanford University launched Alpaca~\citep{Rohan2023_alpaca}. Alpaca is a 7-billion-parameter model fine-tuned from LLaMa-1-7B using 52K instruction-following demonstrations. \textcolor{black}{Despite being significantly smaller,} Alpaca exhibits similar behavior to OpenAI’s text-davinci-003 . It's worth highlighting that Vicuna~\citep{vicuna2023} surged by leveraging the strengths of both LLaMa and Alpaca. However, it slightly exceeds Alpaca in parameter count,  \textcolor{black}{with its} 13 billion parameters.

    \textcolor{black}{Even though the}  models mentioned earlier feature up to 13 billion parameters, a groundbreaking model has emerged: Mistral~\citep{jiang2023_mistral}. Mistral is a 7-billion-parameter LM that enhances inference speed through the use of Grouped-Query Attention (GQA)~\citep{Ainslie2023_GQA}, while simultaneously cost-effectively managing sequences of arbitrary length via Sliding Window Attention (SWA)~\citep{beltagy2020_longformer}. Mistral outperforms the performance of LLaMA-2 13B~\citep{touvron2023_llama2} on all assessed benchmarks, including code generation tasks, as well as the best among released 34B LLaMA-2 variants. Building on  \textcolor{black}{Mistral's success}, the improved Mixtral 8x7B~\citep{jiang2024_mixtral} utilizes a Sparse Mixture of Experts (SMoE) approach. This  \textcolor{black}{improvement} not only enables Mixtral to outperform or match the capabilities of the 70B-parameter Llama-2 and GPT-3.5 model across all benchmarks but also \textcolor{black}{to} maintains the same foundational architecture as the Mistral 7B. The key difference lies in its structure: each layer in Mixtral consists of eight feedforward blocks, or ``experts'', which  \textcolor{black}{enhance} its processing efficiency and model response capability.

    Zephyr~\citep{tunstall2023_zephyr} is an  \textcolor{black}{upgraded} version of Mistral-7B, refined through fine-tuning with the $\sim$ 200,000 samples UltraChat dataset ~\citep{ding2023enhancing, tunstall2023_zephyr}. It has been further aligned using the $\sim$ 64,000 samples UltraFeedback dataset~\citep{cui2023_ultrafeedback} and optimized through the Direct Preference Optimization (DPO) algorithm~\citep{rafailov2023_DPO}. This model outperforms LLaMA2 70B on the MT-Bench benchmark~\citep{zheng2023_MTBench}. However, the MiniCPM model~\citep{minicpm2024} notably surpassed Zephyr-7B-alpha in MT-Bench performance. \textcolor{black}{Just} like Zephyr-7B, it employs supervised fine-tuning and the DPO algorithm, but  is more compact, with only 2 billion parameters.

    Beyond Zephyr, several significant models were refined using Mistral-7B as their foundation, including mistral-7b-openorca~\citep{mukherjee2023_orca}, dolphin-2.6-mistral~\citep{ma2023_dolphins}, and openhermes-2.5-mistral~\citep{Teknium2023_openhermes}. The mistral-7b-openorca model leverages the OpenOrca dataset, focusing on the interaction triad of ``System message, User query, LFM response".  In contrast, dolphin-2.6-mistral enriches its training with three diverse datasets: Capybara~\citep{daniele2023amplify-instruct}, Magicoder-Evol-Instruct-110K~\citep{wei2023magicoder}, and Dolphin~\citep{dolphin_dataset}. Openhermes-2.5-mistral, \textcolor{black}{while} also  \textcolor{black}{building} upon Mistral-7B  \textcolor{black}{set} itself \textcolor{black}{apart} by training on a variety of code benchmarks, including TruthfulQA~\citep{lin2021truthfulqa}, AGIEval~\citep{zhong2023agieval}, and the GPT4All suite~\citep{gpt4all}.

    \subsection{Quantization for model acceleration}

    \textcolor{black}{In light of the success of LLMs in addressing tasks such as question answering~\citep{omar2023chatgpt, zhuang2024toolqa}, and conversational chatbots~\citep{vicuna2023,Rohan2023_alpaca, ramírez2024soloescuchamespanishemotional}, considerable efforts have been directed towards reducing deployment costs. These efforts involve optimizing models for CPU efficiency using compression methods like pruning~\citep{frantar2023sparsegptmassivelanguagemodels,ma2023llm, dery2024everybody} and quantization~\citep{li2023llm,kim2024squeezellmdenseandsparsequantization, huang2024billm}, with the primary aim of decreasing the number of parameters.}

    \textcolor{black}{Regarding quantization, this technique reduces model precision by retaining only the most significant bits, resulting in lower-bit models ($1$-$2$ or $n$ bits). Quantization comes in various forms, including post-training quantization (PTQ)~\citep{xiao2023smoothquant,huang2024billm}, Quantization-Aware Training (QAT)~\citep{chen2024efficientqat,liu2023llm} and weight-only~\citep{frantar2023_GPTQ}. This paper focuses on weight-only quantized models, which address inference time challenges due to memory constraints related to weight parameters~\citep{park2024anyprecisionllmlowcostdeployment}. One notable early work in this area is GPTQ~\citep{frantar2023_GPTQ}. GPTQ (Gradient Post-Training Quantization) uses second-order information for efficient weight quantization in a single step. AWQ (Activation-aware Weight Quantization) \citep{lin2023awq} focuses on activation distributions rather than weight importance and does not require backpropagation or reconstruction. The QuIP (Quantization with Incoherence Processing) technique \citep{chee2023quip} enhances quantization by promoting incoherence in weight and Hessian matrices using random orthogonal matrices. In contrast, SpQR (Sparse-Quantized Representation)~\citep{dettmers2023spqr} addresses quantization errors by storing outlier weights in higher precision while compressing other weights to 3-4 bits. Collectively, these methods offer sophisticated solutions for weight-only quantization, significantly improving model efficiency and accuracy.}

    \textcolor{black}{Building on these advancements in quantization, we utilized the open-source LlamaCPP project \citep{llamaCPP} to conduct experiments on generating Python code from natural language using quantized models. This framework supports quantization for various models, aligning with our focus. Detailed information about the models used in our experiments is provided in Subsection~\ref{subsec:evaluated_models}.}

    \section{Python Code Generation using Low-Cost CPU Models} 
    \label{sec:pythonCodeGeneration}

    \subsection{Problem Definition}
    \label{subsec:problem-definition}

    Given a natural language input text $w = \{w_0, w_1,.., w_{n-1}\}$ of length $|w| = n$, a Python code generation model seeks to generate an equivalent Python source code sequence $c = \{c_0, c_1,.., c_{l-1}\}$ of length $|c| = l$, where $w_i$ and $c_i$ are input and output tokens respectively. For the sake of clarity, we consider ``CPU-friendly'' \textcolor{black}{any} model  \textcolor{black}{that can be} run on CPU  \textcolor{black}{for} inference (not necessarily during training).

    \begin{figure}[!ht]
      \makebox[\textwidth][c]{\includegraphics[width=1\textwidth]{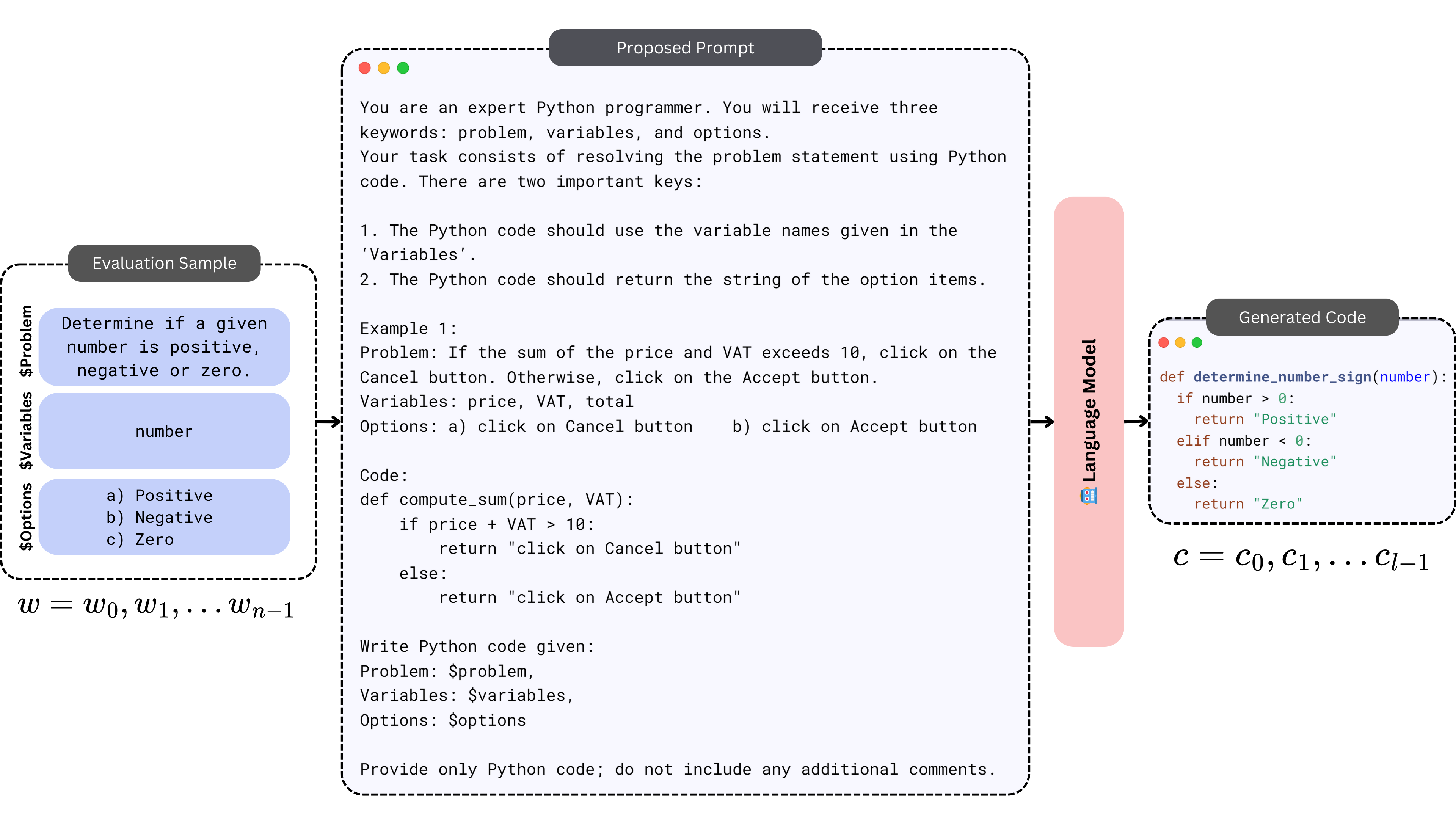}}%
      \caption{Overview of Python code generation with our system. \textcolor{black}{The input consist of a $problem$, $variables$, and $options$. We feed it to the language model along with an engineered prompt to guide the code generation. The language model generates the equivalent Python code.}}
      \label{fig:python_generation}
    \end{figure}

    We provide \textcolor{black}{an overview of our system} in Figure \ref{fig:python_generation} . As mentioned in the introduction, the input text (purple) comprises three key elements: a problem statement, variable names, and a set of options. We feed this input into our Chain-of-Thought prompting module (grey) to generate an efficient prompt that guides the model in its generation. Then, we provide this prompt to the CPU-friendly Language Model (yellow), and the latter finally provides the output Python code (blue). More details about the dataset and the Chain-of-Thought prompting module are provided in Sections \ref{sec:dataset}, and \ref{sec:prompt_eng}, respectively.

   \subsection{Evaluation Protocol within llama.cpp Project}
   \label{subsec:evaluated_models}

    We  \textcolor{black}{performed an evaluation} using GGUF format model files sourced from the llama.cpp \citep{llamaCPP} project, which is a C++ library primarily optimized for CPU usage. This library effectively implements several LLM architectures, including noteworthy models such as LLaMA \citep{touvron2023_llama1, touvron2023_llama2, dubey2024_llama3herdmodels}, Mistral \citep{jiang2023_mistral}, Mixtral \citep{jiang2024_mixtral}, StarCoder \citep{li2023_starcoder}, and Gemma \citep{gemmateam2024_gemma}. These models  \textcolor{black}{are available} in various quantized versions, ranging from $2$ to $8$ bits, and are designed to support both CPU and GPU inference, demonstrating the versatility of llama.cpp in different computing environments. The lightweight design of llama.cpp ensures portability and speed, enabling rapid responses across a wide range of devices. Its customizable low-level features empower developers to deliver effective real-time coding assistance. 

\begin{figure}[!ht]
    \makebox[\textwidth][c]{\includegraphics[width=1\textwidth]{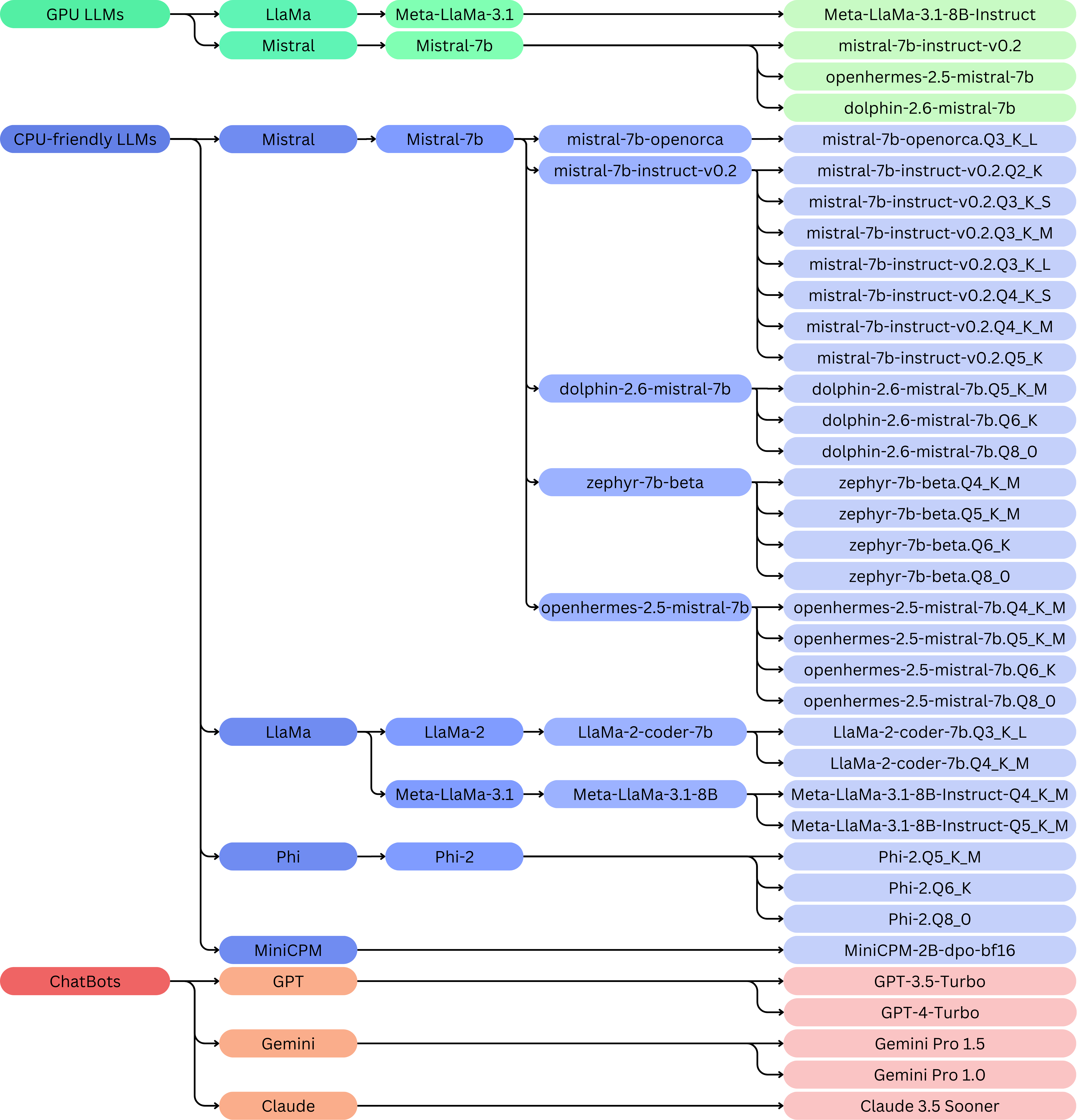}}%
    \caption{Hierarchical organization of all models (CPU, GPU, and Chatbots) examined in the experiments conducted in this paper.}
    \label{fig:all_models}
    \end{figure}

      \textcolor{black}{In the} many model variants \textcolor{black}{we evaluate (the full list of models is shown in Figure ~\ref{fig:all_models}), the name is} noted as such $model$-$name.Q_j\_K\_Size$, where:

     \begin{itemize}
         \item $model$-$name$ is the model name and can be  \textcolor{black}{any} of the following: llama-2-coder-7b, \textcolor{black}{Llama-3.1-8B-Instruct}, phi-2, mistral-7b-instruct-v0.2, mistral-7b-openorca, zephyr-7b-beta, dolphin-2.6-mistral-7b, openhermes-2.5-mistral-7b, and MiniCPM-2B-dpo-bf16.
         \item $Q_j\_K\_Size$ refers to a specific type of quantization method. $j$ is the bit-width used, K refers to the use of K-means clustering in the quantization, $Size$ is the size of the model after quantization,  \textcolor{black}{with} $S$, $M$, $L$  \textcolor{black}{indicating} Small, Medium, and Large, respectively.
     \end{itemize}

    \textcolor{black}{\textbf{Hardware specifications}. 
    To conduct our experiments on the CPU, we used a machine equipped with the following hardware: Processor: 11th Gen Intel(R) Core(TM) i7-11850H @ 2.50GHz, RAM: 32.0 GB, Operating System: 64-bit, x64-based processor.}

    \textcolor{black}{\textbf{Software specifications}.We primarily used the following libraries: Transformers (version $4.43.1$), Python (version $3.10$), BitsandBytes (version $0.43.2$), and Accelerate (version $0.21.0$). Moreover, for the experiments in GPU, we use specifically A100 on Google Colab. Additionally, all GPU-based experiments were conducted using the A100 GPU on Google Colab.}
    
    \newpage
    \section{Datasets}
    \label{sec:dataset}

    We evaluate the models using three Python datasets: our dataset, HumanEval \citep{Chen2021_EvaluatingLL}, and EvalPlus~\citep{liu2023_evalGPTcode}. 

    \subsection{Proposed dataset}

    To assess the functional accuracy of code generated by various models, we designed a dataset  \textcolor{black}{comprising} 60 crafted programming problems, each consisting of a problem statement, variable names, and a set of options. The $problem$ $statement$ describes the conditions and constraints of the user's requirements. The $variables$ keyword, as implied by their name, defines the names of variables to be utilized in the Python code. Lastly, the $options$ represent  \textcolor{black}{the potential outcomes} for each condition outlined in the problem statement. All our dataset samples are generated with the help of GPT-3.5-Turbo API, and  \textcolor{black}{then carefully reviewed by hand} afterward.

    The samples are categorized into three levels of difficulty:
    \textcolor{black}{
    \begin{itemize}
    \setlength\itemsep{-0.5em}
        \item the initial 20 samples represent an easy level ;
        \item the subsequent 20 samples constitute an intermediate level ;
        \item the final 20 samples present a challenging level
    \end{itemize}
    }

\begin{figure}[!ht]
        \centering
        \fbox{\includegraphics[width=\textwidth]{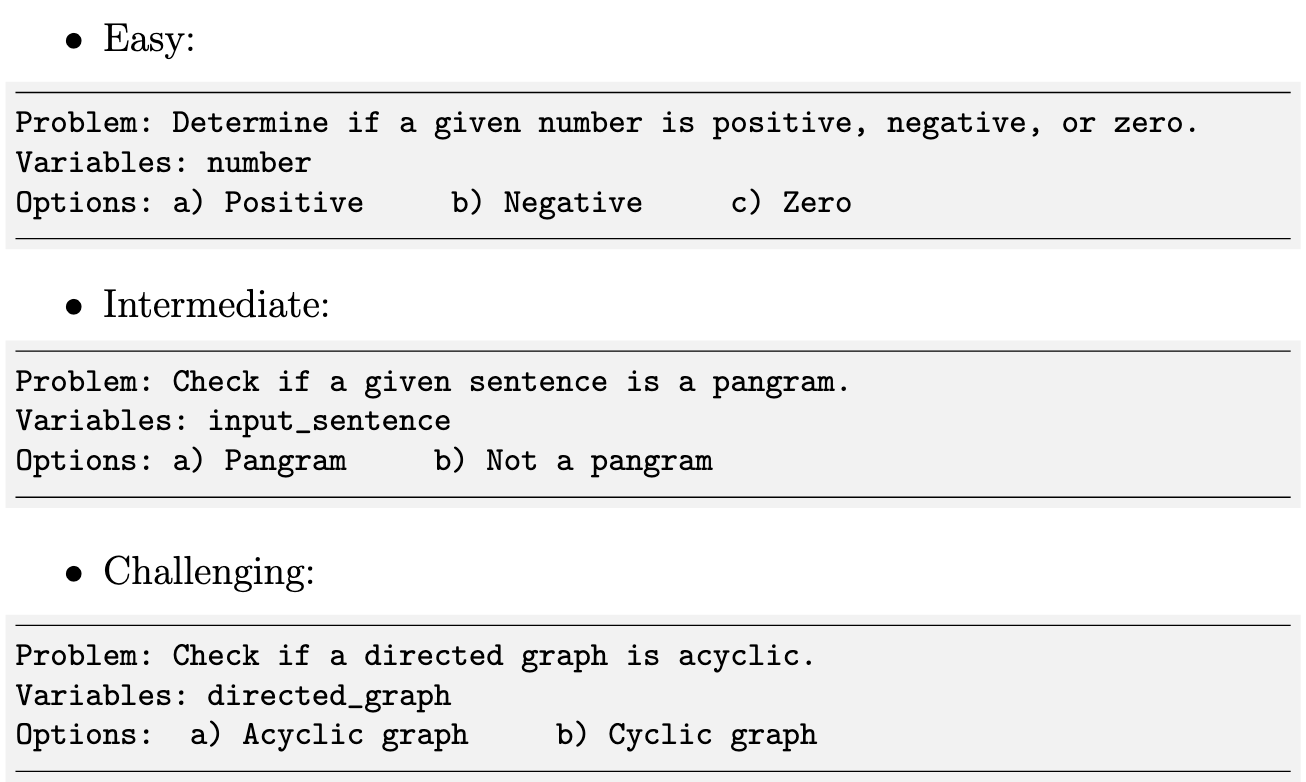}}
        \caption{Examples of different programming languages.}
        \label{fig:exampless}
\end{figure}

\begin{figure}[!ht]
        \centering
        \includegraphics[width=\textwidth]{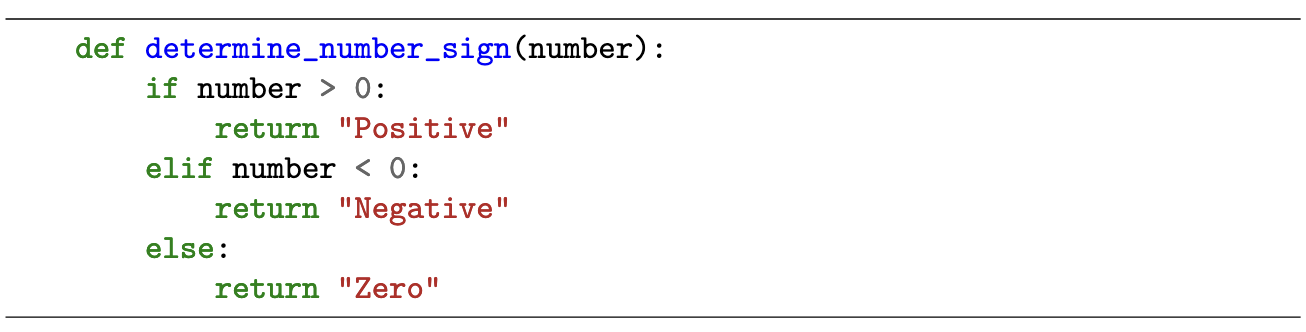}
        \caption{Example solution for the easy problem from Figure \ref{fig:exampless}.}
        \label{fig:sol_first}
\end{figure}

    The samples include mathematical, logic, and reasoning problems. \textcolor{black}{We chose the topics based on popular coding platforms like LeetCode and GeeksforGeeks, ensuring they reflect the types of challenges commonly encountered in real-world coding tasks. At the easy level, tasks involve basic operations such as identifying the type of an angle based on its degree or determining whether one string is an anagram of another. Intermediate problems typically focus on matrices, requiring a solid understanding of concepts like diagonal and orthogonal matrices.} At the most challenging level,  \textcolor{black}{some} problems that involve graphs, trees, and other complex structures. We provide an example of each difficulty level  \textcolor{black}{in Figure \ref{fig:exampless}, and an example output for the easy example in Figure \ref{fig:sol_first}.}

    \subsection{HumanEval and EvalPlus datasets}

    HumanEval~\citep{Chen2021_EvaluatingLL} is a hand-written dataset, developed by OpenAI. It consists of 164 Python problems. Each prompt input contains a function signature and a docstring. Each problem in this dataset is associated with a canonical solution and an average of 7.7 unit tests. The dataset aims to evaluate the models' ability to accurately complete functions based on their signatures and docstrings. The performance of the models is determined by how successfully they pass the given unit tests, ensuring a strict evaluation of the code's functional correctness.

    EvalPlus~\citep{liu2023_evalGPTcode} is an improved version of the original HumanEval dataset. The authors identified that HumanEval did not encompass all potential scenarios for testing the functional correctness of LLM-synthesized code. To address this issue, they significantly expanded the dataset by increasing the number of test cases. EvalPlus comprises an average of 764.1 tests per problem, providing a more robust and comprehensive framework for assessing code correctness.

    \section{Prompt Engineering to Improve Generation}
    \label{sec:prompt_eng}

    To design our prompt  \textcolor{black}{more thoughtfully}, we  \textcolor{black}{have integrated} insights from the research of \citet{yu2023_betterprompts}.  \textcolor{black}{In their work, they recommend} incorporating a Chain-of-Thought (CoT) approach~\citep{Wei2022_COT} to  steer the model in generating  \textcolor{black}{suitable} responses for specific tasks.  They highlight that effective CoT prompts generally include either CoT demonstrations or textual instructions. To ensure the effectiveness of our prompt, we include both an example input with its corresponding output and textual instructions to  \textcolor{black}{guide} the language model. Furthermore, drawing inspiration from the findings of \citet{chen2023_few1shot}, we  \textcolor{black}{opted} to use a single example in our prompt. Their research demonstrates that LLMs can deliver robust performance even with a minimal, 1-shot demonstration,  \textcolor{black}{showcasing} the efficiency of this streamlined approach.  \textcolor{black}{The} full approach to design the prompt \textcolor{black}{is as follows}:

    \begin{itemize}
        \item 
        We designate a specific role for the chatbot: \textit{You are an expert python programmer}.
        \item 
        We identify the keywords the chatbot should pay attention to:  \textit{You will receive three keywords: problem, variables, and options}.
        \item 
        We  \textcolor{black}{explicitly} describe the task that needs to be performed: \textit{Your task consists of resolving the problem statement using python code}.
        \item 
        We used a single example to show the model how to effectively execute the task. \citet{chen2023_few1shot} suggests that \textcolor{black}{providing} just one example is enough for the models, and  \textcolor{black}{also} helps  \textcolor{black}{in keeping} the prompt short \textcolor{black}{and concise}.
        \item 
         \textcolor{black}{At the end of the prompt, we reiterate} the specific format in which we expect the model to return its output. In our case, we specify that we require only a function code to be returned.
    \end{itemize}

    Figure~\ref{fig:python_generation} displays the complete prompt we propose \textcolor{black}{as the input for each evaluated model. Although this is a general prompt, it was adapted for each model using the corresponding tokenizer via the \textit{encode\_chat\_completion} function from Huggingface~\footnote{\url{https://huggingface.co/}} library. This function ensures the correct template with the appropriate special tokens and roles for each model. For instance, Mistral-7B-Instruct-v0.2 utilizes the roles ``user'' and ``assistant'', whereas Meta-Llama-3.1-8B-Instruct employs three roles, ``system'', ``user'' and ``assistant''. Consequently, the prompt was accurately adapted to fit the requirements of each model. For further details, please refer to~\ref{app:prompt_engineering}.}

\section{Evaluation methodology}
\label{sec:evaluation_methodology}

    In this section, we provide a comprehensive breakdown of the metrics and criteria used to evaluate our dataset, HumanEval \citep{Chen2021_EvaluatingLL}, and EvalPlus~\citep{liu2023_evalGPTcode}. 

    \subsection{Proposed dataset}
    \label{subsec:eval_ourDataset}
\begin{figure}[!ht]
        \centering
        \includegraphics[width=\textwidth]{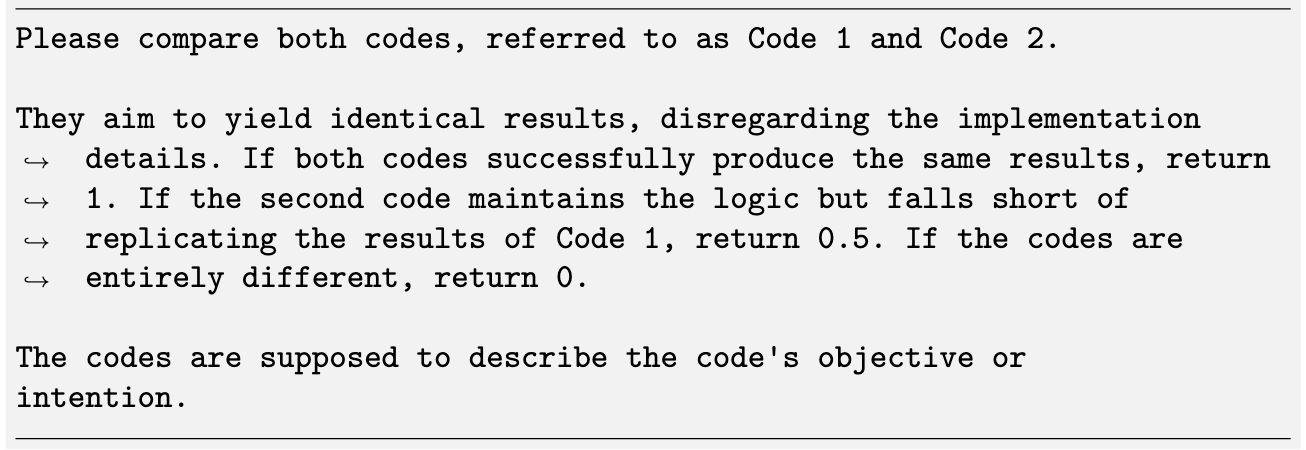}
        \caption{Proposed Prompt to evaluate our results}
        \label{tab:evaluation_prompt}
    \end{figure}

    To assess the performance of CPU-friendly models in generating Python code, we employ semi-manual evaluation.  We provide the evaluation prompt (Figure~\ref{tab:evaluation_prompt}) augmented with the answer generated by the model to GPT-3.5-Turbo~\citep{chatGPT3.5} through its API. The latter returns a correctness score of 0 (wrong answer), 0.5 (passable answer), or 1 (correct answer).  
     \\

    We proposed an approach focuses on thoroughly analyzing code errors without being overly rigid in the assessment. Instead of categorizing outputs as simply correct or incorrect (1 or 0), it accounts for the nuances and subtleties in the generated code. Specifically, it assigns a score of 0.5 to code that, while not yielding the exact results, still adheres to the underlying logic of the problem. For a clearer illustration, please refer to Table~\ref{table:gpt_scores}.

    \textcolor{black}{Table~\ref{table:gpt_scores} presents examples of how GPT-3.5-Turbo scores generated code. In the first example, the code is correct, earning a score of 1. In the second, GPT-3.5-Turbo assigns a $0.5$ because the quantized model failed to convert strings to lowercase, potentially leading to incorrect results. The third example receives a score of $0$ when the code incorrectly attempts to determine if a number is even instead of checking for primality.}

\begin{longtable}{|p{3.7cm}|p{4.1cm}|p{4cm}|p{1cm}|}

\hline
 Problem Statement &  Canonical Solution  & 
Response from the Quantized LLM & GPT Score \\ \hline
\endhead

\small

\vspace{-11em}

\begin{tabular}[t]{@{}l@{}}

\textbf{Problem:} Identify if \\ a given number is a \\ multiple of $3$ \\ 
\textbf{Variables:} n\\ 
\textbf{Options:} a) Multiple \\ of 3 b) Not a multiple \\ c) Other
\end{tabular} 
\vspace{-4em}
&

\begin{minipage}{.28\textwidth}
\includegraphics[width=\textwidth]{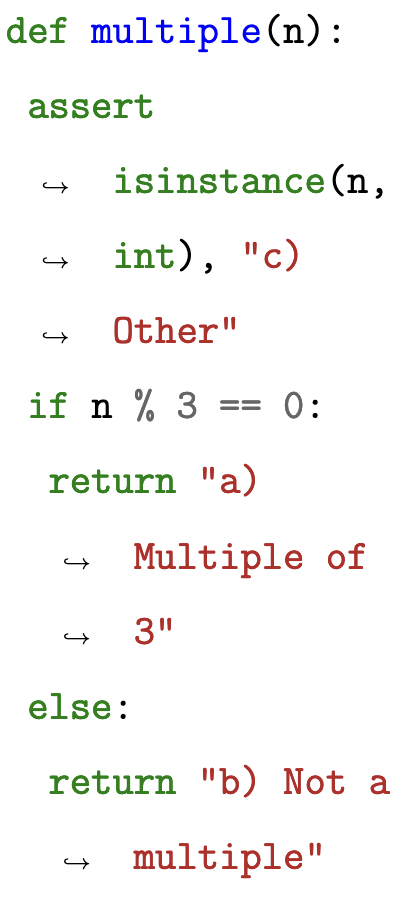}
\end{minipage}
\vspace{-4em}

&

\begin{minipage}{.28\textwidth}
\includegraphics[width=\textwidth]{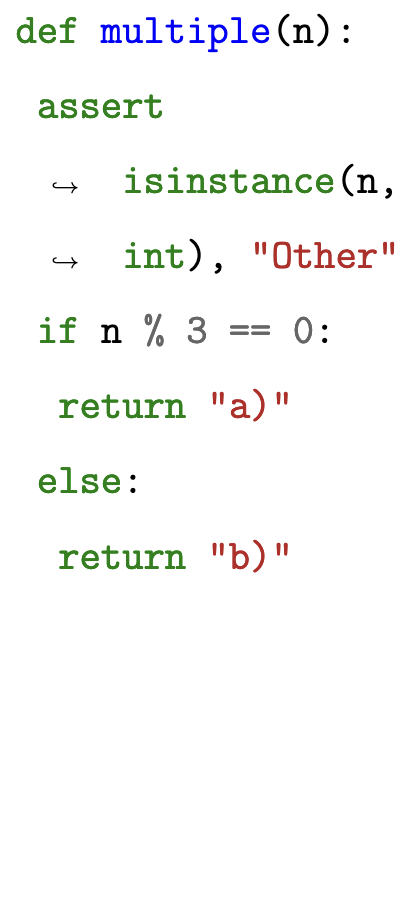}
\end{minipage}
\vspace{-4em}

&
1
\vspace{-4em}
\\ \hline
\small

\vspace{-8em}

\begin{tabular}[t]{@{}l@{}}
\textbf{Problem:} Check if a \\ given string is an \\ anagram of another \\ string\\ 
\textbf{Variables:} s1, s2\\ 
\textbf{Options:} a) Anagram\\ b) Not an anagram\\ c) Other
\end{tabular} 
\vspace{-8em}
&
\vspace{-7em}
\begin{minipage}{.28\textwidth}
\includegraphics[width=\textwidth]{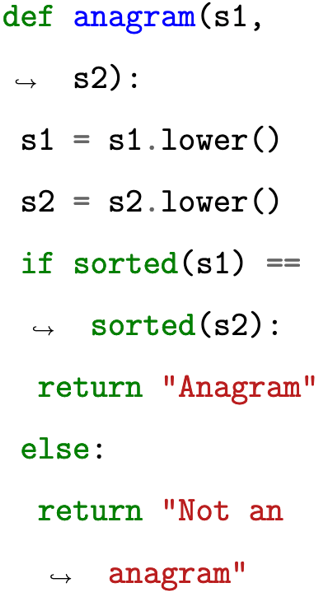}
\end{minipage}
\vspace{-8em}

&

\begin{minipage}{.28\textwidth}
\includegraphics[width=\textwidth]{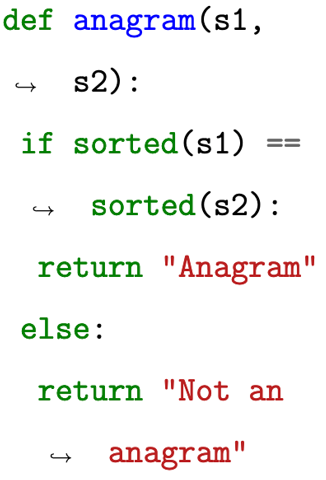}
\end{minipage}
\vspace{-6em}
&

0.5
\vspace{-6em}
\\ \hline

\vspace{1em}
\small
\begin{tabular}[t]{@{}l@{}}

\textbf{Problem:} Determine \\ if a given number is \\ prime. \\ 
\textbf{Variables:} n\\ 
\textbf{Options:} a) Prime \\ number \\ b) Not a prime number \\ c) Other
\end{tabular} 
&
\vspace{1em}
\begin{minipage}{.28\textwidth}
\includegraphics[width=\textwidth]{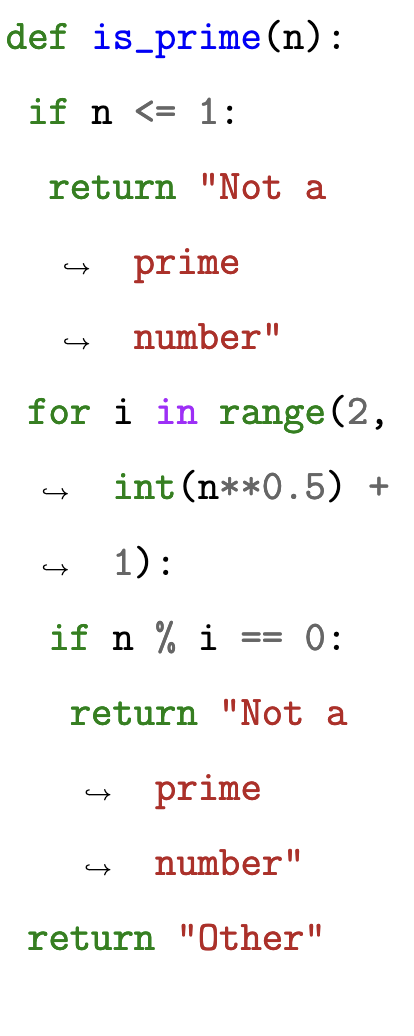}
\end{minipage}
\vspace{-1em}
&
\vspace{1em}
\begin{minipage}{.28\textwidth}
\includegraphics[width=\textwidth]{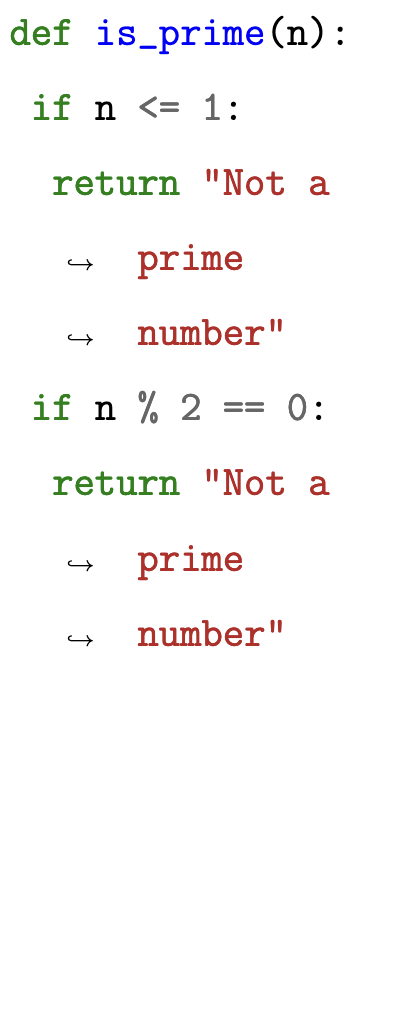}
\end{minipage}
\vspace{-1em}
&
\vspace{1em}
0
\\ \hline

\caption{Examples of scores assigned by GPT-3.5-Turbo using the prompt detailed in Table~\ref{tab:evaluation_prompt}}
\label{table:gpt_scores}
\end{longtable}

    \textcolor{black}{We intervened manually twice. First, we double-checked the scores assigned by GPT-3.5-Turbo to code generated by the quantized models using test cases from our dataset. As described in the dataset section, these tests ensure the code uses the specified variable names, returns the expected options (e.g., "a)", "b)", "c)" or the full text), and passes all performance criteria. For example, Table~\ref{table:penalize_test_cases} shows a case where GPT-3.5-Turbo initially scored the code as $1$. However, after applying our test cases, the score was reduced to $0.5$ due to incorrect variable names (as required in Table~\ref{table:gpt_scores})}.

    \begin{table}[!ht]
        \centering
        \small
        \begin{tabular}{|p{4.5cm}|p{0.5cm}|p{5cm}|p{0.5cm}|} 
        \hline
        \multicolumn{1}{|c|}{\textbf{\begin{tabular}[c]{@{}c@{}}Canonical\\ Solution\end{tabular}}} & 
        \multicolumn{1}{c|}{\textbf{\begin{tabular}[c]{@{}c@{}}GPT\\Score \end{tabular}}} & 
        \multicolumn{1}{c|}{\textbf{\begin{tabular}[c]{@{}c@{}}Response from the\\ Quantized LLM\end{tabular}}} & \multicolumn{1}{c|}{\textbf{\begin{tabular}[c]{@{}c@{}}Test\\ Score\end{tabular}}}
        \\ \hline

\begin{minipage}{.35\textwidth}
\includegraphics[width=\textwidth]{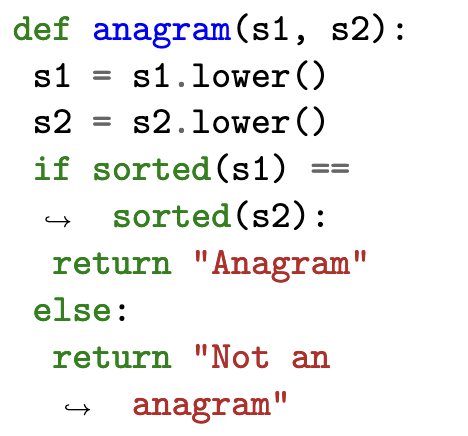}
\end{minipage}

        & 
        1
        & 

\begin{minipage}{.35\textwidth}
\includegraphics[width=\textwidth]{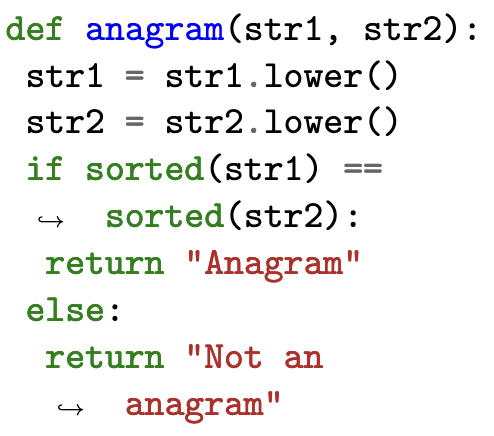}
\end{minipage}

         & 
         0.5
        
       \\ \hline
       
        \end{tabular}
        \caption{An example of penalization in the code: GPT-3.5 initially scored it as 1 (correct code), but after applying our test cases, the score was reduced to 0.5 due to incorrect use of variable names.}
        \label{table:penalize_test_cases}
    \end{table}

    \textcolor{black}{At this stage of our methodology, we ensure that correct answers are accurately identified. However, we manually review the code assigned scores of 0 and 0.5. Our experience shows that the model occasionally assigns a score of 0 when a 0.5 is more appropriate. This occurs because the model lacks the nuanced understanding that some code, despite failing tests, still maintains correct logic and warrants a score of 0.5. Table~\ref{table:correct_score_manually} illustrates an example of this manual adjustment to ensure fair scoring of each generated code.}

   \begin{table}[!ht]
        \centering
        \small
        \begin{tabular}{|p{5cm}|p{5cm}|p{0.5cm}|p{0.5cm}|} 
        \hline
        \multicolumn{1}{|c|}{\textbf{\begin{tabular}[c]{@{}c@{}}Canonical\\ Solution\end{tabular}}} & 
        \multicolumn{1}{c|}{\textbf{\begin{tabular}[c]{@{}c@{}}Response from the\\ Quantized Model\end{tabular}}} & 
        \multicolumn{1}{c|}{\textbf{\begin{tabular}[c]{@{}c@{}}GPT\\Score \end{tabular}}} & 
        \multicolumn{1}{c|}{\textbf{\begin{tabular}[c]{@{}c@{}}Manual\\ Score\end{tabular}}}
        \\ \hline
        

\begin{minipage}{.35\textwidth}
\includegraphics[width=\textwidth]{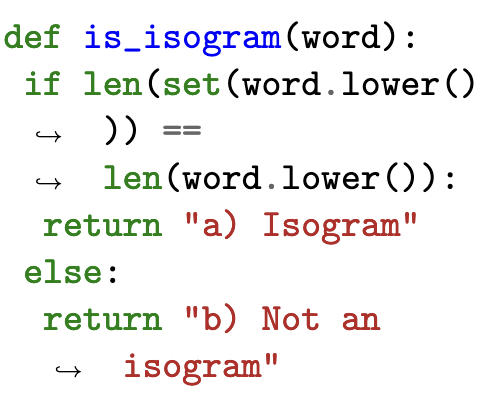}
\end{minipage}

        & 
\begin{minipage}{.35\textwidth}
\includegraphics[width=\textwidth]{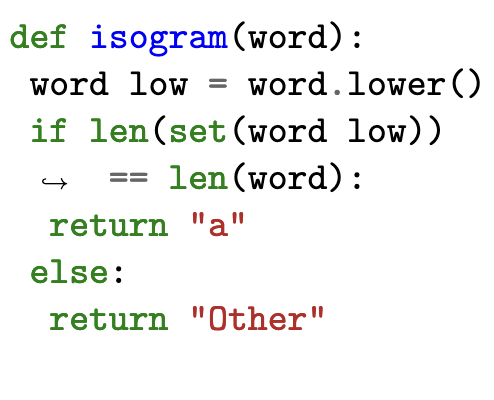}
\end{minipage}
        & 
        0

        & 
        0.5
        
       \\ \hline
       
        \end{tabular}
        \caption{This example demonstrates a case where GPT-3.5-Turbo assigned a score of 0 to code generated by the quantized model. Although the code fails to produce the same results as the canonical solution and returns different responses ("Other" instead of "b" or "b) Not an isogram"), it still maintains the logic of the problem. The code attempts to solve the problem but ultimately fails in its execution.}
        \label{table:correct_score_manually}
    \end{table}
    
    Below, we outline the criteria  \textcolor{black}{for each score in more depth.}

    \begin{itemize}
    \setlength\itemsep{-0.5em}
        \item 
        \textbf{Score = 0 (wrong) -} A score of zero is assigned when the generated code demonstrates no comprehension of the input prompt, indicated by a lack of correlation between the function name, its algorithm, and the requirements specified in the prompt. A score of zero is also given when the function body is empty.
        \item 
        \textbf{Score = 0.5 (passable) -} A score of 0.5 is awarded when the generated code understands the input prompt and attempts to address the problem but fails to deliver the expected response due to syntax issues or logical oversights. For instance, omitting a base case in recursive problems can lead to inefficiency or infinite loops, although the remainder of the algorithm suggests an understanding of the task. Additionally, the code might technically be correct but returns an unexpected output. For example, when asked whether a number is prime with options ``a) Prime number, or b) Not a prime number'', the model might simply return True or False, indicating a lack of deep understanding of the prompt.
        
        \item 
        \textbf{Score = 1 (correct) -} A score of 1 is assigned to the code generation when the model's output matches the gold standard code in functionality, despite possible syntactical differences. It's worth noting that the evaluation criteria are overall not rigid. For instance, if the correct answer is ``a) Prime number'', but the evaluated model omits the ``a)'' and simply returns ``Prime number'' (or vice versa), the generated code is still considered correct.
        
    \end{itemize}

    Our evaluation approach aims to be more flexible. This is achieved by focusing not only on whether the code passes unit tests, but also on understanding the challenges that different models face when generating code across various levels of difficulty.
        
    \subsection{HumanEval and EvalPlus datasets}
    \label{subsec:eval_humaneval}

\begin{figure}[!ht]
        \centering
        \includegraphics[width=\textwidth]{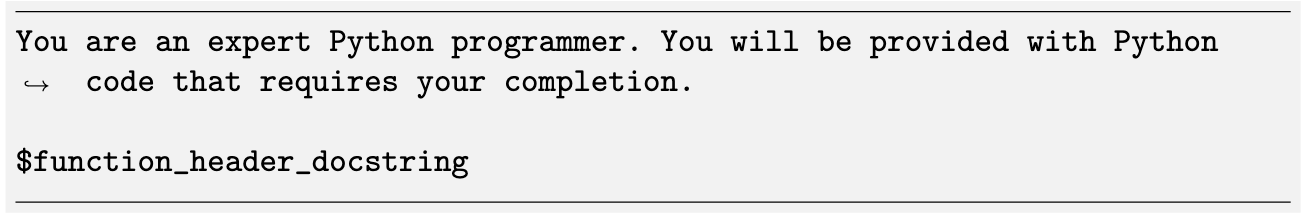}
    \caption{Prompt used to generate Python code for the databases: HumanEval and EvalPlus}
    \label{tab:prompt_humaneval}
    \end{figure}

    \textcolor{black}{To evaluate the performance of the quantized models in the Python code completion task, we used the HumanEval and EvalPlus datasets. In these datasets, the task involves completing a Python function based on the provided function header and corresponding docstring. The prompt used for this task is shown in Figure~\ref{tab:prompt_humaneval}.}



    \textcolor{black}{As detailed in Section~\ref{sec:prompt_eng}, we adapt the template based on the model's specific requirements. For models with both system and user prompts, we use the following approach: the system prompt is ``You are an expert Python programmer. You will be provided with Python code that requires your completion,'' while the user prompt consists of the function header and corresponding docstring.}

    \textcolor{black}{After receiving the LM's response, we perform postprocessing to extract only the generated Python code, as some LMs may include unnecessary text.}  \textcolor{black}{We then evaluate this code using} the pass@k metric \citep{Kulal2019_passK} . Initially, it was introduced to assess functional correctness, where $k$ denotes the number of code samples generated for each problem.  A problem is considered solved if any of these $k$ samples successfully pass the unit tests. The resulting score indicates the proportion of problems solved correctly.

    A slight modification to this metric was proposed by~\citet{Chen2021_EvaluatingLL}. They suggest generating $n$ total samples per task, then counting the correct samples $c \le n$ that pass the unit tests, and computing the unbiased estimator using Equation~\ref{eq:unbiased_estimator}.

    \begin{equation}
        pass@k = \mathbb{E}_{Problems} [1-\frac{\binom{n-c}{k}}{\binom{n}{k}}]
        \label{eq:unbiased_estimator}
    \end{equation}

    However, this estimator often yields large values. Thus, in practice, the score is computed using $1-(1-\hat{p})^k$, where $\hat{p}$ represents the empirical estimate of pass@1.

    \textcolor{black}{Therefore, we evaluated the results using the approach proposed by \citet{Chen2021_EvaluatingLL}. Specifically,} we employ the pass@1 metric, \textcolor{black}{which involves} generating a single sample per problem, \textcolor{black}{using the implementation from the Hugging Face library.}

\section{Results and Discussion: Our dataset}
\label{sec:results}

    \begin{table*}[!ht]
        \centering
        \resizebox{\linewidth}{!}{
        \addtolength{\tabcolsep}{-0.37em}
        \begin{tabular}{|c|l|c|c|c|c|c|c|c|}
        \hline
        & \textbf{Models} & \makecell{\textbf{Size} \\ \textbf{ (GB) $\downarrow$}} &  \makecell{\textbf{Required}\\ \textbf{RAM} \\ \textbf{(GB) $\downarrow$}}  & \makecell{\textbf{Inference}\\ \textbf{Time} \\ \textbf{(ms) $\downarrow$}}
        & \multicolumn{1}{l|}{\makecell{\textbf{Correct} \\ \textbf{(\%) $\uparrow$}}}
        & \multicolumn{1}{l|}{\makecell{\textbf{Passable} \\ \textbf{(\%) $\uparrow$}}}
        & \multicolumn{1}{l|}{\makecell{\textbf{Wrong} \\ \textbf{(\%) $\downarrow$}}}
        & \multicolumn{1}{l|}{\makecell{\textbf{Correct+} \\ \textbf{Passable} \\ \textbf{ (\%) $\uparrow$}}} 
        \\ \hline
        \multirow{4}{*}{\rotatebox[origin=c]{90}{\textcolor{black}{GPU}}} &  
        
        \textcolor{black}{ChatGemini 1.5 Pro} & $\times$ & $\times$ &  $\times$ & \textcolor{black}{\textbf{95.00}} & \textcolor{black}{3.33} & \textcolor{black}{1.66} & \textcolor{black}{96,66} \\ 

        & ChatGPT-3.5 & $\times$ & $\times$ &  $\times$ & 93.33 & 6.67 &  \textbf{0} & \textbf{96.67} \\ 
        
        & ChatGPT-4 & $\times$ & $\times$ &  $\times$ & 86.67 & 11.67 & 1.67 & 92.50 \\ 

        & ChatGemini 1.0 & $\times$ & $\times$ &  $\times$ & 80.00 &  \textbf{18.33} & 1.67 & 89.17  \\ 
        \hline
        \multirow{4}{*}{\rotatebox[origin=c]{90}{\textcolor{black}{GPU}}}  & \textcolor{black}{Meta-Llama-3.1-8B-Instruct} & \textcolor{black}{16.5} & \textcolor{black}{\textbf{29.2}} & \textcolor{black}{\textbf{30.12}} & \textcolor{black}{\textbf{91.66}} & \textcolor{black}{5.00} & \textcolor{black}{\textbf{3.33}} & \textcolor{black}{\textbf{94.16}} \\
        
        & \textcolor{black}{Mistral-7B-Instruct-v0.2} & \textcolor{black}{26.98} & \textcolor{black}{27.7} & \textcolor{black}{23.08} & \textcolor{black}{90.00} & \textcolor{black}{5.00} & \textcolor{black}{5.00} & \textcolor{black}{92.50} \\
        
        & \textcolor{black}{dolphin-2.6-mistral-7b} & \textcolor{black}{26.98} & \textcolor{black}{28.7} & \textcolor{black}{13.16} & \textcolor{black}{60.00} & \textcolor{black}{\textbf{25.00}} & \textcolor{black}{15.00} & \textcolor{black}{72.50}\\
        
        & \textcolor{black}{OpenHermes-2.5-Mistral-7B} & \textcolor{black}{26.98} & \textcolor{black}{28.7} & \textcolor{black}{3.33} & \textcolor{black}{66.66} & \textcolor{black}{18.33} & \textcolor{black}{15.00} & \textcolor{black}{75.82}\\
        
        \hline
        \multirow{27}{*}{\rotatebox[origin=c]{90}{\textcolor{black}{CPU}}}  & \textcolor{black}{Meta-Llama-3.1-8B-Instruct-Q4\_K\_M} & \textcolor{black}{4.92} & \textcolor{black}{7.51} & \textcolor{black}{643.65} & \textcolor{black}{86.66} & \textcolor{black}{\textbf{10.00}}  &\textcolor{black}{3.33}  & \textcolor{black}{91.66} \\
        
        & \textcolor{black}{Meta-Llama-3.1-8B-Instruct-Q5\_K\_M} & \textcolor{black}{\textbf{5.73}} & \textcolor{black}{\textbf{8.14}} & \textcolor{black}{\textbf{662.32}} & \textcolor{black}{\textbf{88.33}} & \textcolor{black}{8.33} & \textcolor{black}{3.33} & \textcolor{black}{\textbf{92.49}} \\

        \cline{2-9}
        
        & llama-2-coder-7b.Q3\_K\_M & 3.30  & 5.80 & 215.51 &  \textbf{16.67} & 16.67 &  \textbf{66.67} &  \textbf{25.00}  \\ 
        
        & llama-2-coder-7b.Q3\_K\_L & \textbf{3.60} & \textbf{6.10} & \textbf{268.36} & 11.67 &  \textbf{20.00} & 68.33 & 21.67 \\ 
        \cline{2-9}
        & phi-2.Q5\_K\_M & \textbf{2.07} &	\textbf{4.57} & 147.48 & 28.33 & 26.67 & 45.00 & 41.67  \\
        
        & phi-2.Q6\_K &2.29 	& 4.79 & \textbf{145.41} & 26.67 & \textbf{41.67} & 31.67 & 47.50  \\  
        
        & phi-2.Q8\_0 & 2.96 &	5.46 & 182.52 & \textbf{30.00} & \textbf{41.67} & \textbf{28.33} & \textbf{50.83}  \\ 
        \cline{2-9}
        & mistral-7b-instruct-v0.2.Q2\_K& \textbf{3.08} & \textbf{5.58} & 209.85 & 33.33 & \textbf{40.00} & 26.67 & 53.33  \\ 
        
        & mistral-7b-instruct-v0.2.Q3\_K\_S & 3.16 &5.66 & 200.09 & 56.67& 20.00 & 23.33 & 66.67 \\ 
        
        & mistral-7b-instruct-v0.2.Q3\_K\_M& 3.52 	& 6.02  & 250.16 & 55.00 & 35.00 & 10.00 & 72.50  \\ 
        
        & mistral-7b-instruct-v0.2.Q3\_K\_L& 3.82 & 6.32 & 261.95 & 50.00 & 21.67 & 28.33 & 60.83  \\ 
        
        & mistral-7b-instruct-v0.2.Q4\_K\_S &  4.14 & 6.64 & 278.04 & 73.33 & 16.67 & 10.00 & 81.67 \\ 
        
        & mistral-7b-instruct-v0.2.Q4\_K\_M& 4.37 	& 6.87 & 304.72 & \textbf{86.67} & 8.33 & \textbf{5.00} & \textbf{90.83} \\ 
        
        & mistral-7b-instruct-v0.2.Q5\_K\_M & 5.13 & 7.63 & 330.64 & 58.33 & 20.00 & 21.67 & 68.33  \\

        \cline{2-9}
        
        & mistral-7b-openorca.Q3\_K\_L & \textbf{3.82} 	& \textbf{6.32} & \textbf{252.96} & \textbf{28.33} & \textbf{40.00} & \textbf{31.67} & \textbf{48.33}   \\ %
        \cline{2-9}
        & zephyr-7b-beta.Q4\_K\_M & \textbf{4.37} & \textbf{6.87} & \textbf{240.36} & \textbf{45.00} & 38.33 & 16.67 & \textbf{64.17}  \\
        
        & zephyr-7b-beta.Q5\_K\_M & 5.13 &	7.63 & 281.22 & 41.67 & 33.33 & 25.00 & 58.33  \\
        
        & zephyr-7b-beta.Q6\_K & 5.94 & 8.44 & 329.11 & 36.67 & 46.67 & 16.67 & 60.00  \\
        
        & zephyr-7b-beta.Q8\_0 & 7.70 & 10.20 & 416.54 & 36.67 & \textbf{53.33} & \textbf{10.00} & 63.33 \\
        \cline{2-9}
        & dolphin-2.6-mistral-7b.Q5\_K\_M & \textbf{5.13} & \textbf{7.63} & \textbf{358.30} & 48.33 & 36.67 & 15.00 & 66.67  \\
        
        & dolphin-2.6-mistral-7b.Q6\_K & $5.94$ & $8.44$ & $393.90$ &$ 30.00$ & \textbf{55.00} & $15.00$ & $57.50$ \\
        
        & dolphin-2.6-mistral-7b.Q8\_0 & 7.70 & 10.20 & 422.94 & \textbf{50.00} & 40.00 & \textbf{10.00} & \textbf{70.00} \\
        \cline{2-9}
        & openhermes-2.5-mistral-7b.Q4\_K\_M & \textbf{4.37} & \textbf{6.87} & \textbf{260.72} & 53.33 & 21.67 & 25.00 & 64.17 \\
        
        & openhermes-2.5-mistral-7b.Q5\_K\_M & 5.13 & 7.63 & 298.93 & 55.00 & \textbf{25.00} & 20.00 & 67.50\\ %
        
        & openhermes-2.5-mistral-7b.Q6\_K & $5.94$ & $8.44$ & $325.59$ & \textbf{58.33} & \textbf{25.00} & \textbf{16.67} & \textbf{70.83} \\
        
        & openhermes-2.5-mistral-7b.Q8\_0 & $7.7$ & $10.20$ & $413.84$ & \textbf{58.33} & $21.67$ & $20.00$ & $69.17$  \\
        \cline{2-9}
        & MiniCPM-2B-dpo-bf16 &\textbf{5.45} & \textbf{10.9} & \textbf{510.25} & \textbf{50.00} & \textbf{36.67} & \textbf{13.33} & \textbf{68.33} \\ 
        \hline
        
        \end{tabular}
       }
        \caption{Evaluation on our dataset. $Q_j$: quantization using $j$ bit width, K: the use of k-means clustering in the quantization, $S$, $M$, $L$: Small, Medium, and Large model size after quantization. The best results of each category are in bold.}
        \label{tab:results_models_ours}
    \end{table*}

    Table~\ref{tab:results_models_ours} presents the \textcolor{black}{results} of our experiments on our dataset. It is  \textcolor{black}{worth noting that}, for comparison, \textcolor{black}{we included the performance of quantized models alongside non-quantized models that require a GPU, as well as the scores for renowned chatbots such as ChatGPT-3.5, ChatGPT-4, and both ChatGemini 1.0 and 1.5-Pro.} Note that, in Table~\ref{tab:results_models_ours}, $Correct+Passable+Wrong=100\%$.

    \paragraph{\textbf{ChatGPT-3.5 vs. ChatGPT-4 vs. Gemini 1.0-1.5 Pro}} Our findings show that for our specific task and input problem formulation, \textcolor{black}{Gemini 1.5 Pro achieved the highest score in the number of correct responses, outperforming ChatGPT-3.5, ChatGPT-4, and Gemini 1.0 by 1.67\%, 8.33\%, and 15\%, respectively. \textcolor{black}{Furthermore,} \textcolor{black}{consistently generated optimized and well-documented algorithms, while also passing unit tests.}}

    \textcolor{black}{It's worth noting that ChatGPT-3.5 outperformed ChatGPT-4 by 6.66\% in terms of correct answers, despite ChatGPT-4's established superiority in tasks like reasoning~\citep{lopez2023_reasoningGPT_bard} and solving mathematical and logic problems~\citep{plevris2023_chats_math}.  In addition, ChatGPT-3.5 produced 0\% wrong answers, compared to ChatGPT-4's 1.67\%. While ChatGPT-4’s algorithms were more optimized and concise, ChatGPT-3.5 tended to be more verbose in its ouputs. The confidential nature of these models' development processes and OpenAI's versioning strategy makes it challenging to explain this behavior. We hypothesize that ChatGPT-3.5's  better understanding of diverse prompts may be due to its more frequent usage, facilitated by free access.}

    \textcolor{black}{Finally, Gemini 1.0 presents the lowest performance in our evaluation, with $80.0$\% correct responses. This model often complicates code generation by adding unnecessary explanations to nearly every line and failing to maintain proper indentation, making the code difficult to copy and paste. Interestingly, both ChatGPT-4 and Gemini 1.0 had the same percentage of wrong answers ($1.67$\%). However, ChatGPT-4 produced fewer passable answers ($11.67$\%) compared to Gemini's $18.33$\%.}

    \begin{figure}[!ht]
        \centering
        \includegraphics[width=\textwidth]{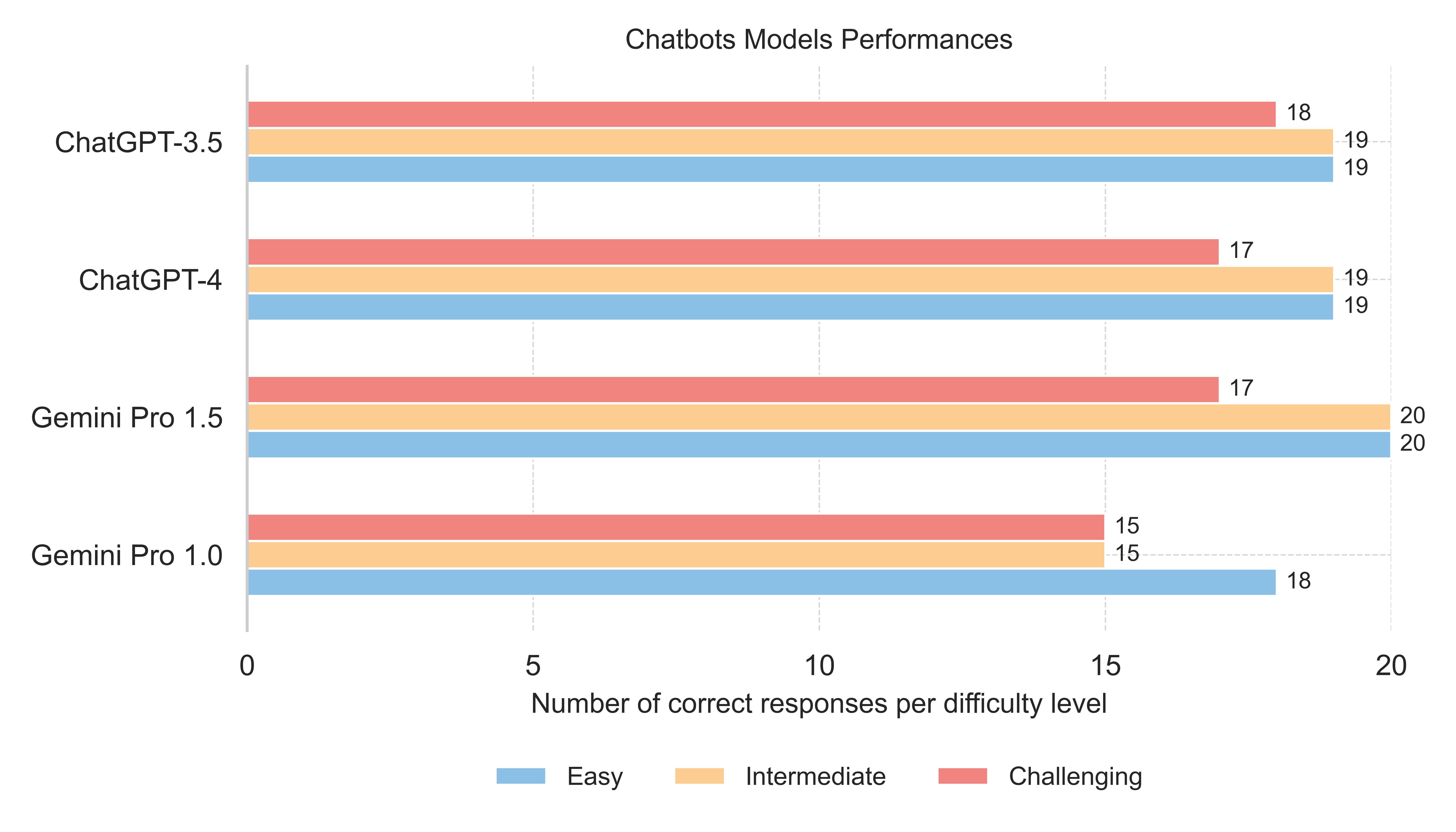}
        \caption{Number of correct answers obtained by Gemini, ChatGPT-4, and ChatGPT-3.5 across easy, intermediate, and challenging levels.}
        \label{fig:performance_chatbots}
    \end{figure}

    Figure~\ref{fig:performance_chatbots} illustrates the number of correct answers obtained by Gemini 1.5 Pro, ChatGPT-3.5, ChatGPT-4, and Gemini 1.0 across easy, intermediate, and challenging levels. \textcolor{black}{On the easy level}, it shows that Gemini 1.5 leads with 20/20, and both  ChatGPT-3.5 and ChatGPT-4 scored $19/20$ , slightly outperforming Gemini 1.0, which scored $18/20$. At the intermediate level, Gemini 1.5 Pro achieved a perfect score of $20/20$. In comparison, ChatGPT-3.5 scored $19/20$, while ChatGPT-4 and Gemini scored $16/20$ and $15/20$, respectively.In the challenging category, ChatGPT-3.5 excelled with a score of $18/20$, followed by ChatGPT-4 and Gemini 1.5 Pro with $17/20$ each, and Gemini 1.0 with $15/20$.  This comparison \textcolor{black}{effectively} highlights the relative strengths and challenges of each model across varying levels of difficulty.

    \paragraph{\textbf{LLaMA models}} In this series of models, we evaluated two versions of LLaMA: LLaMa-2 and LLaMa-3.1. For LLaMA-2, we focused on two quantized models specialized in code: llama-2-coder-7b.Q3\_K\_L and llama-2-coder-7b.Q4\_K\_M. Contrary to the expectations set by \citet{wei2022_emergentAbilities} and \citet{cobbe2021training}, who suggest that larger models perform better, our findings reveal that this is not always true. Specifically, the larger model, llama-2-coder-7b.Q3\_K\_L, performed worse than the medium-sized llama-2-coder-7b.Q4\_K\_M in terms of correct responses, highlighting the significance of bit quantization over model size alone. Overall, the large model scored 21.67\%, while the medium-sized model achieved 25\%, considering both correct and passable answers.

    The primary weakness observed in the llama-2-coder-7 models is a flaw in their background knowledge. For instance, when asked  whether a specific month falls within the winter season, the model generated the following code: if (input\_month  $\ge$ $12$ and input\_month  $\le$ $21$): return "Winter". This response shows a misunderstanding of the problem or a lack of knowledge that a year comprises only $12$ months, making the model incorrectly extend beyond this range. Furthermore, the LLaMA models tend to  \textcolor{black}{generate} verbose code and struggle  \textcolor{black}{offer a variety of} range of options for answers, opting instead to provide direct responses.

    \textcolor{black}{We evaluated the 8B parameter version of Llama-3.1 and for a fair comparison, we tested the non-quantized version, Meta-Llama-3.1-8B-Instruct, which achieved 91.66\% correct responses. In contrast, the 4-bit and 5-bit quantized versions scored 86.66\% and 88.33\% correct responses, respectively. This  clearly demonstrates the impact of quantization on performance. For example, in the quantize version of 4 and 5 bits, Meta-Llama-3.1-8B-Instruct overcome these models with 5\% and 3.33\% in the correct responses, respectively.}

    \textcolor{black}{We also evaluated the 8B parameter version of Llama-3.1. To ensure a fair comparison, we tested the non-quantized version, Meta-Llama-3.1-8B-Instruct, which achieved a 91.66\% accuracy in the correct responses. In contrast, the 4-bit and 5-bit quantized versions scored 86.66\% and 88.33\%, respectively, highlighting the impact of quantization on performance. Notably, the non-quantized Meta-Llama-3.1-8B-Instruct outperformed the quantized versions by 5\% and 3.33\% in accuracy.}

    \begin{figure}[!ht]
        \centering
        \includegraphics[width=\textwidth]{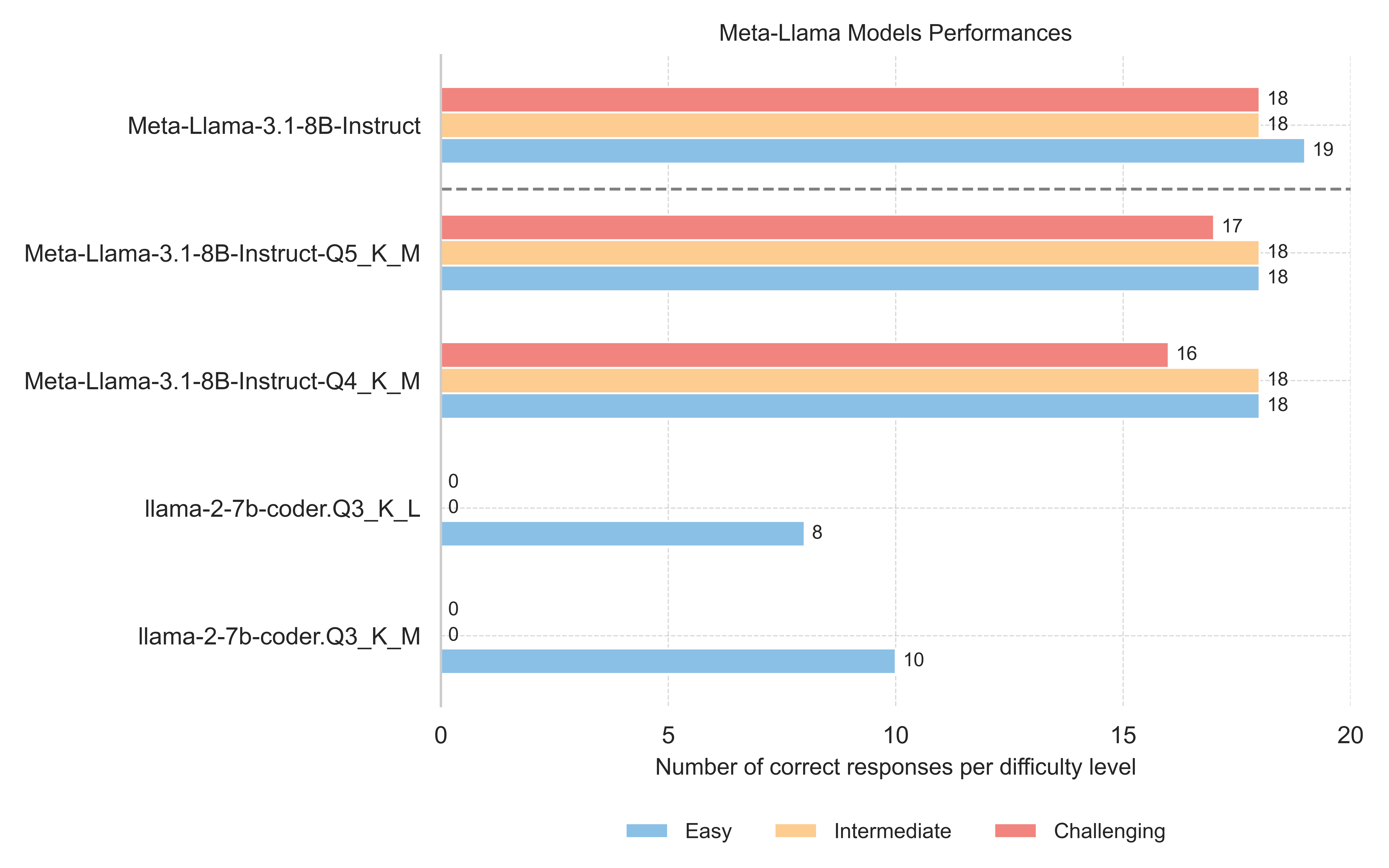}
        \caption{Number of correct answers obtained by LlaMa models across easy, intermediate, and challenging levels.}
        \label{fig:llama_correct_responses}
    \end{figure}

    \textcolor{black}{Among CPU-friendly models for Python code generation, Meta-Llama-3.1-8B-Instruct excelled compared to llama-2-coder-7b, phi-2, mistral-7b-instruct-v0.2, zephyr-7b-beta, dolphin-2.6-mistral-7b, openhermes-2.5-mistral, and MiniCPM-2B-dpo-bf16, across all quantization levels (from 2 to 8 bits). This superior performance can be attributed to the extensive data used during pre-training and post-training, carefully curated through preprocessing. For a detailed breakdown of Llama's performance across easy, medium, and challenging levels, refer to Figure~\ref{fig:llama_correct_responses}.}
    
    \paragraph{\textbf{Phi-2 models}} Our evaluation of Phi-2 models, each with different quantization levels (5, 6, and 8 bits), revealed that Phi-2.Q\_8 achieved the highest accuracy ($30$\%) in correctly generating Python code from prompts, as shown in Table~\ref{tab:results_models_ours}. This surpassed the performance of Phi-2 models quantized to 5 and 6 bits, which obtained $28.33$\% and $26.67$\% correct answers, Interestingly, both Phi-2.Q6\_K and Phi-2.Q\_8 models achieved an identical success rate of $41.67$\% in providing passable responses. Overall, the Phi-2.Q\_8 model  \textcolor{black}{demonstrates} superior performance, with a combined correct and passable accuracy rate of 50.83\%.

\begin{figure}
        \centering
        \includegraphics[width=\textwidth]{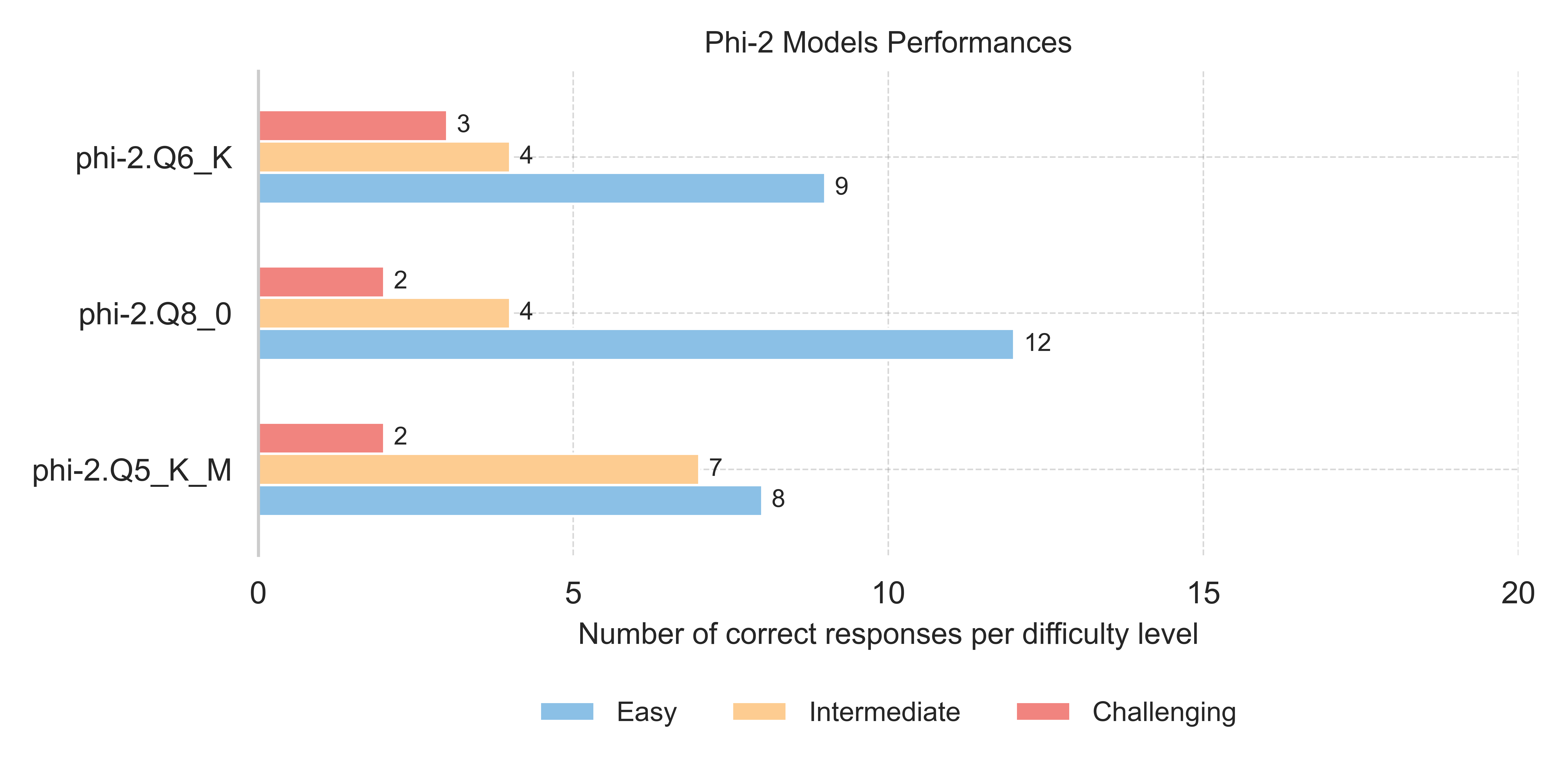}
        \caption{Number of correct answers obtained by Phi-2 models across easy, intermediate, and challenging levels.}
        \label{fig:performance_phi}
    \end{figure}

    \textcolor{black}{However, on our dataset, Phi-2 models generally performed at the lower end compared to other models in terms of correct answers at the easy level, as seen in Figure ~\ref{fig:performance_phi}. The best-performing Phi-2 model, Phi-2.Q8\_0, answered only 12 out of 20 questions correctly. For harder prompts, it yielded less than 20\% accuracy, with only 4 out of 20 and 2 out of 20 correct responses for intermediate and challenging problems, respectively.}

    The primary limitations of the Phi-2 models are their verbosity, including unnecessary comments and often incomplete code, where the models generate the function header and docstring but leave the body empty. Additionally,  \textcolor{black}{there are numerous irrelevant} comments unrelated to the prompt \textcolor{black}{that} indicate the models' tendency to hallucinate. Specifically, the Phi-2.Q5\_K\_M model struggles with proper Python code indentation, making it difficult to read and interpret \textcolor{black}{it}.

    \paragraph{\textbf{Mistral models}} In our study, we  \textcolor{black}{assessed} \textcolor{black}{the performance of the mistral-7b-instruct-v0.2 model and its quantized variants.} We explored quantization levels from $2$ to $5$ bits across three model sizes: small, medium, and large. Additionally, we evaluate \textcolor{black}{a large $3$-bit quantized variant of the} Openorca .

    According to Table~\ref{tab:results_models_ours}, \textcolor{black}{the non-quantized mistral-7b-instruct-v0.2 model unsurprisingly achieved the highest performance among the mistral-7b-instruct family, with 90\% correct answers. Among the quantized versions,} mistral-7b-instruct-v0.2.Q4\_K\_M achieved the best accuracy (86.67\%), followed by the mistral-7b-instruct-v0.2.Q4\_K\_S model ($73.33$\%).  \textcolor{black}{An interesting observation is that} both top-performing models were quantized to $4$ bits. This finding contradicts the hypothesis that models quantized with more bits would necessarily provide higher accuracy. In fact, the mistral-7b-instruct-v0.2.Q5\_K\_M model, quantized with $5$ bits, achieved a $28.34$\% lower accuracy than the mistral-7b-instruct-v0.2.Q4\_K\_M model. Furthermore, the accuracy of the remaining evaluated models ($2$ and $3$ bits) consistently fell below $55.00$\%.

\begin{figure}[t]
        \centering
        \includegraphics[width=\textwidth]{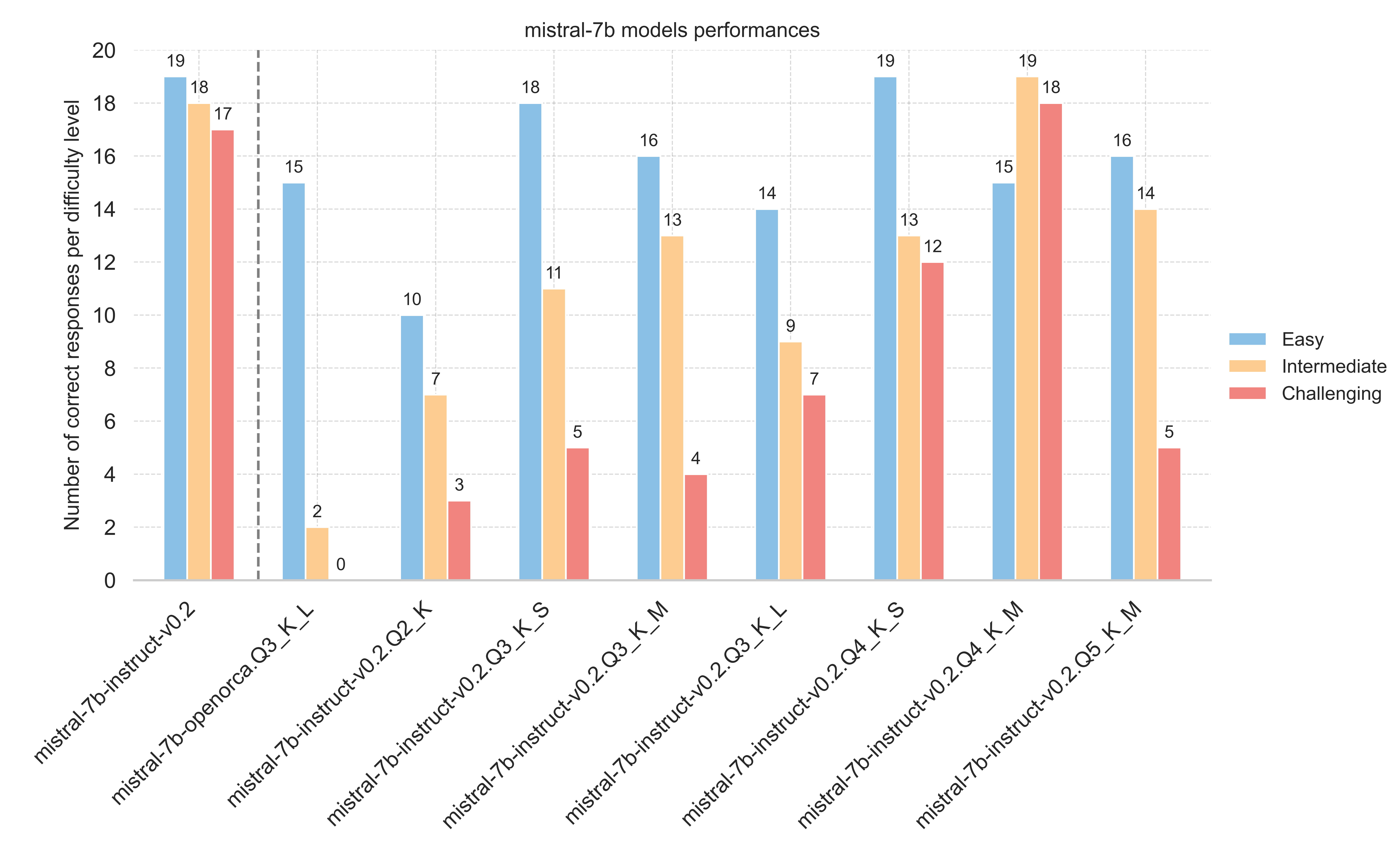}
        \caption{Number of correct answers obtained by mistral models across easy, intermediate, and challenging levels.}
        \label{fig:performance_mistral}
    \end{figure}

    The last column of Table~\ref{tab:results_models_ours} provides the total performance of the model considering both the correct and the passable answers. \textcolor{black}{The non-quantized mistral-7b-instruct-v0.2 achieved the highest score of 92.5\%, followed by mistral-7b-instruct-v0.2.Q4\_K\_M with $90.83$\% and  mistral-7b-instruct-v0.2.Q4\_K\_S with 81.67\%}. Surprisingly, the fourth-best performer was not the model quantized with 5 bits,  \textcolor{black}{instead, it was the} mistral-7b-instruct-v0.2.Q3\_K\_M model  \textcolor{black}{that achived} $72.50$\% of accuracy. This result confirms that the performance of the models in  Python generation  \textcolor{black}{is not solely determined by} the number of bits used to quantize the model.

    Further analysis of Figure~\ref{fig:performance_mistral} revealed interesting trends in correct responses by difficulty level. While mistral-7b-instruct-v0.2.Q4\_K\_S  \textcolor{black}{performed exceptionally well with} easy samples, mistral-7b-instruct-v0.2.Q4\_K\_M  \textcolor{black}{surpassed it} in generating correct responses for intermediate and challenging samples. This suggests that mistral-7b-instruct-v0.2.Q4\_K\_M may be better suited for handling complex Python code generation.

    After finding  \textcolor{black}{that} mistral-7b-instruct-v0.2.Q4\_K\_M \textcolor{black}{has comparable or superior performance} \textcolor{black}{with powerful close-models such as Gemini 1.0 and ChatGPT-4}, we explored the performance of other mistral-based models. Notably, we  \textcolor{black}{looked into} the capabilities of Mistral-7b-openorca quantized to 3-bits and found that it underperformed compared to the original Mistral model. It's important to note that this model was fine-tuned on the OpenOrca dataset, which is not specifically designed for coding tasks. Nonetheless, it achieved $28.33$\% correct responses and $40$\% passable answers.

    In addition, we delved deeper into other models that were fine-tuned on top of Mistral, such as zephyr-7b-beta, dolphin-2.6-mistral, and openhermes-2.5-mistral. We discuss these models in the following sections, and compare their performance at different quantization levels of $4$ bits and higher, allowing us to assess how such adjustments impact their efficiency and output quality. \textcolor{black}{For models with high accuracy in their quantized versions, such as dolphin-2.6-mistral and openhermes-2.5-mistral-7b, we also provided the scores of their non-quantized counterparts.}

\begin{figure}[!ht]
        \centering
        \includegraphics[width=\textwidth]{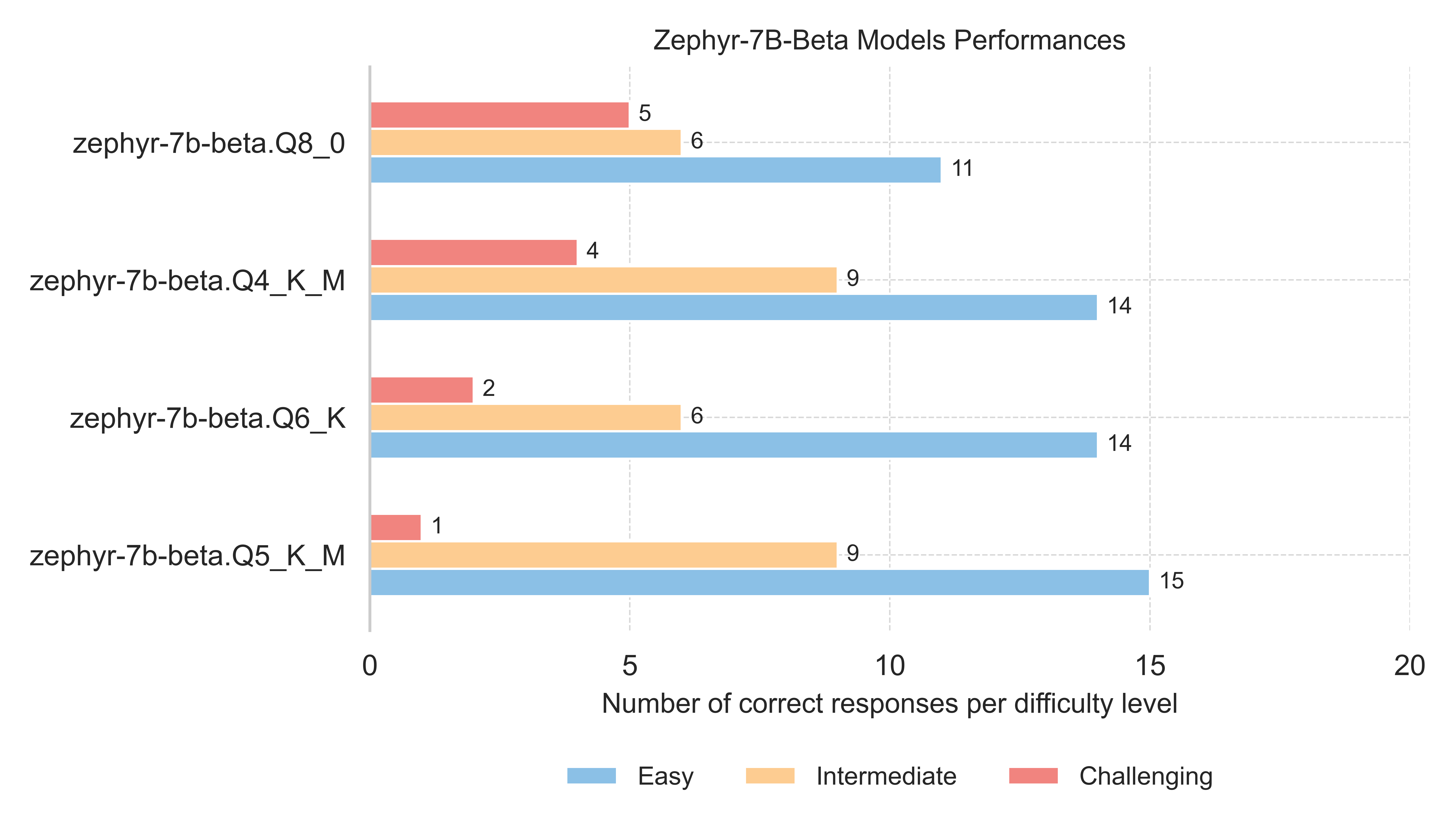}
        \caption{Number of correct answers obtained by zephyr-7b-beta models across easy, intermediate, and challenging levels.}
        \label{fig:performance_zephyr}
    \end{figure}

    \paragraph{\textbf{zephyr-7b-beta models}} Our tests on Zephyr models ($4$, $5$, $6$, and $8$ bits), as seen in Figure ~\ref{fig:performance_zephyr}, showed lower performance compared to mistral-7b-instruct-v0.2.Q4\_K\_M and some other quantized mistral variants. While zephyr-7b-beta.Q4\_K\_M achieved the highest score ($45$\%) within the Zephyr family, it lagged behind mistral-7b-instruct-v0.2.Q4\_K\_M and ChatGPT-3.5 by $41.67$\% and $48.33$\%, respectively.

    Zephyr-7b-beta.Q5\_K\_M achieved the second-best performance of $41.67$\% accuracy. Interestingly, both 6-bit and $8$-bit Zephyr models obtained the same score (36.67\%). However, considering overall performance within the Zephyr family, zephyr-7b-beta.Q4\_K\_M is the best performer, followed by zephyr-7b-beta.Q8\_0. This is because the $8$-bit Zephyr model achieved the highest percentage of passable responses, demonstrating an understanding of the prompt and code logic, albeit with syntactic inaccuracies. It is noteworthy to mention that zephyr-7b-beta.Q8\_0 showcases distinct advantages, such as generating optimized and compact code by leveraging existing libraries like NumPy and utilizing list comprehensions. Contrary to its 5- and 6-bit-width counterparts, zephyr-7b-beta.Q8\_0 seldom produces empty functions.

    Zephyr is fine-tuned on top of Mistral, but worsens its results in Python Code generation.  Designed to align with user intent and handle natural prompts effectively,  it was fine-tuned on the UltraChat 200k dataset~\citep{ding2023enhancing, tunstall2023_zephyr}. UltraChat focuses on three broad areas: understanding real-world concepts, generating various text formats, and working with existing text (rewriting, translation, etc.). Since code generation isn't a core focus in UltraChat, the dataset likely contains minimal code content, which could explain Zephyr's weaker performance.

    \paragraph{\textbf{dolphin-2.6-mistral}} We conducted tests on three quantized versions of the dolphin-2.6-mistral model, using $5$, $6$, and $8$ bits. , \textcolor{black}{T}he dolphin-2.6-mistral-7b.Q8\_0 model, with the highest bit width, achieved the best performance, with 50\% correct responses, closely followed by dolphin-2.6-mistral-7b.Q5\_K\_M at 48.33\%. \textcolor{black}{While the non-quantized version, dolphin-2.6-mistral-7b, scored 60\%, the performance gap with the quantized models is modest.} However, a significant drop is observed with the dolphin-2.6-mistral-7b.Q8\_0 model, which only achieved 30\% accuracy.

\begin{figure}
        \centering
        \includegraphics[width=\textwidth]{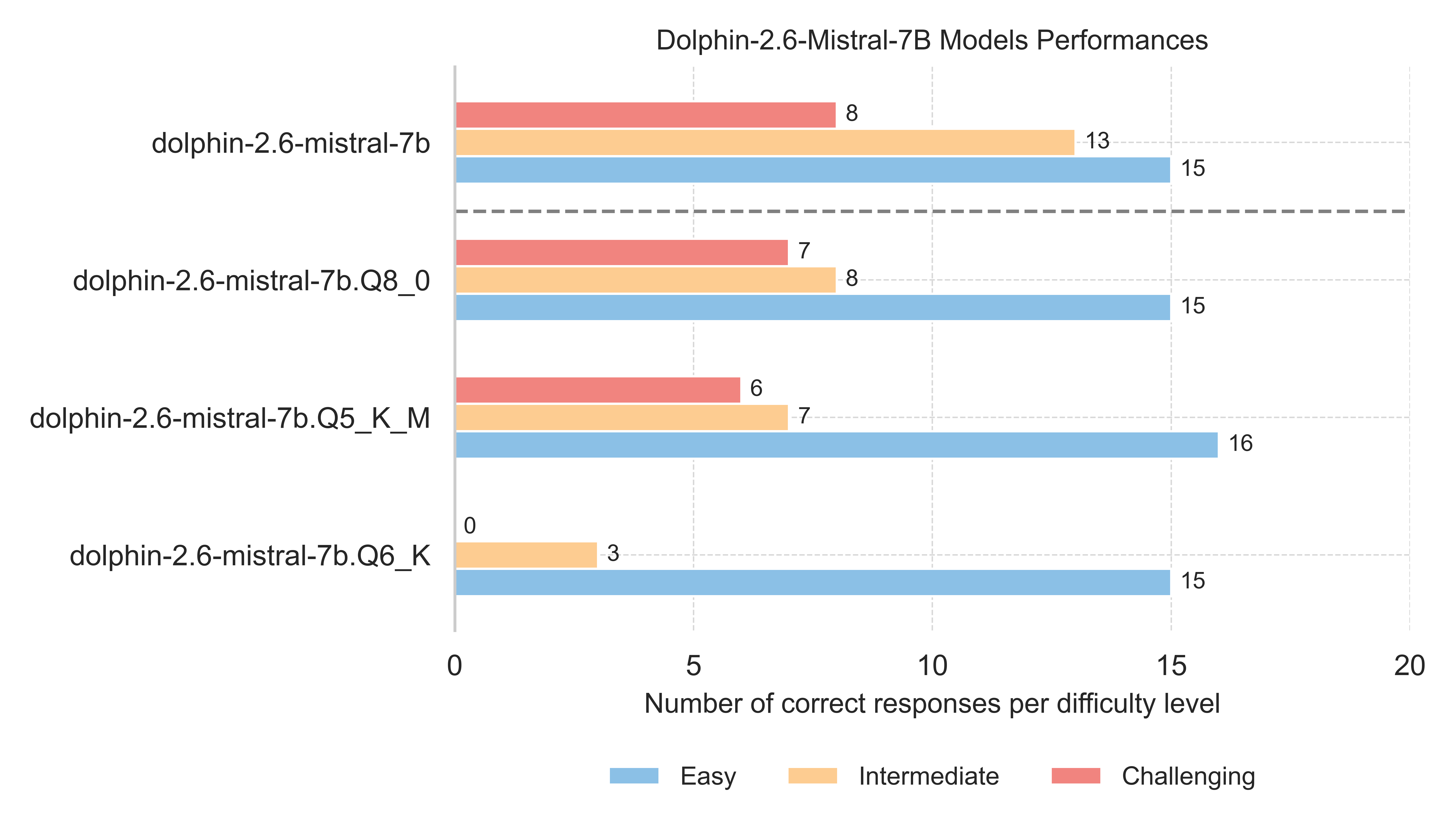}
        \caption{Number of correct answers obtained by dolphin-2.6-mistral models across easy, intermediate, and challenging levels.}
        \label{fig:performance_dolphin}
    \end{figure}

    When considering only correct responses, Figure~\ref{fig:performance_dolphin} highlights the superior performance of the dolphin-2.6-mistral-7b model, followed by the dolphin-2.6-mistral-7b.Q8\_0 variant, across difficult and intermediate levels. Notably, dolphin-2.6-mistral-7b.Q5\_K\_M achieved the highest number of correct answers at the easy level ($16/20$), while the other two models closely followed with $15/20$ each. Both the 8-bit and 5-bit dolphin-2.6-mistral models maintained consistent performance in the intermediate (8/20 and 7/20, respectively) and difficult levels ($7/20$ and $6/20$, respectively). In contrast, the dolphin-2.6-mistral-7b.Q6\_K\_M model experienced a sharp decline, failing to generate any correct answers at the difficult level ($0/20$).

    The enhanced performance of the Dolphin-2.6-Mistral models over Zephyr models can be explained by the datasets used for fine-tuning. Zephyr models were fine-tuned using the UltraChat 200k dataset~\citep{ding2023enhancing, tunstall2023_zephyr}, which has a limited portion of source code. In contrast, Dolphin-2.6-Mistral models benefited from a fine-tuning with Magicoder-Evol-Instruct-110K~\citep{wei2023magicoder} and Magicoder-OSS-Instruct-75K datasets \citep{Magicoder-OSS_75k}, among others. This selection of datasets is a key factor in Dolphin-2.6-Mistral being better than Zephyr models.

    \paragraph{\textbf{openhermes-2.5-mistral}} We evaluated four quantized versions of the openhermes-2.5-mistral model, utilizing $4$, $5$, $6$, and $8$ bits. The analysis revealed that the 4-bit version and the 5-bit version yielded accuracy rates of $53.33$\% and $55$\% in the correct responses, respectively. In contrast, both the $6$-bit and $8$-bit versions demonstrated superior accuracy, achieving a rate of $58.33$\%. \textcolor{black}{While these scores are close to the 66.66\% accuracy of the non-quantized OpenHermes-2.5-Mistral-7B, they highlight the effectiveness of CPU-friendly models for generating Python code from natural language.}

    \textcolor{black}{Moreover,} as shown in Table~\ref{tab:results_models_ours}, the $6$-bit model (openhermes-2.5-mistral-7b.Q6\_K) outperforms the $8$-bit model (openhermes-2.5-mistral-7b.Q8\_0) in terms of overall quality. This is because the $6$-bit model produced more passable responses and fewer incorrect answers. While the $8$-bit model achieved a combined accuracy of $69.17$\% (correct + passable), the 6-bit model reached a higher accuracy of $70.83$\%.

    \begin{figure}[!ht]
        \centering
        \includegraphics[width=\textwidth]{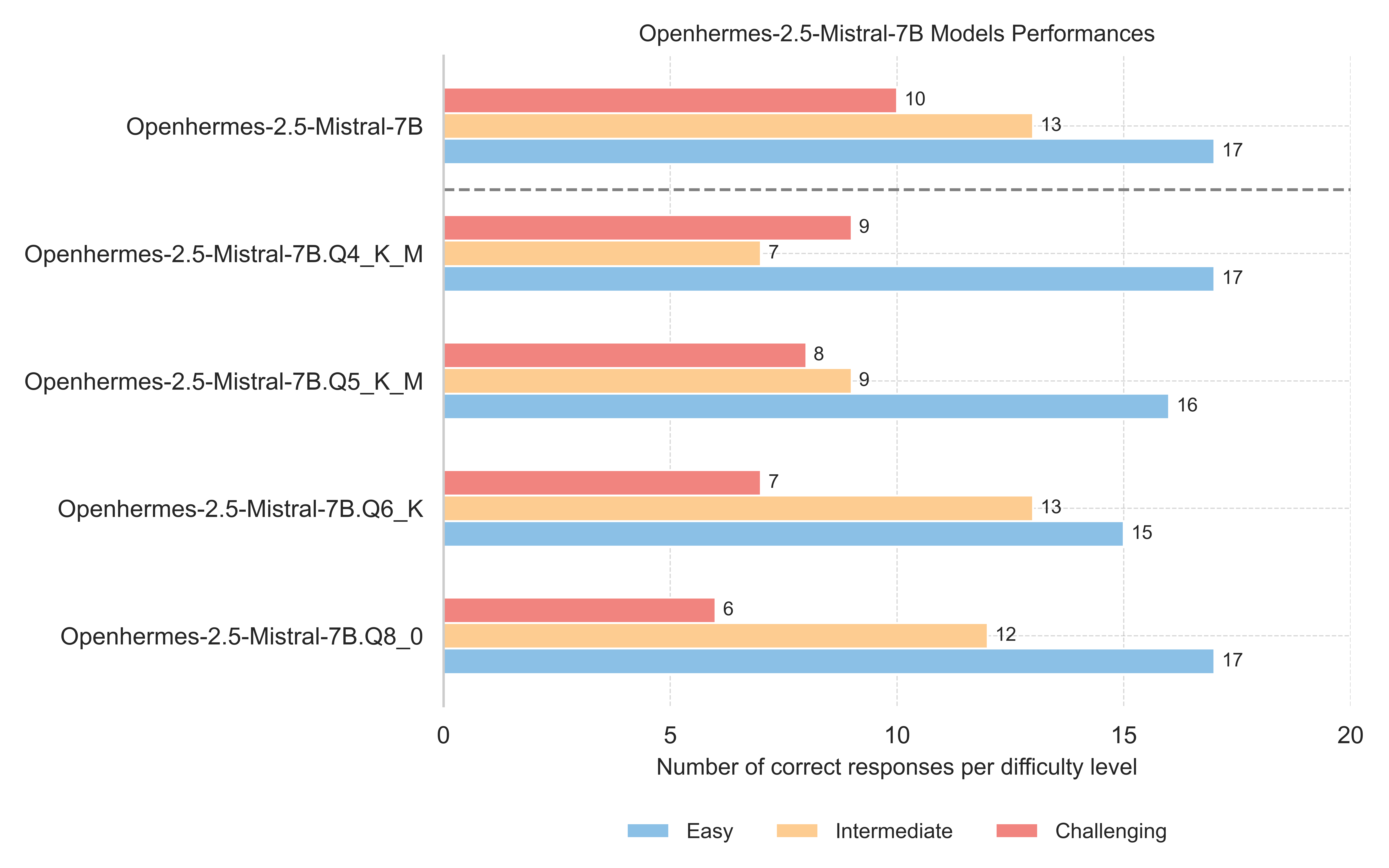}
        \caption{Number of correct answers obtained by openhermes-2.5-mistral models across easy, intermediate, and challenging levels.}
        \label{fig:performance_openhermes}
    \end{figure}

    Examining correct answers across different difficulty levels (Figure~\ref{fig:performance_openhermes}), the findings indicate that both the openhermes-2.5-mistral-7b.Q4\_K\_M and openhermes-2.5-mistral-7b.Q8\_0 models excelled in the easy category, each securing $17$ correct responses. They were closely followed by the 5-bit model with $16$ correct responses, while the 6-bit version recorded the lowest success rate in this category. Furthermore, the openhermes-2.5-mistral-7b.Q4\_K\_M and openhermes-2.5-mistral-7b.Q5\_K\_M models secured the highest success rate in the challenging category, each achieving 8 out of $20$ correct responses. While not the best in either easy or challenging categories, the 6-bit openhermes-2.5-mistral-7b.Q6\_K\_M model achieves consistent performance across all difficulty levels. This stability makes it the overall best choice, considering both correct and passable responses.

    The main weaknesses of the openhermes-2.5-mistral family are hallucinations and lack of code conciseness. For instance, the model references nonexistent NumPy functions or incorrectly attributes SciPy functions to the NumPy library. Moreover, the generated code lacks compactness; for example, rather than employing the more concise condition: $18.5$ $\le$ bmi $\le$ $24.9$, it opts for the lengthier if bmi $\ge$ $18.5$ and bmi $\le$ $24.9$.  Additionally, it's noteworthy that the openhermes-2.5-mistral $4$-bit version tends to leverage native structures more effectively for algorithm development, in contrast to the $6$ and $8$-bit versions, which more frequently utilize library functions.

\begin{figure}[!ht]
        \centering
        \includegraphics[width=\textwidth]{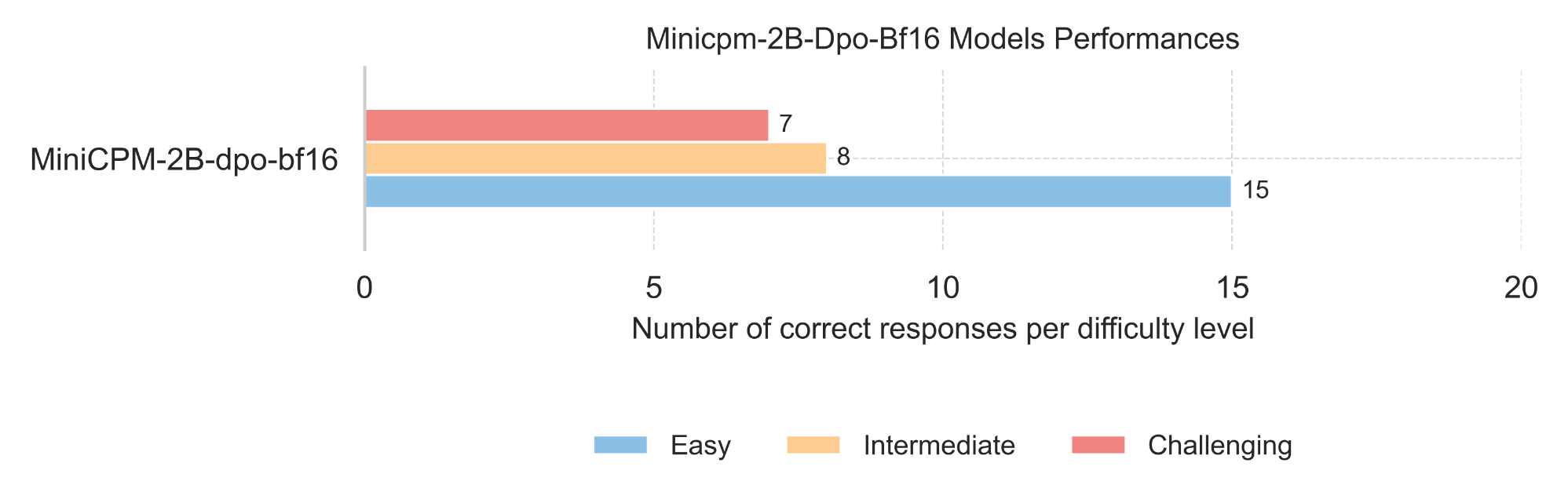}
        \caption{Number of correct answers obtained by the MiniCPM model across easy, intermediate, and challenging levels.}
        \label{fig:performance_minicpm}
    \end{figure}

    \paragraph{\textbf{MiniCPM-2B-dpo-bf16}} Lastly, we assessed the performance of the MiniCPM model, which outperformed models like Llama2-7B, Mistral-7B, and Llama2-13B in code and mathematical reasoning~\citep{minicpm2024}, including on the HumanEval dataset. While MiniCPM also surpassed Llama2-7B in our dataset, it did not consistently outperform all Mistral variants. Specifically, MiniCPM scored higher in terms of correct responses compared to Mistral models up to $3$ bits. However, Mistral models from $4$ bits and above achieved better results. It is important to note that MiniCPM evaluation was impacted by its non-compliance to the expected output format. Despite this penalization, MiniCPM achieved $50$\% of correct responses, $36.67$\% of passable responses, and $13.33$\% of incorrect answers, as shown in Figure ~\ref{fig:performance_minicpm}.

    \paragraph{\textbf{Size, RAM and Inference Time}} The analysis presented in Table~\ref{tab:results_models_ours} offers an overview of the performances and system requirements of various models. Notably, the models that need GPU to run the inference, are the heaviest exceeding the 15GB in space. With respect to quantize models, the heaviest models are zephyr-7b-beta.Q8\_0, dolphin-2.6-mistral-7b.Q8\_0, and openhermes-2.5-mistral-7b.Q8\_0, with a size of $7.70$ GB, which is still manageable on a conventional machine. In terms of RAM usage, MiniCPM is the most demanding, requiring $10.9$ GB. This moderate use of resources allows for the possibility of performing other tasks in parallel, thus optimizing machine utilization.

     The inference times vary among the models. The Phi family of models is the fastest, followed by Mistral. Conversely,  \textcolor{black}{Meta-Llama-3.1-8B-Instruct quantize models} exhibits the longest inference time, followed by   \textcolor{black}{MiniCPM} and  \textcolor{black}{Dolphin}. Overall, Mistral stands out as offering the best balance between size, RAM usage, inference speed, and overall performance on this dataset.

\section{Results and Discussion: HumanEval and EvalPlus dataset}
\label{sec:results_human_eval}

    \begin{table*}[!ht]
        \centering
        \resizebox{\linewidth}{!}{
        
        \begin{tabular}{|c|l|c|c|c|}
        \cline{4-5}
        \multicolumn{3}{c|}{}   & \multicolumn{1}{c|}{\textbf{HumanEval}}  & \multicolumn{1}{c|}{\textbf{EvalPlus }}     \\ \hline
        & \textbf{Models}  & \makecell{\textbf{Inference}\\ \textbf{Time (ms) $\downarrow$}}
        & \makecell{\textbf{Pass@1} \\ \textbf{0-shot $\uparrow$}} 
        & \makecell{\textbf{Pass@1} \\ \textbf{0-shot $\uparrow$}}
        \\ \hline
        \multirow{5}{*}{\rotatebox[origin=c]{90}{\textcolor{black}{Closed}}} & \textcolor{black}{Claude 3.5 Sooner} & $\times$ & \textcolor{black}{\textbf{92.0}} & \textcolor{black}{82.3} \\
        & GPT-4-Turbo (April 2024) & $\times$ & 90.2 & \textbf{86.6} \\ 
        & \textcolor{black}{Gemini 1.5 Pro} & $\times$ & \textcolor{black}{84.1} & \textcolor{black}{77.25} \\
        & GPT-3.5-Turbo (Nov 2023)  & $\times$ & 76.8 & 70.7 \\ 
         & Gemini 1.0 Pro  &$\times$ & 63.4  & 55.5 \\ 
         \hline
        \multirow{4}{*}{\rotatebox[origin=c]{90}{\textcolor{black}{GPU}}} & \textcolor{black}{Meta-Llama-3.1-8B-Instruct} & \textcolor{black}{28.26} & \textcolor{black}{\textbf{60.4}} & \textcolor{black}{\textbf{57.35}} \\
        
        & \textcolor{black}{Mistral-7B-Instruct-v0.2} & \textcolor{black}{14.96} & \textcolor{black}{42.10} & \textcolor{black}{36.00} \\
        
        & \textcolor{black}{dolphin-2.6-mistral-7b} & \textcolor{black}{8.89} & \textcolor{black}{60.25} & \textcolor{black}{52.32} \\
        
        & \textcolor{black}{OpenHermes-2.5-Mistral-7B} & \textcolor{black}{7.56} & \textcolor{black}{50.12} & \textcolor{black}{46.25} \\
        
        \hline
        \multirow{20}{*}{\rotatebox[origin=c]{90}{\textcolor{black}{CPU}}} & \textcolor{black}{Meta-Llama-3.1-8B-Instruct-Q4\_K\_M} & \textcolor{black}{458.25} & \textcolor{black}{54.2} & \textcolor{black}{50.7} \\
        & \textcolor{black}{Meta-Llama-3.1-8B-Instruct-Q5\_K\_M} & \textcolor{black}{478.12} & \textcolor{black}{\textbf{54.32}} & \textcolor{black}{\textbf{51.87}} \\
        
        \cline{2-5}
         & llama-2-coder-7b.Q3\_K\_M & 209.93 & 11.52 &  10.21 \\ 
        
         & llama-2-coder-7b.Q3\_K\_L & 224.33  & 12.52 & 10.87  \\ 

         & llama-2-coder-7b.Q4\_K\_M & 279.63 & \textbf{14.12} &  \textbf{11.36} \\ 

        \cline{2-5}
        
        
        & mistral-7b-instruct-v0.2.Q2\_K & 188.38  & 15.85 &  14.02 \\ 
        
        & mistral-7b-instruct-v0.2.Q3\_K\_S  & 207.47 & \textbf{32.92}  &  \textbf{28.65} \\ 
        
        & mistral-7b-instruct-v0.2.Q3\_K\_M & 233.43 & 27.40 &  25.00 \\ 
        
        & mistral-7b-instruct-v0.2.Q3\_K\_L & 240.58  & 23.78 &  21.34 \\ 
        
        & mistral-7b-instruct-v0.2.Q4\_K\_S & 242.29  & 23.17 &  21.95 \\ 
        
        & mistral-7b-instruct-v0.2.Q4\_K\_M & 245.57 & 23.17  &  24.39\\ 
        
        & mistral-7b-instruct-v0.2.Q5\_K\_M  & 292.74 & 13.41 & 12.19 \\

        \cline{2-5}
        
        & mistral-7b-openorca.Q3\_K\_L & 219.98 & \textbf{34.75} & \textbf{32.31}  \\ %
        \cline{2-5}
        
        
        
        & dolphin-2.6-mistral-7b.Q5\_K\_M & 288.57 & 51.82 & 45.12 \\
        
        & dolphin-2.6-mistral-7b.Q6\_K  &  332.83 & \textbf{54.26} &  \textbf{48.17} \\
        
        & dolphin-2.6-mistral-7b.Q8\_0  & 425.05 & 49.39 &  44.51\\
        \cline{2-5}
        & openhermes-2.5-mistral-7b.Q4\_K\_M  & 265.82 & 43.29 &  35.97 \\
        
        & openhermes-2.5-mistral-7b.Q5\_K\_M  & 294.62 &  \textbf{45.12} &  \textbf{37.80} \\ %
        
        & openhermes-2.5-mistral-7b.Q6\_K  & 328.12  & 43.90 &  36.58 \\
        
        & openhermes-2.5-mistral-7b.Q8\_0  &  426.11 & 39.63  &  32.31 \\

        \hline
        
        \end{tabular}
       }
        \caption{Evaluation on HumanEval and EvalPlus datasets. $Q_j$: quantization using $j$ bit width, K: the use of k-means clustering in the quantization, $S$, $M$, $L$: Small, Medium, and Large model size after quantization. Best results of each category are in bold.}
        \label{tab:results_models_humaneval_evalplus}
    \end{table*}

    In this section, we discuss the results obtained by the evaluated models on the HumanEval and EvalPlus datasets.  Our primary reference for comparison is the EvalPlus Leaderboard~\citep{evalplus_scores}, as outlined by~\citet{liu2023_evalGPTcode}. However, it is noteworthy that the samples in this leaderboard were generated from scratch and  \textcolor{black}{underwent preprocessing} with a sanitizer script. 

    To ensure a fair comparison with the scores reported on the EvalPlus Leaderboard, we used the greedy search decoding algorithm to generate code outputs. We evaluated both quantized models \textcolor{black}{and the original, missing models that require GPU for inference}, as shown in Table~\ref{tab:results_models_humaneval_evalplus}. Additionally, we assessed code correctness using the pass@1 metric and and the $0$-shot approach in the input prompts.

    \paragraph{\textbf{\textcolor{black}{Claude 3.5 Sooner} vs ChatGPT-3.5 vs ChatGPT-4 vs Gemini \textcolor{black}{1.0-1.5 Pro}}}

    \textcolor{black}{According to \citep{dubey2024_llama3herdmodels, ClaudeSonner_2024} and  Table~\ref{tab:results_models_humaneval_evalplus},Claude 3.5 Sonnet outperforms both ChatGPT-3.5, ChatGPT-4, and the Gemini models in the HumanEval dataset, achieving 92.0\% accuracy. GPT-4-Turbo follows closely with 90.2\%.Gemini 1.5 Pro shows a significant drop in performance, scoring 84.1\%, which is 7.3\% higher than GPT-3.5-Turbo's 76.8\%. Finally, Gemini Pro 1.0 ranks fifth with 63.4\% accuracy.}

    \textcolor{black}{In the EvalPlus evaluation, the rankings shifted. GPT-4-Turbo leads at 86.6\% accuracy, followed by Claude 3.5 Sonnet at 82.3\%. The other positions remain consistent: Gemini 1.5 Pro is third with 77.25\%, GPT-3.5-Turbo is fourth with 70.7\%, and Gemini Pro 1.0 is fifth with 55.5\%.}

    \textcolor{black}{These results highlight the rapid advancements in language model capabilities, especially in code generation. Since the release of ChatGPT-3.5 in November 2022, the community has achieved impressive progress in less than two years.}
   
    \paragraph{\textbf{LLaMA Models}} \textcolor{black}{In this set of experiments, we evaluated the Meta-Llama-3.1-8B-Instruct model alongside two of its quantized variants (4-bit and 5-bit), as well as three quantized versions of the Llama-2-Coder-7B model (Q3\_K\_M, Q3\_K\_L, Q4\_K\_M). As shown in Table~\ref{tab:results_models_humaneval_evalplus}, the non-quantized Meta-Llama-3.1-8B-Instruct achieved the highest scores in both the HumanEval and EvalPlus datasets, with 60.4\% and 57.35\% accuracy, respectively. The quantized versions were close behind, with the 4-bit and 5-bit models scoring 54.2\% and 50.7\%, and 54.32\% and 51.87\%, respectively.}

    \textcolor{black}{In contrast, the Llama-2-Coder-7B quantized models recorded the lowest scores in both datasets. In the HumanEval dataset, they achieved 11.52\%, 12.52\%, and 14.12\% accuracy for the Q3\_K\_M, Q3\_K\_L, and Q4\_K\_M models, respectively. Performance on the EvalPlus dataset was even lower, with the Llama-2-Coder-7B.Q4\_K\_M emerging as the most effective variant at 11.36\% accuracy.}

    \textcolor{black}{These results highlight the superiority of LLaMA-3.1 in Python code generation tasks. Despite being specialized for coding, the Llama-2-Coder-7B model was outperformed by the Llama-3.1-8B-Instruct, and our findings consistently revealed that the code produced by Llama-2-Coder was frequently incomplete and riddled with syntax errors.}

    \paragraph{\textbf{Mistral}} \textcolor{black}{In this set of experiments, we evaluate the performance of the non-quantized Mistral-7B-Instruct-v0.2 model and its quantized versions. According to the EvalPlus Leaderboard and Table~\ref{tab:results_models_humaneval_evalplus}, the non-quantized model achieved 42.1\% accuracy on HumanEval and 36\% on EvalPlus. The performance difference between the non-quantized and quantized models is minimal. For instance,} the best quantized performance on the HumanEval dataset comes from the Mistral-7B-Instruct-v0.2.Q3\_K\_S model with a pass@1 score of 32.92\%, followed by Mistral-7B-Instruct-v0.2.Q3\_K\_M at 27.40\%. Surprisingly, the lowest scores were obtained by Mistral-7B-Instruct-v0.2.Q2\_K and Mistral-7B-Instruct-v0.2.Q5\_K\_M, with pass@1 scores of 15.85\% and 13.41\% respectively. This highlights, once again, that quantizing with more bits does not necessarily lead to better results. 

    Unsurprisingly, the quantized Mistral models showed lower performance on the EvalPlus dataset, which includes a greater number of unit tests. The models achieved an average score 1.6 points lower than on the HumanEval dataset. Nevertheless, the overall ranking of the tested models remained consistent across both datasets.
    
    Finally, we evaluated the performance of the Mistral-7b-openorca.Q3\_K\_L model, where it surpassed other quantized Mistral variants with accuracies of 34.75\% and 32.31\%, respectively. However, as detailed in Section~\ref{sec:results}, its performance on our dataset was significantly lower than that of other quantized Mistral models due to penalties for failing to generate outputs in the correct format. This discrepancy highlights the model's difficulties in interpreting input prompts with restrictive conditions for generating Python code.

    \paragraph{\textbf{dolphin-2.6-mistral-7b}} In our evaluation, \textcolor{black}{we assessed the non-quantized dolphin-2.6-mistral-7b model alongside its 5-bit, 6-bit, and 8-bit quantized versions. Unsurprisingly, the non-quantized model achieved the highest accuracy, scoring 60.25\% on HumanEval and 52.32\% on EvalPlus, with only minor differences compared to the quantized versions.  All dolphin models outperformed the mistral models.} On HumanEval, the 6-bit model led the quantized versions with 54.26\%, followed by the 5-bit model at 51.82\%, and the 8-bit model at 49.39\%. Similarly, in EvalPlus, the 6-bit model scored highest among the quantized versions with 48.17\%, while the 8-bit model had the lowest score at 44.51\%.

    These findings show that the dolphin-2.6-mistral-7b quantized models outperformed the Mistral quantized models in Python code generation across both the HumanEval and EvalPlus datasets. However, it's worth noting that on our dataset, the dolphin-2.6-mistral-7b model scored lower than Mistral. It's important to note that although dolphin-2.6-mistral-7b generates correct code, it was penalized for not following the specific output format we provided. This highlights the importance to adhere to the correct format, even if the code is functionally correct. Thus, while Mistral excels in understanding the required output format, dolphin-2.6 demonstrates a high capability in generating Python code but often fails to deliver it in the prescribed format.

    \paragraph{\textbf{openhermes-2.5-mistral-7}} \textcolor{black}{The evaluation results for the non-quantized OpenHermes-2.5-Mistral-7B model were 50.12\% on HumanEval and 46.25\% on EvalPlus. These results are competitive compared to more complex models like Meta-Llama-3.1-8B-Instruct, which scored 60.4\% and 57.35\%, respectively. Additionally,} the quantized OpenHermes-2.5-Mistral-7B outperformed the quantized Mistral models. However, it scores lower than the dolphin-2.6-mistral-7b models, which is expected since the latter were trained using code datasets, enhancing their performance in such tasks. Although mistral models achieved better scores in our dataset but lower scores on HumanEval and EvalPlus, it's important to note that openhermes-2.5-mistral-7b also produces correct and functional codes. However, similarly to dolphin-2.6-mistral-7b models, these were also penalized due to not adhering to the required output format, resulting in a reduced score of 0.5. 

    \textcolor{black}{Consistent with our previous findings, the openhermes-2.5-mistral-7 model often struggles with variable name consistency when generating code. Although it adheres to the variable names specified in the prompt, it frequently introduces errors when using new variables. For example, it might refer to the same variable as 'time' initially and then as 'times' later, leading to failures in unit tests. It would be advantageous to develop strategies to address this issue, such as through post-processing or improved token generation techniques.}

    \paragraph{\textbf{Inference Time}}Table~\ref{tab:results_models_humaneval_evalplus} shows minimal differences in inference times among the evaluated models. Mistral and LLaMa are the fastest, with a small difference of 104.36 milliseconds between Mistral's 2-bit and 8-bit quantized versions. The  \textcolor{black}{Meta-Llama-3.1-8B-Instruct} family of models have longer inference times on average, with the  \textcolor{black}{5}-bit quantized version being the slowest at  \textcolor{black}{478.12} milliseconds. These results highlight the efficiency and practicality of these CPU-friendly models in terms of both inference speed and performance.

\section{Conclusions and Future Directions}
\label{sec:Conclusions}

    In this paper, we examine the  \textcolor{black}{effectiveness} of CPU-friendly models in generating Python code  \textcolor{black}{under} various scenarios. Specifically, our dataset tasked models with generating Python code  \textcolor{black}{based on} a given problem statement, using predetermined variable names and returning a specified set of options included in the input prompt.  \textcolor{black}{In contrast}, in the HumanEval and EvalPlus datasets, the challenge was to generate Python code from the function's signature and docstring, allowing us to assess the functional correctness of LLM-synthesized code.
    
    Our evaluation of CPU-friendly models across different datasets assesses not only their ability to generate correct Python code but also  \textcolor{black}{how well they followed the} specified variable names and output formats.

    \textcolor{black}{The advantages of this work are to: (1) facilitate performance comparisons of CPU-friendly models, enabling individuals or small companies with limited computational resources to better choose a smaller model as an alternative to large, costly models; (2) propose a new Python generation dataset consisting of 60 programming problems. This dataset extends the existing benchmark datasets, HumanEval and EvalPlus, allowing for deeper understanding of these models; (3) propose engineered prompts for generating and evaluating Python code; (4) propose a semi-manual evaluation of state-of-the-art methods for Python code generation; and (5) make all of our work, experiments, and dataset publicly available to advance research in code generation and provide access to a thorough analysis of CPU-friendly models, which can serve as strong alternatives to very large or closed models. We define a set of disadvantages in Section \ref{sec: limitations}.}

     \textcolor{black}{Our} findings reveal two key points: 1) CPU-friendly models achieve competitive scores compared to advanced chatbot models like ChatGPT-3.5, ChatGPT-4, and Gemini, which require significant GPU resources for processing; 2) some models, such as dolphin-2.6-mistral-7b, excel in generating Python code but fall short in meeting the required output formats. Often, even if the generated code is correct, these models are penalized for not returning the specified options from the input prompt. A similar issue affects the openhermes-2.5-mistral-7b model, albeit with slightly lower scores than dolphin-2.6-mistral-7b. In contrast, models like Mistral demonstrate consistent performance in understanding prompts and generating code, leading to superior results on our dataset, though they perform worse on HumanEval and EvalPlus datasets.

    Remarkably, llama.cpp project has  \textcolor{black}{made these} competitive outcomes using standard computers \textcolor{black}{possible}, a development that seemed highly unlikely just a few months or years ago. This progress highlights the feasibility of running sophisticated language models on CPUs today. However, it is important to  \textcolor{black}{mention} that we did not utilize more powerful models such as Mixtral, an enhanced version of Mistral. Despite existing quantized versions, Mixtral still requires considerable computational resources which our machines could not support. For instance, the 2-bit quantized version of Mixtral requires 15.6 GB of storage and 18.14 GB of RAM, while the 8-bit version needs 49.62 GB of storage and 52.12 GB of RAM. Thus, these models do not align with our  \textcolor{black}{objective} of demonstrating the performance of CPU-friendly models on  \textcolor{black}{standard} machines.

    \textcolor{black}{In our future work, we plan to explore we plan to explore the expanded capabilities of CPU-friendly models across various code-related tasks, such as defect detection, cloze tests, code refinement, and code translation. Our key focus will be on creating a comprehensive framework to evaluate the performance of diverse LLMs in their quantized forms at different bit levels. This framework will integrate the Automatic Prompt Engineer (APE)~\citep{zhou2023_APE_framework} framework, which generates various candidate prompts, executes them to generate responses, and evaluates the responses to select the most effective prompt. This integration can effectively enhance the quality of the prompts used in our experiments.}

    \textcolor{black}{Moreover, the proposed framework will allow for a thorough and systematic analysis of how quantization affects both precision and task-specific outcomes in code-related contexts. Through this approach, we aim to gain deeper insights into the trade-offs between computational efficiency and model accuracy, ultimately guiding the optimization of LLMs for code-based applications.}
    
\section{\textbf{Challenges and Limitations}}
\label{sec: limitations}
    \textcolor{black}{Our work faces three main limitations. The first is the rapid emergence of new LLMs. At the time of this publication, several new models have been introduced, and we cannot yet assess the performance of their quantized versions. The second limitation concerns the dataset used. Our dataset presents a significant challenge for LLMs, as it requires them to accurately relate and comprehend three key elements: the problem, the context, and the possible solutions. This complexity tests the models' ability to understand natural language and correctly generate Python code.}

    \textcolor{black}{Another limitation of our work is that LLMs are usually trained on high-level imperative programming languages like Python, resulting in the under-representation of low-resource, declarative, and low-level languages. This bias limits the model's versatility across different programming paradigms. A comprehensive review of cross-language generation and providing a specialized dataset for these underrepresented languages would be essential to address this gap and study the model's applicability in more languages.}

    \textcolor{black}{Similarly, existing datasets like HumanEval and EvalPlus are designed to evaluate LLMs' ability to generate Python code based on natural language prompts. However, these datasets lack a variety of scenarios that would more thoroughly test the capabilities of quantized models in code generation, such as varying the structure of input text. This gap highlights the need for more comprehensive evaluation methods to fully understand the impact of quantization on code generation tasks.}

\textbf{Acknowledgments}

We would like to acknowledge Novelis for their support in publishing this article. We are especially grateful for the assistance and
contributions of their research team.



\newpage
\bibliographystyle{elsarticle-harv} 
\bibliography{thebibliography}

\begin{thebibliography}{109}
\expandafter\ifx\csname natexlab\endcsname\relax\def\natexlab#1{#1}\fi
\providecommand{\url}[1]{\texttt{#1}}
\providecommand{\href}[2]{#2}
\providecommand{\path}[1]{#1}
\providecommand{\DOIprefix}{doi:}
\providecommand{\ArXivprefix}{arXiv:}
\providecommand{\URLprefix}{URL: }
\providecommand{\Pubmedprefix}{pmid:}
\providecommand{\doi}[1]{\href{http://dx.doi.org/#1}{\path{#1}}}
\providecommand{\Pubmed}[1]{\href{pmid:#1}{\path{#1}}}
\providecommand{\bibinfo}[2]{#2}
\ifx\xfnm\relax \def\xfnm[#1]{\unskip,\space#1}\fi
\bibitem[{Agarap(2019)}]{agarap2019_relu}
\bibinfo{author}{Agarap, A.F.}, \bibinfo{year}{2019}.
\newblock \bibinfo{title}{Deep learning using rectified linear units (relu)}.
\newblock \href{http://arxiv.org/abs/1803.08375}{{\tt arXiv:1803.08375}}.
\bibitem[{Aghajanyan et~al.(2022)Aghajanyan, Huang, Ross, Karpukhin, Xu, Goyal, Okhonko, Joshi, Ghosh, Lewis and Zettlemoyer}]{aghajanyan2022cm3}
\bibinfo{author}{Aghajanyan, A.}, \bibinfo{author}{Huang, B.}, \bibinfo{author}{Ross, C.}, \bibinfo{author}{Karpukhin, V.}, \bibinfo{author}{Xu, H.}, \bibinfo{author}{Goyal, N.}, \bibinfo{author}{Okhonko, D.}, \bibinfo{author}{Joshi, M.}, \bibinfo{author}{Ghosh, G.}, \bibinfo{author}{Lewis, M.}, \bibinfo{author}{Zettlemoyer, L.}, \bibinfo{year}{2022}.
\newblock \bibinfo{title}{Cm3: A causal masked multimodal model of the internet}.
\newblock \href{http://arxiv.org/abs/2201.07520}{{\tt arXiv:2201.07520}}.
\bibitem[{Ahmad et~al.(2021)Ahmad, Chakraborty, Ray and Chang}]{ahmad-etal-2021-PLBART}
\bibinfo{author}{Ahmad, W.}, \bibinfo{author}{Chakraborty, S.}, \bibinfo{author}{Ray, B.}, \bibinfo{author}{Chang, K.W.}, \bibinfo{year}{2021}.
\newblock \bibinfo{title}{Unified pre-training for program understanding and generation}, in: \bibinfo{booktitle}{Proceedings of the 2021 Conference of the North American Chapter of the Association for Computational Linguistics: Human Language Technologies}, \bibinfo{publisher}{Association for Computational Linguistics}, \bibinfo{address}{Online}. pp. \bibinfo{pages}{2655--2668}.
\newblock \URLprefix \url{https://aclanthology.org/2021.naacl-main.211}, \DOIprefix\doi{10.18653/v1/2021.naacl-main.211}.
\bibitem[{Ahmed et~al.(2023)Ahmed, Roy, Kajol, Hasan, Datta and Reza}]{ahmed2023_chatgptBard}
\bibinfo{author}{Ahmed, I.}, \bibinfo{author}{Roy, A.}, \bibinfo{author}{Kajol, M.}, \bibinfo{author}{Hasan, U.}, \bibinfo{author}{Datta, P.P.}, \bibinfo{author}{Reza, M.R.}, \bibinfo{year}{2023}.
\newblock \bibinfo{title}{Chatgpt vs. bard: a comparative study}.
\newblock \bibinfo{journal}{Authorea Preprints} .
\bibitem[{AI@Meta(2024)}]{llama3modelcard}
\bibinfo{author}{AI@Meta}, \bibinfo{year}{2024}.
\newblock \bibinfo{title}{Llama 3 model card}.
\newblock \URLprefix \url{https://github.com/meta-llama/llama3/blob/main/MODEL_CARD.md}.
\bibitem[{Ainslie et~al.(2023)Ainslie, Lee-Thorp, de~Jong, Zemlyanskiy, Lebron and Sanghai}]{Ainslie2023_GQA}
\bibinfo{author}{Ainslie, J.}, \bibinfo{author}{Lee-Thorp, J.}, \bibinfo{author}{de~Jong, M.}, \bibinfo{author}{Zemlyanskiy, Y.}, \bibinfo{author}{Lebron, F.}, \bibinfo{author}{Sanghai, S.}, \bibinfo{year}{2023}.
\newblock \bibinfo{title}{{GQA}: Training generalized multi-query transformer models from multi-head checkpoints}, in: \bibinfo{editor}{Bouamor, H.}, \bibinfo{editor}{Pino, J.}, \bibinfo{editor}{Bali, K.} (Eds.), \bibinfo{booktitle}{Proceedings of the 2023 Conference on Empirical Methods in Natural Language Processing}, \bibinfo{publisher}{Association for Computational Linguistics}, \bibinfo{address}{Singapore}. pp. \bibinfo{pages}{4895--4901}.
\newblock \URLprefix \url{https://aclanthology.org/2023.emnlp-main.298}, \DOIprefix\doi{10.18653/v1/2023.emnlp-main.298}.
\bibitem[{Anand et~al.(2023)Anand, Nussbaum, Duderstadt, Schmidt and Mulyar}]{gpt4all}
\bibinfo{author}{Anand, Y.}, \bibinfo{author}{Nussbaum, Z.}, \bibinfo{author}{Duderstadt, B.}, \bibinfo{author}{Schmidt, B.}, \bibinfo{author}{Mulyar, A.}, \bibinfo{year}{2023}.
\newblock \bibinfo{title}{Gpt4all: Training an assistant-style chatbot with large scale data distillation from gpt-3.5-turbo}.
\newblock \bibinfo{howpublished}{\url{https://github.com/nomic-ai/gpt4all}}.
\bibitem[{Anthropic(2024)}]{ClaudeSonner_2024}
\bibinfo{author}{Anthropic}, \bibinfo{year}{2024}.
\newblock \bibinfo{title}{Claude 3.5 sonnet model card addendum}.
\newblock \URLprefix \url{https://www-cdn.anthropic.com/fed9cc193a14b84131812372d8d5857f8f304c52/Model_Card_Claude_3_Addendum.pdf}.
\bibitem[{Austin et~al.(2021)Austin, Odena, Nye, Bosma, Michalewski, Dohan, Jiang, Cai, Terry, Le and Sutton}]{austin2021_program}
\bibinfo{author}{Austin, J.}, \bibinfo{author}{Odena, A.}, \bibinfo{author}{Nye, M.}, \bibinfo{author}{Bosma, M.}, \bibinfo{author}{Michalewski, H.}, \bibinfo{author}{Dohan, D.}, \bibinfo{author}{Jiang, E.}, \bibinfo{author}{Cai, C.}, \bibinfo{author}{Terry, M.}, \bibinfo{author}{Le, Q.}, \bibinfo{author}{Sutton, C.}, \bibinfo{year}{2021}.
\newblock \bibinfo{title}{Program synthesis with large language models}.
\newblock \href{http://arxiv.org/abs/2108.07732}{{\tt arXiv:2108.07732}}.
\bibitem[{Ba et~al.(2016)Ba, Kiros and Hinton}]{ba2016_layernormalization}
\bibinfo{author}{Ba, J.L.}, \bibinfo{author}{Kiros, J.R.}, \bibinfo{author}{Hinton, G.E.}, \bibinfo{year}{2016}.
\newblock \bibinfo{title}{Layer normalization}.
\newblock \href{http://arxiv.org/abs/1607.06450}{{\tt arXiv:1607.06450}}.
\bibitem[{Beltagy et~al.(2020)Beltagy, Peters and Cohan}]{beltagy2020_longformer}
\bibinfo{author}{Beltagy, I.}, \bibinfo{author}{Peters, M.E.}, \bibinfo{author}{Cohan, A.}, \bibinfo{year}{2020}.
\newblock \bibinfo{title}{Longformer: The long-document transformer}.
\newblock \URLprefix \url{https://arxiv.org/abs/2004.05150}, \href{http://arxiv.org/abs/2004.05150}{{\tt arXiv:2004.05150}}.
\bibitem[{Brown et~al.(2020)Brown, Mann, Ryder, Subbiah, Kaplan, Dhariwal, Neelakantan, Shyam, Sastry, Askell and et~al.}]{Brown2020_LLMs-ZeroShot_GPT3}
\bibinfo{author}{Brown, T.}, \bibinfo{author}{Mann, B.}, \bibinfo{author}{Ryder, N.}, \bibinfo{author}{Subbiah, M.}, \bibinfo{author}{Kaplan, J.D.}, \bibinfo{author}{Dhariwal, P.}, \bibinfo{author}{Neelakantan, A.}, \bibinfo{author}{Shyam, P.}, \bibinfo{author}{Sastry, G.}, \bibinfo{author}{Askell, A.}, \bibinfo{author}{et~al.}, \bibinfo{year}{2020}.
\newblock \bibinfo{title}{Language models are few-shot learners}, in: \bibinfo{editor}{Larochelle, H.}, \bibinfo{editor}{Ranzato, M.}, \bibinfo{editor}{Hadsell, R.}, \bibinfo{editor}{Balcan, M.}, \bibinfo{editor}{Lin, H.} (Eds.), \bibinfo{booktitle}{Advances in Neural Information Processing Systems}, \bibinfo{publisher}{Curran Associates, Inc.}. pp. \bibinfo{pages}{1877--1901}.
\newblock \URLprefix \url{https://proceedings.neurips.cc/paper_files/paper/2020/file/1457c0d6bfcb4967418bfb8ac142f64a-Paper.pdf}.
\bibitem[{Chee et~al.(2023)Chee, Cai, Kuleshov and Sa}]{chee2023quip}
\bibinfo{author}{Chee, J.}, \bibinfo{author}{Cai, Y.}, \bibinfo{author}{Kuleshov, V.}, \bibinfo{author}{Sa, C.D.}, \bibinfo{year}{2023}.
\newblock \bibinfo{title}{Qu{IP}: 2-bit quantization of large language models with guarantees}.
\newblock \bibinfo{journal}{Thirty-seventh Conference on Neural Information Processing Systems} \URLprefix \url{https://openreview.net/forum?id=xrk9g5vcXR}.
\bibitem[{Chen et~al.(2024)Chen, Shao, Xu, Wang, Gao, Zhang, Qiao and Luo}]{chen2024efficientqat}
\bibinfo{author}{Chen, M.}, \bibinfo{author}{Shao, W.}, \bibinfo{author}{Xu, P.}, \bibinfo{author}{Wang, J.}, \bibinfo{author}{Gao, P.}, \bibinfo{author}{Zhang, K.}, \bibinfo{author}{Qiao, Y.}, \bibinfo{author}{Luo, P.}, \bibinfo{year}{2024}.
\newblock \bibinfo{title}{Efficientqat: Efficient quantization-aware training for large language models}.
\newblock \bibinfo{journal}{arXiv preprint arXiv:2407.11062} .
\bibitem[{Chen et~al.(2021)Chen, Tworek, Jun, Yuan, Pinto, Kaplan, Edwards, Burda, Joseph and Brockman}]{Chen2021_EvaluatingLL}
\bibinfo{author}{Chen, M.}, \bibinfo{author}{Tworek, J.}, \bibinfo{author}{Jun, H.}, \bibinfo{author}{Yuan, Q.}, \bibinfo{author}{Pinto, H.P.d.O.}, \bibinfo{author}{Kaplan, J.}, \bibinfo{author}{Edwards, H.}, \bibinfo{author}{Burda, Y.}, \bibinfo{author}{Joseph, N.}, \bibinfo{author}{Brockman, G.e.a.}, \bibinfo{year}{2021}.
\newblock \bibinfo{title}{Evaluating large language models trained on code}.
\newblock \bibinfo{journal}{ArXiv} \bibinfo{volume}{abs/2107.03374}.
\newblock \URLprefix \url{https://api.semanticscholar.org/CorpusID:235755472}.
\bibitem[{Chen(2023)}]{chen2023_few1shot}
\bibinfo{author}{Chen, W.}, \bibinfo{year}{2023}.
\newblock \bibinfo{title}{Large language models are few(1)-shot table reasoners}.
\newblock \href{http://arxiv.org/abs/2210.06710}{{\tt arXiv:2210.06710}}.
\bibitem[{Chiang et~al.(2023)Chiang, Li, Lin, Sheng, Wu, Zhang, Zheng, Zhuang, Zhuang, Gonzalez, Stoica and Xing}]{vicuna2023}
\bibinfo{author}{Chiang, W.L.}, \bibinfo{author}{Li, Z.}, \bibinfo{author}{Lin, Z.}, \bibinfo{author}{Sheng, Y.}, \bibinfo{author}{Wu, Z.}, \bibinfo{author}{Zhang, H.}, \bibinfo{author}{Zheng, L.}, \bibinfo{author}{Zhuang, S.}, \bibinfo{author}{Zhuang, Y.}, \bibinfo{author}{Gonzalez, J.E.}, \bibinfo{author}{Stoica, I.}, \bibinfo{author}{Xing, E.P.}, \bibinfo{year}{2023}.
\newblock \bibinfo{title}{Vicuna: An open-source chatbot impressing gpt-4 with 90\%* chatgpt quality}.
\newblock \URLprefix \url{https://lmsys.org/blog/2023-03-30-vicuna/}.
\bibitem[{Chowdhery et~al.(2022)Chowdhery, Narang, Devlin, Bosma, Mishra, Roberts, Barham, Chung, Sutton, Gehrmann and et~al.}]{chowdhery2022_palm}
\bibinfo{author}{Chowdhery, A.}, \bibinfo{author}{Narang, S.}, \bibinfo{author}{Devlin, J.}, \bibinfo{author}{Bosma, M.}, \bibinfo{author}{Mishra, G.}, \bibinfo{author}{Roberts, A.}, \bibinfo{author}{Barham, P.}, \bibinfo{author}{Chung, H.W.}, \bibinfo{author}{Sutton, C.}, \bibinfo{author}{Gehrmann, S.}, \bibinfo{author}{et~al.}, \bibinfo{year}{2022}.
\newblock \bibinfo{title}{Palm: Scaling language modeling with pathways}.
\newblock \href{http://arxiv.org/abs/2204.02311}{{\tt arXiv:2204.02311}}.
\bibitem[{Christiano et~al.(2017)Christiano, Leike, Brown, Martic, Legg and Amodei}]{Christiano2017_RL}
\bibinfo{author}{Christiano, P.F.}, \bibinfo{author}{Leike, J.}, \bibinfo{author}{Brown, T.}, \bibinfo{author}{Martic, M.}, \bibinfo{author}{Legg, S.}, \bibinfo{author}{Amodei, D.}, \bibinfo{year}{2017}.
\newblock \bibinfo{title}{Deep reinforcement learning from human preferences}.
\newblock \bibinfo{journal}{Advances in Neural Information Processing Systems} \bibinfo{volume}{30}.
\newblock \URLprefix \url{https://proceedings.neurips.cc/paper_files/paper/2017/file/d5e2c0adad503c91f91df240d0cd4e49-Paper.pdf}.
\bibitem[{Cobbe et~al.(2021)Cobbe, Kosaraju, Bavarian, Chen, Jun, Kaiser, Plappert, Tworek, Hilton, Nakano, Hesse and Schulman}]{cobbe2021training}
\bibinfo{author}{Cobbe, K.}, \bibinfo{author}{Kosaraju, V.}, \bibinfo{author}{Bavarian, M.}, \bibinfo{author}{Chen, M.}, \bibinfo{author}{Jun, H.}, \bibinfo{author}{Kaiser, L.}, \bibinfo{author}{Plappert, M.}, \bibinfo{author}{Tworek, J.}, \bibinfo{author}{Hilton, J.}, \bibinfo{author}{Nakano, R.}, \bibinfo{author}{Hesse, C.}, \bibinfo{author}{Schulman, J.}, \bibinfo{year}{2021}.
\newblock \bibinfo{title}{Training verifiers to solve math word problems}.
\newblock \href{http://arxiv.org/abs/2110.14168}{{\tt arXiv:2110.14168}}.
\bibitem[{Coursera(2023)}]{coursera2023_pl}
\bibinfo{author}{Coursera}, \bibinfo{year}{2023}.
\newblock \bibinfo{title}{Most popular programming languages in 2024}.
\newblock \URLprefix \url{https://www.coursera.org/articles/popular-programming-languages}.
\bibitem[{Cui et~al.(2023)Cui, Yuan, Ding, Yao, Zhu, Ni, Xie, Liu and Sun}]{cui2023_ultrafeedback}
\bibinfo{author}{Cui, G.}, \bibinfo{author}{Yuan, L.}, \bibinfo{author}{Ding, N.}, \bibinfo{author}{Yao, G.}, \bibinfo{author}{Zhu, W.}, \bibinfo{author}{Ni, Y.}, \bibinfo{author}{Xie, G.}, \bibinfo{author}{Liu, Z.}, \bibinfo{author}{Sun, M.}, \bibinfo{year}{2023}.
\newblock \bibinfo{title}{Ultrafeedback: Boosting language models with high-quality feedback}.
\newblock \href{http://arxiv.org/abs/2310.01377}{{\tt arXiv:2310.01377}}.
\bibitem[{Dai et~al.(2019)Dai, Yang, Yang, Carbonell, Le and Salakhutdinov}]{dai2019_transformerxl}
\bibinfo{author}{Dai, Z.}, \bibinfo{author}{Yang, Z.}, \bibinfo{author}{Yang, Y.}, \bibinfo{author}{Carbonell, J.}, \bibinfo{author}{Le, Q.V.}, \bibinfo{author}{Salakhutdinov, R.}, \bibinfo{year}{2019}.
\newblock \bibinfo{title}{Transformer-xl: Attentive language models beyond a fixed-length context}.
\newblock \href{http://arxiv.org/abs/1901.02860}{{\tt arXiv:1901.02860}}.
\bibitem[{Daniele and Suphavadeeprasit(2023)}]{daniele2023amplify-instruct}
\bibinfo{author}{Daniele, L.}, \bibinfo{author}{Suphavadeeprasit}, \bibinfo{year}{2023}.
\newblock \bibinfo{title}{Amplify-instruct: Synthetically generated diverse multi-turn conversations for effecient llm training.}
\newblock \bibinfo{journal}{arXiv preprint arXiv:(coming soon)} \URLprefix \url{https://huggingface.co/datasets/LDJnr/Capybara}.
\bibitem[{Dao et~al.(2022)Dao, Fu, Ermon, Rudra and R{\'e}}]{dao2022_flashattention}
\bibinfo{author}{Dao, T.}, \bibinfo{author}{Fu, D.}, \bibinfo{author}{Ermon, S.}, \bibinfo{author}{Rudra, A.}, \bibinfo{author}{R{\'e}, C.}, \bibinfo{year}{2022}.
\newblock \bibinfo{title}{Flashattention: Fast and memory-efficient exact attention with io-awareness}.
\newblock \bibinfo{journal}{Advances in Neural Information Processing Systems} \bibinfo{volume}{35}, \bibinfo{pages}{16344--16359}.
\bibitem[{Dauphin et~al.(2017)Dauphin, Fan, Auli and Grangier}]{Dauphin2017_glu}
\bibinfo{author}{Dauphin, Y.N.}, \bibinfo{author}{Fan, A.}, \bibinfo{author}{Auli, M.}, \bibinfo{author}{Grangier, D.}, \bibinfo{year}{2017}.
\newblock \bibinfo{title}{Language modeling with gated convolutional networks}, in: \bibinfo{booktitle}{Proceedings of the 34th International Conference on Machine Learning - Volume 70}, \bibinfo{publisher}{JMLR.org}. p. \bibinfo{pages}{933–941}.
\bibitem[{Dery et~al.(2024)Dery, Kolawole, Kagey, Smith, Neubig and Talwalkar}]{dery2024everybody}
\bibinfo{author}{Dery, L.}, \bibinfo{author}{Kolawole, S.}, \bibinfo{author}{Kagey, J.F.}, \bibinfo{author}{Smith, V.}, \bibinfo{author}{Neubig, G.}, \bibinfo{author}{Talwalkar, A.}, \bibinfo{year}{2024}.
\newblock \bibinfo{title}{Everybody prune now: Structured pruning of llms with only forward passes}.
\newblock \bibinfo{journal}{arXiv preprint arXiv:2402.05406} .
\bibitem[{Dettmers et~al.(2023)Dettmers, Svirschevski, Egiazarian, Kuznedelev, Frantar, Ashkboos, Borzunov, Hoefler and Alistarh}]{dettmers2023spqr}
\bibinfo{author}{Dettmers, T.}, \bibinfo{author}{Svirschevski, R.}, \bibinfo{author}{Egiazarian, V.}, \bibinfo{author}{Kuznedelev, D.}, \bibinfo{author}{Frantar, E.}, \bibinfo{author}{Ashkboos, S.}, \bibinfo{author}{Borzunov, A.}, \bibinfo{author}{Hoefler, T.}, \bibinfo{author}{Alistarh, D.}, \bibinfo{year}{2023}.
\newblock \bibinfo{title}{Spqr: A sparse-quantized representation for near-lossless llm weight compression}.
\newblock \URLprefix \url{https://arxiv.org/abs/2306.03078}, \href{http://arxiv.org/abs/2306.03078}{{\tt arXiv:2306.03078}}.
\bibitem[{Devlin et~al.(2019)Devlin, Chang, Lee and Toutanova}]{devlin2019_bert}
\bibinfo{author}{Devlin, J.}, \bibinfo{author}{Chang, M.W.}, \bibinfo{author}{Lee, K.}, \bibinfo{author}{Toutanova, K.}, \bibinfo{year}{2019}.
\newblock \bibinfo{title}{{BERT}: Pre-training of deep bidirectional transformers for language understanding}, in: \bibinfo{editor}{Burstein, J.}, \bibinfo{editor}{Doran, C.}, \bibinfo{editor}{Solorio, T.} (Eds.), \bibinfo{booktitle}{Proceedings of the 2019 Conference of the North {A}merican Chapter of the Association for Computational Linguistics: Human Language Technologies, Volume 1 (Long and Short Papers)}, \bibinfo{publisher}{Association for Computational Linguistics}, \bibinfo{address}{Minneapolis, Minnesota}. pp. \bibinfo{pages}{4171--4186}.
\newblock \URLprefix \url{https://aclanthology.org/N19-1423}, \DOIprefix\doi{10.18653/v1/N19-1423}.
\bibitem[{Ding et~al.(2023)Ding, Chen, Xu, Qin, Zheng, Hu, Liu, Sun and Zhou}]{ding2023enhancing}
\bibinfo{author}{Ding, N.}, \bibinfo{author}{Chen, Y.}, \bibinfo{author}{Xu, B.}, \bibinfo{author}{Qin, Y.}, \bibinfo{author}{Zheng, Z.}, \bibinfo{author}{Hu, S.}, \bibinfo{author}{Liu, Z.}, \bibinfo{author}{Sun, M.}, \bibinfo{author}{Zhou, B.}, \bibinfo{year}{2023}.
\newblock \bibinfo{title}{Enhancing chat language models by scaling high-quality instructional conversations}.
\newblock \href{http://arxiv.org/abs/2305.14233}{{\tt arXiv:2305.14233}}.
\bibitem[{Dong et~al.(2023)Dong, Jiang, Jin and Li}]{dong2023selfcollaboration}
\bibinfo{author}{Dong, Y.}, \bibinfo{author}{Jiang, X.}, \bibinfo{author}{Jin, Z.}, \bibinfo{author}{Li, G.}, \bibinfo{year}{2023}.
\newblock \bibinfo{title}{Self-collaboration code generation via chatgpt}.
\newblock \href{http://arxiv.org/abs/2304.07590}{{\tt arXiv:2304.07590}}.
\bibitem[{Dubey et~al.(2024)Dubey, Jauhri, Pandey, Kadian, Al-Dahle, Letman, Mathur, Schelten, Yang, Fan and et~al.}]{dubey2024_llama3herdmodels}
\bibinfo{author}{Dubey, A.}, \bibinfo{author}{Jauhri, A.}, \bibinfo{author}{Pandey, A.}, \bibinfo{author}{Kadian, A.}, \bibinfo{author}{Al-Dahle, A.}, \bibinfo{author}{Letman, A.}, \bibinfo{author}{Mathur, A.}, \bibinfo{author}{Schelten, A.}, \bibinfo{author}{Yang, A.}, \bibinfo{author}{Fan, A.}, \bibinfo{author}{et~al.}, \bibinfo{year}{2024}.
\newblock \bibinfo{title}{The llama 3 herd of models}.
\newblock \URLprefix \url{https://arxiv.org/abs/2407.21783}, \href{http://arxiv.org/abs/2407.21783}{{\tt arXiv:2407.21783}}.
\bibitem[{Frantar and Alistarh(2023)}]{frantar2023sparsegptmassivelanguagemodels}
\bibinfo{author}{Frantar, E.}, \bibinfo{author}{Alistarh, D.}, \bibinfo{year}{2023}.
\newblock \bibinfo{title}{Sparsegpt: Massive language models can be accurately pruned in one-shot}.
\newblock \URLprefix \url{https://arxiv.org/abs/2301.00774}, \href{http://arxiv.org/abs/2301.00774}{{\tt arXiv:2301.00774}}.
\bibitem[{Frantar et~al.(2023)Frantar, Ashkboos, Hoefler and Alistarh}]{frantar2023_GPTQ}
\bibinfo{author}{Frantar, E.}, \bibinfo{author}{Ashkboos, S.}, \bibinfo{author}{Hoefler, T.}, \bibinfo{author}{Alistarh, D.}, \bibinfo{year}{2023}.
\newblock \bibinfo{title}{Gptq: Accurate post-training quantization for generative pre-trained transformers}.
\newblock \URLprefix \url{https://arxiv.org/abs/2210.17323}, \href{http://arxiv.org/abs/2210.17323}{{\tt arXiv:2210.17323}}.
\bibitem[{Fried et~al.(2023)Fried, Aghajanyan, Lin, Wang, Wallace, Shi, Zhong, Yih, Zettlemoyer and Lewis}]{fried2023_incoder}
\bibinfo{author}{Fried, D.}, \bibinfo{author}{Aghajanyan, A.}, \bibinfo{author}{Lin, J.}, \bibinfo{author}{Wang, S.}, \bibinfo{author}{Wallace, E.}, \bibinfo{author}{Shi, F.}, \bibinfo{author}{Zhong, R.}, \bibinfo{author}{Yih, S.}, \bibinfo{author}{Zettlemoyer, L.}, \bibinfo{author}{Lewis, M.}, \bibinfo{year}{2023}.
\newblock \bibinfo{title}{Incoder: A generative model for code infilling and synthesis}.
\newblock \bibinfo{journal}{The Eleventh International Conference on Learning Representations} \URLprefix \url{https://openreview.net/forum?id=hQwb-lbM6EL}.
\bibitem[{Gerganov(2023)}]{llamaCPP}
\bibinfo{author}{Gerganov, G.}, \bibinfo{year}{2023}.
\newblock \bibinfo{title}{llama.cpp}.
\newblock \URLprefix \url{https://github.com/ggerganov/llama.cpp}.
\bibitem[{Gerganov(2024)}]{Gerganov2023_GGUF}
\bibinfo{author}{Gerganov, G.}, \bibinfo{year}{2024}.
\newblock \bibinfo{title}{Gguf}.
\newblock \bibinfo{howpublished}{\url{https://huggingface.co/docs/hub/en/gguf}}.
\bibitem[{Gunasekar et~al.(2023)Gunasekar, Zhang, Aneja, Mendes, Giorno, Gopi, Javaheripi, Kauffmann and et~al.}]{gunasekar2023_phi1}
\bibinfo{author}{Gunasekar, S.}, \bibinfo{author}{Zhang, Y.}, \bibinfo{author}{Aneja, J.}, \bibinfo{author}{Mendes, C.C.T.}, \bibinfo{author}{Giorno, A.D.}, \bibinfo{author}{Gopi, S.}, \bibinfo{author}{Javaheripi, M.}, \bibinfo{author}{Kauffmann, P.}, \bibinfo{author}{et~al.}, \bibinfo{year}{2023}.
\newblock \bibinfo{title}{Textbooks are all you need}.
\newblock \href{http://arxiv.org/abs/2306.11644}{{\tt arXiv:2306.11644}}.
\bibitem[{Hartford(2023)}]{dolphin_dataset}
\bibinfo{author}{Hartford, E.}, \bibinfo{year}{2023}.
\newblock \bibinfo{title}{Dolphin dataset}.
\newblock \URLprefix \url{https://huggingface.co/datasets/cognitivecomputations/dolphin}.
\bibitem[{Heidari et~al.(2024)Heidari, Navimipour, Zeadally and Chamola}]{heidarieverything}
\bibinfo{author}{Heidari, A.}, \bibinfo{author}{Navimipour, N.J.}, \bibinfo{author}{Zeadally, S.}, \bibinfo{author}{Chamola, V.}, \bibinfo{year}{2024}.
\newblock \bibinfo{title}{Everything you wanted to know about chatgpt: Components, capabilities, applications, and opportunities}.
\newblock \bibinfo{journal}{Internet Technology Letters} , \bibinfo{pages}{e530}.
\bibitem[{Hendrycks et~al.(2021)Hendrycks, Burns, Basart, Zou, Mazeika, Song and Steinhardt}]{hendrycks2021_MMLU}
\bibinfo{author}{Hendrycks, D.}, \bibinfo{author}{Burns, C.}, \bibinfo{author}{Basart, S.}, \bibinfo{author}{Zou, A.}, \bibinfo{author}{Mazeika, M.}, \bibinfo{author}{Song, D.}, \bibinfo{author}{Steinhardt, J.}, \bibinfo{year}{2021}.
\newblock \bibinfo{title}{Measuring massive multitask language understanding}.
\newblock \href{http://arxiv.org/abs/2009.03300}{{\tt arXiv:2009.03300}}.
\bibitem[{{Hoffmann} et~al.(2022){Hoffmann}, {Borgeaud}, {Mensch}, {Buchatskaya}, {Cai}, {Rutherford}, {de Las Casas}, {Hendricks}, {Welbl}, {Clark} and et~al.}]{Hoffmann2022_chinchilla}
\bibinfo{author}{{Hoffmann}, J.}, \bibinfo{author}{{Borgeaud}, S.}, \bibinfo{author}{{Mensch}, A.}, \bibinfo{author}{{Buchatskaya}, E.}, \bibinfo{author}{{Cai}, T.}, \bibinfo{author}{{Rutherford}, E.}, \bibinfo{author}{{de Las Casas}, D.}, \bibinfo{author}{{Hendricks}, L.A.}, \bibinfo{author}{{Welbl}, J.}, \bibinfo{author}{{Clark}, A.}, \bibinfo{author}{et~al.}, \bibinfo{year}{2022}.
\newblock \bibinfo{title}{{Training Compute-Optimal Large Language Models}}.
\newblock \bibinfo{journal}{arXiv e-prints} .
\bibitem[{Huang et~al.(2024)Huang, Liu, Qin, Li, Zhang, Liu, Magno and Qi}]{huang2024billm}
\bibinfo{author}{Huang, W.}, \bibinfo{author}{Liu, Y.}, \bibinfo{author}{Qin, H.}, \bibinfo{author}{Li, Y.}, \bibinfo{author}{Zhang, S.}, \bibinfo{author}{Liu, X.}, \bibinfo{author}{Magno, M.}, \bibinfo{author}{Qi, X.}, \bibinfo{year}{2024}.
\newblock \bibinfo{title}{Billm: Pushing the limit of post-training quantization for llms}.
\newblock \bibinfo{journal}{arXiv preprint arXiv:2402.04291} .
\bibitem[{Inc. and TsinghuaNLP(2024)}]{minicpm2024}
\bibinfo{author}{Inc., M.}, \bibinfo{author}{TsinghuaNLP}, \bibinfo{year}{2024}.
\newblock \bibinfo{title}{Minicpm: Unveiling the potential of end-side large language models}.
\bibitem[{Jiang et~al.(2023)Jiang, Sablayrolles, Mensch, Bamford, Chaplot, de~las Casas, Bressand, Lengyel, Lample, Saulnier, Lavaud, Lachaux, Stock, Scao, Lavril, Wang, Lacroix and Sayed}]{jiang2023_mistral}
\bibinfo{author}{Jiang, A.Q.}, \bibinfo{author}{Sablayrolles, A.}, \bibinfo{author}{Mensch, A.}, \bibinfo{author}{Bamford, C.}, \bibinfo{author}{Chaplot, D.S.}, \bibinfo{author}{de~las Casas, D.}, \bibinfo{author}{Bressand, F.}, \bibinfo{author}{Lengyel, G.}, \bibinfo{author}{Lample, G.}, \bibinfo{author}{Saulnier, L.}, \bibinfo{author}{Lavaud, L.R.}, \bibinfo{author}{Lachaux, M.A.}, \bibinfo{author}{Stock, P.}, \bibinfo{author}{Scao, T.L.}, \bibinfo{author}{Lavril, T.}, \bibinfo{author}{Wang, T.}, \bibinfo{author}{Lacroix, T.}, \bibinfo{author}{Sayed, W.E.}, \bibinfo{year}{2023}.
\newblock \bibinfo{title}{Mistral 7b}.
\newblock \href{http://arxiv.org/abs/2310.06825}{{\tt arXiv:2310.06825}}.
\bibitem[{Jiang et~al.(2024)Jiang, Sablayrolles, Roux, Mensch, Savary, Bamford, Chaplot, de~las Casas, Hanna, Bressand and et~al.}]{jiang2024_mixtral}
\bibinfo{author}{Jiang, A.Q.}, \bibinfo{author}{Sablayrolles, A.}, \bibinfo{author}{Roux, A.}, \bibinfo{author}{Mensch, A.}, \bibinfo{author}{Savary, B.}, \bibinfo{author}{Bamford, C.}, \bibinfo{author}{Chaplot, D.S.}, \bibinfo{author}{de~las Casas, D.}, \bibinfo{author}{Hanna, E.B.}, \bibinfo{author}{Bressand, F.}, \bibinfo{author}{et~al.}, \bibinfo{year}{2024}.
\newblock \bibinfo{title}{Mixtral of experts}.
\newblock \href{http://arxiv.org/abs/2401.04088}{{\tt arXiv:2401.04088}}.
\bibitem[{Kim et~al.(2024)Kim, Hooper, Gholami, Dong, Li, Shen, Mahoney and Keutzer}]{kim2024squeezellmdenseandsparsequantization}
\bibinfo{author}{Kim, S.}, \bibinfo{author}{Hooper, C.}, \bibinfo{author}{Gholami, A.}, \bibinfo{author}{Dong, Z.}, \bibinfo{author}{Li, X.}, \bibinfo{author}{Shen, S.}, \bibinfo{author}{Mahoney, M.W.}, \bibinfo{author}{Keutzer, K.}, \bibinfo{year}{2024}.
\newblock \bibinfo{title}{Squeezellm: Dense-and-sparse quantization}.
\newblock \URLprefix \url{https://arxiv.org/abs/2306.07629}, \href{http://arxiv.org/abs/2306.07629}{{\tt arXiv:2306.07629}}.
\bibitem[{Kulal et~al.(2019)Kulal, Pasupat, Chandra, Lee, Padon, Aiken and Liang}]{Kulal2019_passK}
\bibinfo{author}{Kulal, S.}, \bibinfo{author}{Pasupat, P.}, \bibinfo{author}{Chandra, K.}, \bibinfo{author}{Lee, M.}, \bibinfo{author}{Padon, O.}, \bibinfo{author}{Aiken, A.}, \bibinfo{author}{Liang, P.}, \bibinfo{year}{2019}.
\newblock \bibinfo{title}{Spoc: Search-based pseudocode to code}.
\newblock \URLprefix \url{https://arxiv.org/abs/1906.04908}, \href{http://arxiv.org/abs/1906.04908}{{\tt arXiv:1906.04908}}.
\bibitem[{Lancaster(2023)}]{lancaster2023artificial}
\bibinfo{author}{Lancaster, T.}, \bibinfo{year}{2023}.
\newblock \bibinfo{title}{Artificial intelligence, text generation tools and chatgpt--does digital watermarking offer a solution?}
\newblock \bibinfo{journal}{International Journal for Educational Integrity} \bibinfo{volume}{19}, \bibinfo{pages}{10}.
\bibitem[{Leswing(2024)}]{Leswing2024_paramsGPT4}
\bibinfo{author}{Leswing, K.}, \bibinfo{year}{2024}.
\newblock \bibinfo{title}{Nvidia ceo jensen huang announces new ai chips: ‘we need bigger gpus’}.
\newblock \URLprefix \url{https://www.cnbc.com/2024/03/18/nvidia-announces-gb200-blackwell-ai-chip-launching-later-this-year.html}.
\bibitem[{Li et~al.(2023a)Li, allal, Zi, Muennighoff, Kocetkov, Mou, Marone, Akiki, LI, Chim and et~al.}]{li2023_starcoder}
\bibinfo{author}{Li, R.}, \bibinfo{author}{allal, L.B.}, \bibinfo{author}{Zi, Y.}, \bibinfo{author}{Muennighoff, N.}, \bibinfo{author}{Kocetkov, D.}, \bibinfo{author}{Mou, C.}, \bibinfo{author}{Marone, M.}, \bibinfo{author}{Akiki, C.}, \bibinfo{author}{LI, J.}, \bibinfo{author}{Chim, J.}, \bibinfo{author}{et~al.}, \bibinfo{year}{2023}a.
\newblock \bibinfo{title}{Starcoder: may the source be with you!}
\newblock \bibinfo{journal}{Transactions on Machine Learning Research} \URLprefix \url{https://openreview.net/forum?id=KoFOg41haE}. \bibinfo{note}{reproducibility Certification}.
\bibitem[{Li et~al.(2023b)Li, Ning, Hong, Liu, Wang, Li, Zhong, Dai, Yang and Wang}]{li2023llm}
\bibinfo{author}{Li, S.}, \bibinfo{author}{Ning, X.}, \bibinfo{author}{Hong, K.}, \bibinfo{author}{Liu, T.}, \bibinfo{author}{Wang, L.}, \bibinfo{author}{Li, X.}, \bibinfo{author}{Zhong, K.}, \bibinfo{author}{Dai, G.}, \bibinfo{author}{Yang, H.}, \bibinfo{author}{Wang, Y.}, \bibinfo{year}{2023}b.
\newblock \bibinfo{title}{Llm-mq: Mixed-precision quantization for efficient llm deployment}.
\newblock \bibinfo{journal}{The Efficient Natural Language and Speech Processing Workshop with NeurIPS} \bibinfo{volume}{9}.
\bibitem[{Li et~al.(2023c)Li, Bubeck, Eldan, Giorno, Gunasekar and Lee}]{li2023_phi1.5}
\bibinfo{author}{Li, Y.}, \bibinfo{author}{Bubeck, S.}, \bibinfo{author}{Eldan, R.}, \bibinfo{author}{Giorno, A.D.}, \bibinfo{author}{Gunasekar, S.}, \bibinfo{author}{Lee, Y.T.}, \bibinfo{year}{2023}c.
\newblock \bibinfo{title}{Textbooks are all you need ii: phi-1.5 technical report}.
\newblock \href{http://arxiv.org/abs/2309.05463}{{\tt arXiv:2309.05463}}.
\bibitem[{Li et~al.(2022)Li, Choi, Chung, Kushman, Schrittwieser, Leblond, Eccles, Keeling, Gimeno, Dal~Lago and et~al.}]{Li2022_AlphaCode}
\bibinfo{author}{Li, Y.}, \bibinfo{author}{Choi, D.}, \bibinfo{author}{Chung, J.}, \bibinfo{author}{Kushman, N.}, \bibinfo{author}{Schrittwieser, J.}, \bibinfo{author}{Leblond, R.}, \bibinfo{author}{Eccles, T.}, \bibinfo{author}{Keeling, J.}, \bibinfo{author}{Gimeno, F.}, \bibinfo{author}{Dal~Lago, A.}, \bibinfo{author}{et~al.}, \bibinfo{year}{2022}.
\newblock \bibinfo{title}{Competition-level code generation with alphacode}.
\newblock \bibinfo{journal}{Science} \bibinfo{volume}{378}, \bibinfo{pages}{1092–1097}.
\newblock \URLprefix \url{http://dx.doi.org/10.1126/science.abq1158}, \DOIprefix\doi{10.1126/science.abq1158}.
\bibitem[{Lin et~al.(2023)Lin, Tang, Tang, Yang, Dang, Gan and Han}]{lin2023awq}
\bibinfo{author}{Lin, J.}, \bibinfo{author}{Tang, J.}, \bibinfo{author}{Tang, H.}, \bibinfo{author}{Yang, S.}, \bibinfo{author}{Dang, X.}, \bibinfo{author}{Gan, C.}, \bibinfo{author}{Han, S.}, \bibinfo{year}{2023}.
\newblock \bibinfo{title}{Awq: Activation-aware weight quantization for llm compression and acceleration}.
\newblock \href{http://arxiv.org/abs/2306.00978}{{\tt arXiv:2306.00978}}.
\bibitem[{Lin et~al.(2021)Lin, Hilton and Evans}]{lin2021truthfulqa}
\bibinfo{author}{Lin, S.}, \bibinfo{author}{Hilton, J.}, \bibinfo{author}{Evans, O.}, \bibinfo{year}{2021}.
\newblock \bibinfo{title}{Truthfulqa: Measuring how models mimic human falsehoods}.
\newblock \href{http://arxiv.org/abs/2109.07958}{{\tt arXiv:2109.07958}}.
\bibitem[{Liu et~al.(2023a)Liu, Bao, Zhang, Zhang, Hu, Zhang and Yan}]{liu2023improving}
\bibinfo{author}{Liu, C.}, \bibinfo{author}{Bao, X.}, \bibinfo{author}{Zhang, H.}, \bibinfo{author}{Zhang, N.}, \bibinfo{author}{Hu, H.}, \bibinfo{author}{Zhang, X.}, \bibinfo{author}{Yan, M.}, \bibinfo{year}{2023}a.
\newblock \bibinfo{title}{Improving chatgpt prompt for code generation}.
\newblock \href{http://arxiv.org/abs/2305.08360}{{\tt arXiv:2305.08360}}.
\bibitem[{Liu et~al.(2023b)Liu, Xia, Wang and ZHANG}]{liu2023_evalGPTcode}
\bibinfo{author}{Liu, J.}, \bibinfo{author}{Xia, C.S.}, \bibinfo{author}{Wang, Y.}, \bibinfo{author}{ZHANG, L.}, \bibinfo{year}{2023}b.
\newblock \bibinfo{title}{Is your code generated by chat{GPT} really correct? rigorous evaluation of large language models for code generation}.
\newblock \bibinfo{journal}{Thirty-seventh Conference on Neural Information Processing Systems} \URLprefix \url{https://openreview.net/forum?id=1qvx610Cu7}.
\bibitem[{Liu et~al.(2023c)Liu, Xia, Wang and ZHANG}]{Liu2023_ChatGPTCode}
\bibinfo{author}{Liu, J.}, \bibinfo{author}{Xia, C.S.}, \bibinfo{author}{Wang, Y.}, \bibinfo{author}{ZHANG, L.}, \bibinfo{year}{2023}c.
\newblock \bibinfo{title}{Is your code generated by chatgpt really correct? rigorous evaluation of large language models for code generation}, in: \bibinfo{editor}{Oh, A.}, \bibinfo{editor}{Neumann, T.}, \bibinfo{editor}{Globerson, A.}, \bibinfo{editor}{Saenko, K.}, \bibinfo{editor}{Hardt, M.}, \bibinfo{editor}{Levine, S.} (Eds.), \bibinfo{booktitle}{Advances in Neural Information Processing Systems}, \bibinfo{publisher}{Curran Associates, Inc.}. pp. \bibinfo{pages}{21558--21572}.
\newblock \URLprefix \url{https://proceedings.neurips.cc/paper_files/paper/2023/file/43e9d647ccd3e4b7b5baab53f0368686-Paper-Conference.pdf}.
\bibitem[{Liu et~al.(2023d)Liu, Oguz, Zhao, Chang, Stock, Mehdad, Shi, Krishnamoorthi and Chandra}]{liu2023llm}
\bibinfo{author}{Liu, Z.}, \bibinfo{author}{Oguz, B.}, \bibinfo{author}{Zhao, C.}, \bibinfo{author}{Chang, E.}, \bibinfo{author}{Stock, P.}, \bibinfo{author}{Mehdad, Y.}, \bibinfo{author}{Shi, Y.}, \bibinfo{author}{Krishnamoorthi, R.}, \bibinfo{author}{Chandra, V.}, \bibinfo{year}{2023}d.
\newblock \bibinfo{title}{Llm-qat: Data-free quantization aware training for large language models}.
\newblock \bibinfo{journal}{arXiv preprint arXiv:2305.17888} .
\bibitem[{{López Espejel} et~al.(2023a){López Espejel}, Ettifouri, {Yahaya Alassan}, Chouham and Dahhane}]{lopez2023_reasoningGPT_bard}
\bibinfo{author}{{López Espejel}, J.}, \bibinfo{author}{Ettifouri, E.H.}, \bibinfo{author}{{Yahaya Alassan}, M.S.}, \bibinfo{author}{Chouham, E.M.}, \bibinfo{author}{Dahhane, W.}, \bibinfo{year}{2023}a.
\newblock \bibinfo{title}{Gpt-3.5, gpt-4, or bard? evaluating llms reasoning ability in zero-shot setting and performance boosting through prompts}.
\newblock \bibinfo{journal}{Natural Language Processing Journal} \bibinfo{volume}{5}, \bibinfo{pages}{100032}.
\newblock \URLprefix \url{https://www.sciencedirect.com/science/article/pii/S2949719123000298}, \DOIprefix\doi{https://doi.org/10.1016/j.nlp.2023.100032}.
\bibitem[{{López Espejel} et~al.(2023b){López Espejel}, {Yahaya Alassan}, Chouham, Dahhane and Ettifouri}]{Lopez2023_comprehensive}
\bibinfo{author}{{López Espejel}, J.}, \bibinfo{author}{{Yahaya Alassan}, M.S.}, \bibinfo{author}{Chouham, E.M.}, \bibinfo{author}{Dahhane, W.}, \bibinfo{author}{Ettifouri, E.H.}, \bibinfo{year}{2023}b.
\newblock \bibinfo{title}{A comprehensive review of state-of-the-art methods for java code generation from natural language text}.
\newblock \bibinfo{journal}{Natural Language Processing Journal} \bibinfo{volume}{3}, \bibinfo{pages}{100013}.
\newblock \URLprefix \url{https://www.sciencedirect.com/science/article/pii/S2949719123000109}, \DOIprefix\doi{https://doi.org/10.1016/j.nlp.2023.100013}.
\bibitem[{Ma et~al.(2023a)Ma, Fang and Wang}]{ma2023llm}
\bibinfo{author}{Ma, X.}, \bibinfo{author}{Fang, G.}, \bibinfo{author}{Wang, X.}, \bibinfo{year}{2023}a.
\newblock \bibinfo{title}{Llm-pruner: On the structural pruning of large language models}.
\newblock \bibinfo{journal}{Advances in neural information processing systems} \bibinfo{volume}{36}, \bibinfo{pages}{21702--21720}.
\bibitem[{Ma et~al.(2023b)Ma, Cao, Sun, Pavone and Xiao}]{ma2023_dolphins}
\bibinfo{author}{Ma, Y.}, \bibinfo{author}{Cao, Y.}, \bibinfo{author}{Sun, J.}, \bibinfo{author}{Pavone, M.}, \bibinfo{author}{Xiao, C.}, \bibinfo{year}{2023}b.
\newblock \bibinfo{title}{Dolphins: Multimodal language model for driving}.
\newblock \href{http://arxiv.org/abs/2312.00438}{{\tt arXiv:2312.00438}}.
\bibitem[{Manyika(2023)}]{manyika2023_BARD}
\bibinfo{author}{Manyika, J.}, \bibinfo{year}{2023}.
\newblock \bibinfo{title}{An overview of bard: an early experiment with generative ai}.
\bibitem[{Mukherjee et~al.(2023)Mukherjee, Mitra, Jawahar, Agarwal, Palangi and Awadallah}]{mukherjee2023_orca}
\bibinfo{author}{Mukherjee, S.}, \bibinfo{author}{Mitra, A.}, \bibinfo{author}{Jawahar, G.}, \bibinfo{author}{Agarwal, S.}, \bibinfo{author}{Palangi, H.}, \bibinfo{author}{Awadallah, A.}, \bibinfo{year}{2023}.
\newblock \bibinfo{title}{Orca: Progressive learning from complex explanation traces of gpt-4}.
\newblock \href{http://arxiv.org/abs/2306.02707}{{\tt arXiv:2306.02707}}.
\bibitem[{Mulia et~al.(2023)Mulia, Piri and Tho}]{Mulia2023_text-generation-chatgpt}
\bibinfo{author}{Mulia, A.P.}, \bibinfo{author}{Piri, P.R.}, \bibinfo{author}{Tho, C.}, \bibinfo{year}{2023}.
\newblock \bibinfo{title}{Usability analysis of text generation by chatgpt openai using system usability scale method}.
\newblock \bibinfo{journal}{Procedia Computer Science} \bibinfo{volume}{227}, \bibinfo{pages}{381--388}.
\newblock \URLprefix \url{https://www.sciencedirect.com/science/article/pii/S1877050923017040}, \DOIprefix\doi{https://doi.org/10.1016/j.procs.2023.10.537}. \bibinfo{note}{8th International Conference on Computer Science and Computational Intelligence (ICCSCI 2023)}.
\bibitem[{Omar et~al.(2023)Omar, Mangukiya, Kalnis and Mansour}]{omar2023chatgpt}
\bibinfo{author}{Omar, R.}, \bibinfo{author}{Mangukiya, O.}, \bibinfo{author}{Kalnis, P.}, \bibinfo{author}{Mansour, E.}, \bibinfo{year}{2023}.
\newblock \bibinfo{title}{Chatgpt versus traditional question answering for knowledge graphs: Current status and future directions towards knowledge graph chatbots}.
\newblock \bibinfo{journal}{arXiv preprint arXiv:2302.06466} .
\bibitem[{OpenAI(2022)}]{chatGPT3.5}
\bibinfo{author}{OpenAI}, \bibinfo{year}{2022}.
\newblock \bibinfo{title}{Introducing chatgpt}.
\newblock \bibinfo{howpublished}{\url{https://openai.com/blog/chatgpt}}.
\bibitem[{OpenAI(2023)}]{chatGPT4}
\bibinfo{author}{OpenAI}, \bibinfo{year}{2023}.
\newblock \bibinfo{title}{Gpt-4 technical report}.
\newblock \bibinfo{journal}{arxiv} \URLprefix \url{https://arxiv.org/pdf/2303.08774.pdf}.
\bibitem[{Ouyang et~al.(2022)Ouyang, Wu, Jiang, Almeida, Wainwright, Mishkin, Zhang, Agarwal, Slama, Ray, Schulman, Hilton, Kelton, Miller, Simens, Askell, Welinder, Christiano, Leike and Lowe}]{Ouyang2022TrainingLM}
\bibinfo{author}{Ouyang, L.}, \bibinfo{author}{Wu, J.}, \bibinfo{author}{Jiang, X.}, \bibinfo{author}{Almeida, D.}, \bibinfo{author}{Wainwright, C.L.}, \bibinfo{author}{Mishkin, P.}, \bibinfo{author}{Zhang, C.}, \bibinfo{author}{Agarwal, S.}, \bibinfo{author}{Slama, K.}, \bibinfo{author}{Ray, A.}, \bibinfo{author}{Schulman, J.}, \bibinfo{author}{Hilton, J.}, \bibinfo{author}{Kelton, F.}, \bibinfo{author}{Miller, L.E.}, \bibinfo{author}{Simens, M.}, \bibinfo{author}{Askell, A.}, \bibinfo{author}{Welinder, P.}, \bibinfo{author}{Christiano, P.F.}, \bibinfo{author}{Leike, J.}, \bibinfo{author}{Lowe, R.J.}, \bibinfo{year}{2022}.
\newblock \bibinfo{title}{Training language models to follow instructions with human feedback}.
\newblock \bibinfo{journal}{ArXiv} \bibinfo{volume}{abs/2203.02155}.
\bibitem[{Park et~al.(2024)Park, Hyun, Cho, Sim and Lee}]{park2024anyprecisionllmlowcostdeployment}
\bibinfo{author}{Park, Y.}, \bibinfo{author}{Hyun, J.}, \bibinfo{author}{Cho, S.}, \bibinfo{author}{Sim, B.}, \bibinfo{author}{Lee, J.W.}, \bibinfo{year}{2024}.
\newblock \bibinfo{title}{Any-precision llm: Low-cost deployment of multiple, different-sized llms}.
\newblock \URLprefix \url{https://arxiv.org/abs/2402.10517}, \href{http://arxiv.org/abs/2402.10517}{{\tt arXiv:2402.10517}}.
\bibitem[{Plevris et~al.(2023)Plevris, Papazafeiropoulos and Jim{\'e}nez~Rios}]{plevris2023_chats_math}
\bibinfo{author}{Plevris, V.}, \bibinfo{author}{Papazafeiropoulos, G.}, \bibinfo{author}{Jim{\'e}nez~Rios, A.}, \bibinfo{year}{2023}.
\newblock \bibinfo{title}{Chatbots put to the test in math and logic problems: A comparison and assessment of chatgpt-3.5, chatgpt-4, and google bard}.
\newblock \bibinfo{journal}{AI} \bibinfo{volume}{4}, \bibinfo{pages}{949--969}.
\bibitem[{Radford et~al.(2018a)Radford, Narasimhan, Salimans and Sutskever}]{radford2018improving}
\bibinfo{author}{Radford, A.}, \bibinfo{author}{Narasimhan, K.}, \bibinfo{author}{Salimans, T.}, \bibinfo{author}{Sutskever, I.}, \bibinfo{year}{2018}a.
\newblock \bibinfo{title}{Improving language understanding by generative pre-training}.
\newblock \bibinfo{journal}{arxiv} .
\bibitem[{Radford et~al.(2018b)Radford, Wu, Child, Luan, Amodei and Sutskever}]{Radford2018_LLMMultitask}
\bibinfo{author}{Radford, A.}, \bibinfo{author}{Wu, J.}, \bibinfo{author}{Child, R.}, \bibinfo{author}{Luan, D.}, \bibinfo{author}{Amodei, D.}, \bibinfo{author}{Sutskever, I.}, \bibinfo{year}{2018}b.
\newblock \bibinfo{title}{Language models are unsupervised multitask learners}.
\newblock \bibinfo{journal}{arxiv} \URLprefix \url{https://d4mucfpksywv.cloudfront.net/better-language-models/language-models.pdf}.
\bibitem[{{Rae} et~al.(2021){Rae}, {Borgeaud}, {Cai}, {Millican}, {Hoffmann}, {Song}, {Aslanides}, {Henderson}, {Ring}, {Young} and et~al.}]{Jack2021_Gopher}
\bibinfo{author}{{Rae}, J.W.}, \bibinfo{author}{{Borgeaud}, S.}, \bibinfo{author}{{Cai}, T.}, \bibinfo{author}{{Millican}, K.}, \bibinfo{author}{{Hoffmann}, J.}, \bibinfo{author}{{Song}, F.}, \bibinfo{author}{{Aslanides}, J.}, \bibinfo{author}{{Henderson}, S.}, \bibinfo{author}{{Ring}, R.}, \bibinfo{author}{{Young}, S.}, \bibinfo{author}{et~al.}, \bibinfo{year}{2021}.
\newblock \bibinfo{title}{{Scaling Language Models: Methods, Analysis \& Insights from Training Gopher}}.
\newblock \bibinfo{journal}{arXiv e-prints} , \bibinfo{pages}{arXiv:2112.11446}\DOIprefix\doi{10.48550/arXiv.2112.11446}, \href{http://arxiv.org/abs/2112.11446}{{\tt arXiv:2112.11446}}.
\bibitem[{Rafailov et~al.(2023)Rafailov, Sharma, Mitchell, Ermon, Manning and Finn}]{rafailov2023_DPO}
\bibinfo{author}{Rafailov, R.}, \bibinfo{author}{Sharma, A.}, \bibinfo{author}{Mitchell, E.}, \bibinfo{author}{Ermon, S.}, \bibinfo{author}{Manning, C.D.}, \bibinfo{author}{Finn, C.}, \bibinfo{year}{2023}.
\newblock \bibinfo{title}{Direct preference optimization: Your language model is secretly a reward model}.
\newblock \href{http://arxiv.org/abs/2305.18290}{{\tt arXiv:2305.18290}}.
\bibitem[{Ramachandran et~al.(2017)Ramachandran, Zoph and Le}]{Ramachandran2017_SwishAS}
\bibinfo{author}{Ramachandran, P.}, \bibinfo{author}{Zoph, B.}, \bibinfo{author}{Le, Q.V.}, \bibinfo{year}{2017}.
\newblock \bibinfo{title}{Swish: a self-gated activation function}.
\newblock \bibinfo{journal}{arXiv: Neural and Evolutionary Computing} \URLprefix \url{https://api.semanticscholar.org/CorpusID:196158220}.
\bibitem[{Ramírez et~al.(2024)Ramírez, Espejel, del Carmen Santiago~Díaz and Linares}]{ramírez2024soloescuchamespanishemotional}
\bibinfo{author}{Ramírez, B.G.}, \bibinfo{author}{Espejel, J.L.}, \bibinfo{author}{del Carmen Santiago~Díaz, M.}, \bibinfo{author}{Linares, G.T.R.}, \bibinfo{year}{2024}.
\newblock \bibinfo{title}{S\'olo esc\'uchame: Spanish emotional accompaniment chatbot}.
\newblock \URLprefix \url{https://arxiv.org/abs/2408.01852}, \href{http://arxiv.org/abs/2408.01852}{{\tt arXiv:2408.01852}}.
\bibitem[{Rozière et~al.(2024)Rozière, Gehring, Gloeckle, Sootla, Gat, Tan, Adi, Liu and et~al.}]{roziere2024_codeLLaMA}
\bibinfo{author}{Rozière, B.}, \bibinfo{author}{Gehring, J.}, \bibinfo{author}{Gloeckle, F.}, \bibinfo{author}{Sootla, S.}, \bibinfo{author}{Gat, I.}, \bibinfo{author}{Tan, X.E.}, \bibinfo{author}{Adi, Y.}, \bibinfo{author}{Liu, J.}, \bibinfo{author}{et~al.}, \bibinfo{year}{2024}.
\newblock \bibinfo{title}{Code llama: Open foundation models for code}.
\newblock \href{http://arxiv.org/abs/2308.12950}{{\tt arXiv:2308.12950}}.
\bibitem[{Scholak et~al.(2021)Scholak, Schucher and Bahdanau}]{Scholak2021_PICARD}
\bibinfo{author}{Scholak, T.}, \bibinfo{author}{Schucher, N.}, \bibinfo{author}{Bahdanau, D.}, \bibinfo{year}{2021}.
\newblock \bibinfo{title}{{PICARD}: Parsing incrementally for constrained auto-regressive decoding from language models}, in: \bibinfo{booktitle}{Proceedings of the 2021 Conference on Empirical Methods in Natural Language Processing}, \bibinfo{publisher}{Association for Computational Linguistics}. pp. \bibinfo{pages}{9895--9901}.
\newblock \URLprefix \url{https://aclanthology.org/2021.emnlp-main.779}.
\bibitem[{Shazeer(2020)}]{shazeer2020_glu}
\bibinfo{author}{Shazeer, N.}, \bibinfo{year}{2020}.
\newblock \bibinfo{title}{Glu variants improve transformer}.
\newblock \href{http://arxiv.org/abs/2002.05202}{{\tt arXiv:2002.05202}}.
\bibitem[{Smith et~al.(2022)Smith, Patwary, Norick, LeGresley, Rajbhandari, Casper, Liu, Prabhumoye, Zerveas, Korthikanti and et~al.}]{smith2022_megatron}
\bibinfo{author}{Smith, S.}, \bibinfo{author}{Patwary, M.}, \bibinfo{author}{Norick, B.}, \bibinfo{author}{LeGresley, P.}, \bibinfo{author}{Rajbhandari, S.}, \bibinfo{author}{Casper, J.}, \bibinfo{author}{Liu, Z.}, \bibinfo{author}{Prabhumoye, S.}, \bibinfo{author}{Zerveas, G.}, \bibinfo{author}{Korthikanti, V.}, \bibinfo{author}{et~al.}, \bibinfo{year}{2022}.
\newblock \bibinfo{title}{Using deepspeed and megatron to train megatron-turing nlg 530b, a large-scale generative language model}.
\newblock \href{http://arxiv.org/abs/2201.11990}{{\tt arXiv:2201.11990}}.
\bibitem[{Taori et~al.(2023)Taori, Gulrajani, Zhang, Dubois, Li, Guestrin, Liang and Hashimoto}]{Rohan2023_alpaca}
\bibinfo{author}{Taori, R.}, \bibinfo{author}{Gulrajani, I.}, \bibinfo{author}{Zhang, T.}, \bibinfo{author}{Dubois, Y.}, \bibinfo{author}{Li, X.}, \bibinfo{author}{Guestrin, C.}, \bibinfo{author}{Liang, P.}, \bibinfo{author}{Hashimoto, T.B.}, \bibinfo{year}{2023}.
\newblock \bibinfo{title}{Stanford alpaca: An instruction-following llama model}.
\newblock \bibinfo{howpublished}{\url{https://github.com/tatsu-lab/stanford_alpaca}}.
\bibitem[{Team et~al.(2023)Team, Anil, Borgeaud, Wu, Alayrac, Yu, Soricut, Schalkwyk, Dai, Hauth and et~al.}]{geminiteam2023_gemini}
\bibinfo{author}{Team, G.}, \bibinfo{author}{Anil, R.}, \bibinfo{author}{Borgeaud, S.}, \bibinfo{author}{Wu, Y.}, \bibinfo{author}{Alayrac, J.B.}, \bibinfo{author}{Yu, J.}, \bibinfo{author}{Soricut, R.}, \bibinfo{author}{Schalkwyk, J.}, \bibinfo{author}{Dai, A.M.}, \bibinfo{author}{Hauth, A.}, \bibinfo{author}{et~al., K.M.}, \bibinfo{year}{2023}.
\newblock \bibinfo{title}{Gemini: A family of highly capable multimodal models}.
\newblock \href{http://arxiv.org/abs/2312.11805}{{\tt arXiv:2312.11805}}.
\bibitem[{Team et~al.(2024)Team, Mesnard, Hardin, Dadashi, Bhupatiraju, Pathak, Sifre, Rivière, Kale, Love and et~al.}]{gemmateam2024_gemma}
\bibinfo{author}{Team, G.}, \bibinfo{author}{Mesnard, T.}, \bibinfo{author}{Hardin, C.}, \bibinfo{author}{Dadashi, R.}, \bibinfo{author}{Bhupatiraju, S.}, \bibinfo{author}{Pathak, S.}, \bibinfo{author}{Sifre, L.}, \bibinfo{author}{Rivière, M.}, \bibinfo{author}{Kale, M.S.}, \bibinfo{author}{Love, J.}, \bibinfo{author}{et~al.}, \bibinfo{year}{2024}.
\newblock \bibinfo{title}{Gemma: Open models based on gemini research and technology}.
\newblock \href{http://arxiv.org/abs/2403.08295}{{\tt arXiv:2403.08295}}.
\bibitem[{Team(2024a)}]{evalplus_scores}
\bibinfo{author}{Team, H.}, \bibinfo{year}{2024}a.
\newblock \bibinfo{title}{Evalplus leaderboard}.
\newblock \bibinfo{howpublished}{\url{https://evalplus.github.io/leaderboard.html}}.
\bibitem[{Team(2024b)}]{Magicoder-OSS_75k}
\bibinfo{author}{Team, H.}, \bibinfo{year}{2024}b.
\newblock \bibinfo{title}{Magicoder-oss-instruct-75k}.
\newblock \bibinfo{howpublished}{\url{https://huggingface.co/datasets/ise-uiuc/Magicoder-OSS-Instruct-75K}}.
\bibitem[{Teknium(2023)}]{Teknium2023_openhermes}
\bibinfo{author}{Teknium}, \bibinfo{year}{2023}.
\newblock \bibinfo{title}{Openhermes 2.5 mistral 7b - gguf}.
\newblock \URLprefix \url{https://huggingface.co/teknium/OpenHermes-2.5-Mistral-7B}.
\bibitem[{Thoppilan et~al.(2022)Thoppilan, Freitas, Hall, Shazeer, Kulshreshtha, Cheng, Jin, Bos, Baker, Du, Li, Lee and et~al.}]{Thoppilan2022_LaMDA}
\bibinfo{author}{Thoppilan, R.}, \bibinfo{author}{Freitas, D.}, \bibinfo{author}{Hall, J.}, \bibinfo{author}{Shazeer, N.}, \bibinfo{author}{Kulshreshtha, A.}, \bibinfo{author}{Cheng, H.T.}, \bibinfo{author}{Jin, A.}, \bibinfo{author}{Bos, T.}, \bibinfo{author}{Baker, L.}, \bibinfo{author}{Du, Y.}, \bibinfo{author}{Li, Y.}, \bibinfo{author}{Lee, H.}, \bibinfo{author}{et~al.}, \bibinfo{year}{2022}.
\newblock \bibinfo{title}{Lamda: Language models for dialog applications}.
\newblock \bibinfo{journal}{arXiv} .
\bibitem[{Touvron et~al.(2023a)Touvron, Lavril, Izacard, Martinet, Lachaux, Lacroix, Rozière, Goyal, Hambro, Azhar, Rodriguez, Joulin, Grave and Lample}]{touvron2023_llama1}
\bibinfo{author}{Touvron, H.}, \bibinfo{author}{Lavril, T.}, \bibinfo{author}{Izacard, G.}, \bibinfo{author}{Martinet, X.}, \bibinfo{author}{Lachaux, M.A.}, \bibinfo{author}{Lacroix, T.}, \bibinfo{author}{Rozière, B.}, \bibinfo{author}{Goyal, N.}, \bibinfo{author}{Hambro, E.}, \bibinfo{author}{Azhar, F.}, \bibinfo{author}{Rodriguez, A.}, \bibinfo{author}{Joulin, A.}, \bibinfo{author}{Grave, E.}, \bibinfo{author}{Lample, G.}, \bibinfo{year}{2023}a.
\newblock \bibinfo{title}{Llama: Open and efficient foundation language models}.
\newblock \href{http://arxiv.org/abs/2302.13971}{{\tt arXiv:2302.13971}}.
\bibitem[{Touvron et~al.(2023b)Touvron, Martin, Stone, Albert, Almahairi, Babaei, Bashlykov, Batra, Bhargava, Bhosale and et~al.}]{touvron2023_llama2}
\bibinfo{author}{Touvron, H.}, \bibinfo{author}{Martin, L.}, \bibinfo{author}{Stone, K.}, \bibinfo{author}{Albert, P.}, \bibinfo{author}{Almahairi, A.}, \bibinfo{author}{Babaei, Y.}, \bibinfo{author}{Bashlykov, N.}, \bibinfo{author}{Batra, S.}, \bibinfo{author}{Bhargava, P.}, \bibinfo{author}{Bhosale, S.}, \bibinfo{author}{et~al.}, \bibinfo{year}{2023}b.
\newblock \bibinfo{title}{Llama 2: Open foundation and fine-tuned chat models}.
\newblock \href{http://arxiv.org/abs/2307.09288}{{\tt arXiv:2307.09288}}.
\bibitem[{Tunstall et~al.(2023)Tunstall, Beeching, Lambert, Rajani, Rasul, Belkada, Huang, von Werra, Fourrier, Habib and et~al.}]{tunstall2023_zephyr}
\bibinfo{author}{Tunstall, L.}, \bibinfo{author}{Beeching, E.}, \bibinfo{author}{Lambert, N.}, \bibinfo{author}{Rajani, N.}, \bibinfo{author}{Rasul, K.}, \bibinfo{author}{Belkada, Y.}, \bibinfo{author}{Huang, S.}, \bibinfo{author}{von Werra, L.}, \bibinfo{author}{Fourrier, C.}, \bibinfo{author}{Habib, N.}, \bibinfo{author}{et~al.}, \bibinfo{year}{2023}.
\newblock \bibinfo{title}{Zephyr: Direct distillation of lm alignment}.
\newblock \href{http://arxiv.org/abs/2310.16944}{{\tt arXiv:2310.16944}}.
\bibitem[{Vailshery(2024)}]{Vailshery2024_pl}
\bibinfo{author}{Vailshery, L.S.}, \bibinfo{year}{2024}.
\newblock \bibinfo{title}{Most used programming languages among developers worldwide as of 2023}.
\newblock \URLprefix \url{https://www.statista.com/statistics/793628/worldwide-developer-survey-most-used-languages/}.
\bibitem[{Vaswani et~al.(2017)Vaswani, Shazeer, Parmar, Uszkoreit, Jones, Gomez, Kaiser and Polosukhin}]{Vaswani2017_transformers}
\bibinfo{author}{Vaswani, A.}, \bibinfo{author}{Shazeer, N.}, \bibinfo{author}{Parmar, N.}, \bibinfo{author}{Uszkoreit, J.}, \bibinfo{author}{Jones, L.}, \bibinfo{author}{Gomez, A.N.}, \bibinfo{author}{Kaiser, L.u.}, \bibinfo{author}{Polosukhin, I.}, \bibinfo{year}{2017}.
\newblock \bibinfo{title}{Attention is all you need}.
\newblock \bibinfo{journal}{Advances in Neural Information Processing Systems} \bibinfo{volume}{30}.
\bibitem[{Wang and Komatsuzaki(2021)}]{Wang2021_gptj}
\bibinfo{author}{Wang, B.}, \bibinfo{author}{Komatsuzaki, A.}, \bibinfo{year}{2021}.
\newblock \bibinfo{title}{Gpt-j-6b: A 6 billion parameter autoregressive language model}.
\newblock \bibinfo{howpublished}{\url{https://github.com/kingoflolz/mesh-transformer-jax}}.
\newblock \URLprefix \url{https://github.com/kingoflolz/mesh-transformer-jax}.
\bibitem[{Wang et~al.(2021)Wang, Wang, Joty and Hoi}]{wang-etal-2021-codet5}
\bibinfo{author}{Wang, Y.}, \bibinfo{author}{Wang, W.}, \bibinfo{author}{Joty, S.}, \bibinfo{author}{Hoi, S.C.}, \bibinfo{year}{2021}.
\newblock \bibinfo{title}{{C}ode{T}5: Identifier-aware unified pre-trained encoder-decoder models for code understanding and generation}.
\newblock \bibinfo{journal}{Proceedings of the 2021 Conference on Empirical Methods in Natural Language Processing} .
\bibitem[{Wei et~al.(2022a)Wei, Tay, Bommasani, Raffel, Zoph, Borgeaud, Yogatama, Bosma, Zhou, Metzler and et~al.}]{wei2022_emergentAbilities}
\bibinfo{author}{Wei, J.}, \bibinfo{author}{Tay, Y.}, \bibinfo{author}{Bommasani, R.}, \bibinfo{author}{Raffel, C.}, \bibinfo{author}{Zoph, B.}, \bibinfo{author}{Borgeaud, S.}, \bibinfo{author}{Yogatama, D.}, \bibinfo{author}{Bosma, M.}, \bibinfo{author}{Zhou, D.}, \bibinfo{author}{Metzler, D.}, \bibinfo{author}{et~al.}, \bibinfo{year}{2022}a.
\newblock \bibinfo{title}{Emergent abilities of large language models}.
\newblock \href{http://arxiv.org/abs/2206.07682}{{\tt arXiv:2206.07682}}.
\bibitem[{Wei et~al.(2022b)Wei, Wang, Schuurmans, Bosma, hsin Chi, Xia, Le and Zhou}]{Wei2022_COT}
\bibinfo{author}{Wei, J.}, \bibinfo{author}{Wang, X.}, \bibinfo{author}{Schuurmans, D.}, \bibinfo{author}{Bosma, M.}, \bibinfo{author}{hsin Chi, E.H.}, \bibinfo{author}{Xia, F.}, \bibinfo{author}{Le, Q.}, \bibinfo{author}{Zhou, D.}, \bibinfo{year}{2022}b.
\newblock \bibinfo{title}{Chain of thought prompting elicits reasoning in large language models}.
\newblock \bibinfo{journal}{ArXiv} \bibinfo{volume}{abs/2201.11903}.
\newblock \URLprefix \url{https://api.semanticscholar.org/CorpusID:246411621}.
\bibitem[{Wei et~al.(2023)Wei, Wang, Liu, Ding and Zhang}]{wei2023magicoder}
\bibinfo{author}{Wei, Y.}, \bibinfo{author}{Wang, Z.}, \bibinfo{author}{Liu, J.}, \bibinfo{author}{Ding, Y.}, \bibinfo{author}{Zhang, L.}, \bibinfo{year}{2023}.
\newblock \bibinfo{title}{Magicoder: Source code is all you need}.
\newblock \bibinfo{journal}{arXiv preprint arXiv:2312.02120} .
\bibitem[{Wong et~al.(2023)Wong, Guo, Hang, Ho and Tan}]{wong2023natural}
\bibinfo{author}{Wong, M.F.}, \bibinfo{author}{Guo, S.}, \bibinfo{author}{Hang, C.N.}, \bibinfo{author}{Ho, S.W.}, \bibinfo{author}{Tan, C.W.}, \bibinfo{year}{2023}.
\newblock \bibinfo{title}{Natural language generation and understanding of big code for ai-assisted programming: A review}.
\newblock \bibinfo{journal}{Entropy} \bibinfo{volume}{25}, \bibinfo{pages}{888}.
\bibitem[{Wu et~al.(2023)Wu, Duan and Ni}]{Wu2023_security}
\bibinfo{author}{Wu, X.}, \bibinfo{author}{Duan, R.}, \bibinfo{author}{Ni, J.}, \bibinfo{year}{2023}.
\newblock \bibinfo{title}{Unveiling security, privacy, and ethical concerns of chatgpt}.
\newblock \bibinfo{journal}{Journal of Information and Intelligence} \URLprefix \url{https://www.sciencedirect.com/science/article/pii/S2949715923000707}, \DOIprefix\doi{https://doi.org/10.1016/j.jiixd.2023.10.007}.
\bibitem[{Xiao et~al.(2023)Xiao, Lin, Seznec, Wu, Demouth and Han}]{xiao2023smoothquant}
\bibinfo{author}{Xiao, G.}, \bibinfo{author}{Lin, J.}, \bibinfo{author}{Seznec, M.}, \bibinfo{author}{Wu, H.}, \bibinfo{author}{Demouth, J.}, \bibinfo{author}{Han, S.}, \bibinfo{year}{2023}.
\newblock \bibinfo{title}{Smoothquant: Accurate and efficient post-training quantization for large language models}, in: \bibinfo{booktitle}{International Conference on Machine Learning}, \bibinfo{organization}{PMLR}. pp. \bibinfo{pages}{38087--38099}.
\bibitem[{Yu et~al.(2023)Yu, He, Wu, Dai and Chen}]{yu2023_betterprompts}
\bibinfo{author}{Yu, Z.}, \bibinfo{author}{He, L.}, \bibinfo{author}{Wu, Z.}, \bibinfo{author}{Dai, X.}, \bibinfo{author}{Chen, J.}, \bibinfo{year}{2023}.
\newblock \bibinfo{title}{Towards better chain-of-thought prompting strategies: A survey}.
\newblock \href{http://arxiv.org/abs/2310.04959}{{\tt arXiv:2310.04959}}.
\bibitem[{Zhang and Sennrich(2019)}]{zhang2019_rms}
\bibinfo{author}{Zhang, B.}, \bibinfo{author}{Sennrich, R.}, \bibinfo{year}{2019}.
\newblock \bibinfo{title}{Root mean square layer normalization}.
\newblock \href{http://arxiv.org/abs/1910.07467}{{\tt arXiv:1910.07467}}.
\bibitem[{Zheng et~al.(2023)Zheng, Chiang, Sheng, Zhuang, Wu, Zhuang, Lin, Li, Li and et~al.}]{zheng2023_MTBench}
\bibinfo{author}{Zheng, L.}, \bibinfo{author}{Chiang, W.L.}, \bibinfo{author}{Sheng, Y.}, \bibinfo{author}{Zhuang, S.}, \bibinfo{author}{Wu, Z.}, \bibinfo{author}{Zhuang, Y.}, \bibinfo{author}{Lin, Z.}, \bibinfo{author}{Li, Z.}, \bibinfo{author}{Li, D.}, \bibinfo{author}{et~al., E.P.X.}, \bibinfo{year}{2023}.
\newblock \bibinfo{title}{Judging llm-as-a-judge with mt-bench and chatbot arena}.
\newblock \href{http://arxiv.org/abs/2306.05685}{{\tt arXiv:2306.05685}}.
\bibitem[{Zhong et~al.(2023)Zhong, Cui, Guo, Liang, Lu, Wang, Saied, Chen and Duan}]{zhong2023agieval}
\bibinfo{author}{Zhong, W.}, \bibinfo{author}{Cui, R.}, \bibinfo{author}{Guo, Y.}, \bibinfo{author}{Liang, Y.}, \bibinfo{author}{Lu, S.}, \bibinfo{author}{Wang, Y.}, \bibinfo{author}{Saied, A.}, \bibinfo{author}{Chen, W.}, \bibinfo{author}{Duan, N.}, \bibinfo{year}{2023}.
\newblock \bibinfo{title}{Agieval: A human-centric benchmark for evaluating foundation models}.
\newblock \href{http://arxiv.org/abs/2304.06364}{{\tt arXiv:2304.06364}}.
\bibitem[{Zhou et~al.(2023)Zhou, Muresanu, Han, Paster, Pitis, Chan and Ba}]{zhou2023_APE_framework}
\bibinfo{author}{Zhou, Y.}, \bibinfo{author}{Muresanu, A.I.}, \bibinfo{author}{Han, Z.}, \bibinfo{author}{Paster, K.}, \bibinfo{author}{Pitis, S.}, \bibinfo{author}{Chan, H.}, \bibinfo{author}{Ba, J.}, \bibinfo{year}{2023}.
\newblock \bibinfo{title}{Large language models are human-level prompt engineers}.
\newblock \bibinfo{journal}{The Eleventh International Conference on Learning Representations} \URLprefix \url{https://openreview.net/forum?id=92gvk82DE-}.
\bibitem[{Zhuang et~al.(2024)Zhuang, Yu, Wang, Sun and Zhang}]{zhuang2024toolqa}
\bibinfo{author}{Zhuang, Y.}, \bibinfo{author}{Yu, Y.}, \bibinfo{author}{Wang, K.}, \bibinfo{author}{Sun, H.}, \bibinfo{author}{Zhang, C.}, \bibinfo{year}{2024}.
\newblock \bibinfo{title}{Toolqa: A dataset for llm question answering with external tools}.
\newblock \bibinfo{journal}{Advances in Neural Information Processing Systems} \bibinfo{volume}{36}.

\end{thebibliography}






\newpage 

\appendix

\section{Prompt Engineering}
\label{app:prompt_engineering}

\begin{figure}
        \centering
        \includegraphics[width=\textwidth]{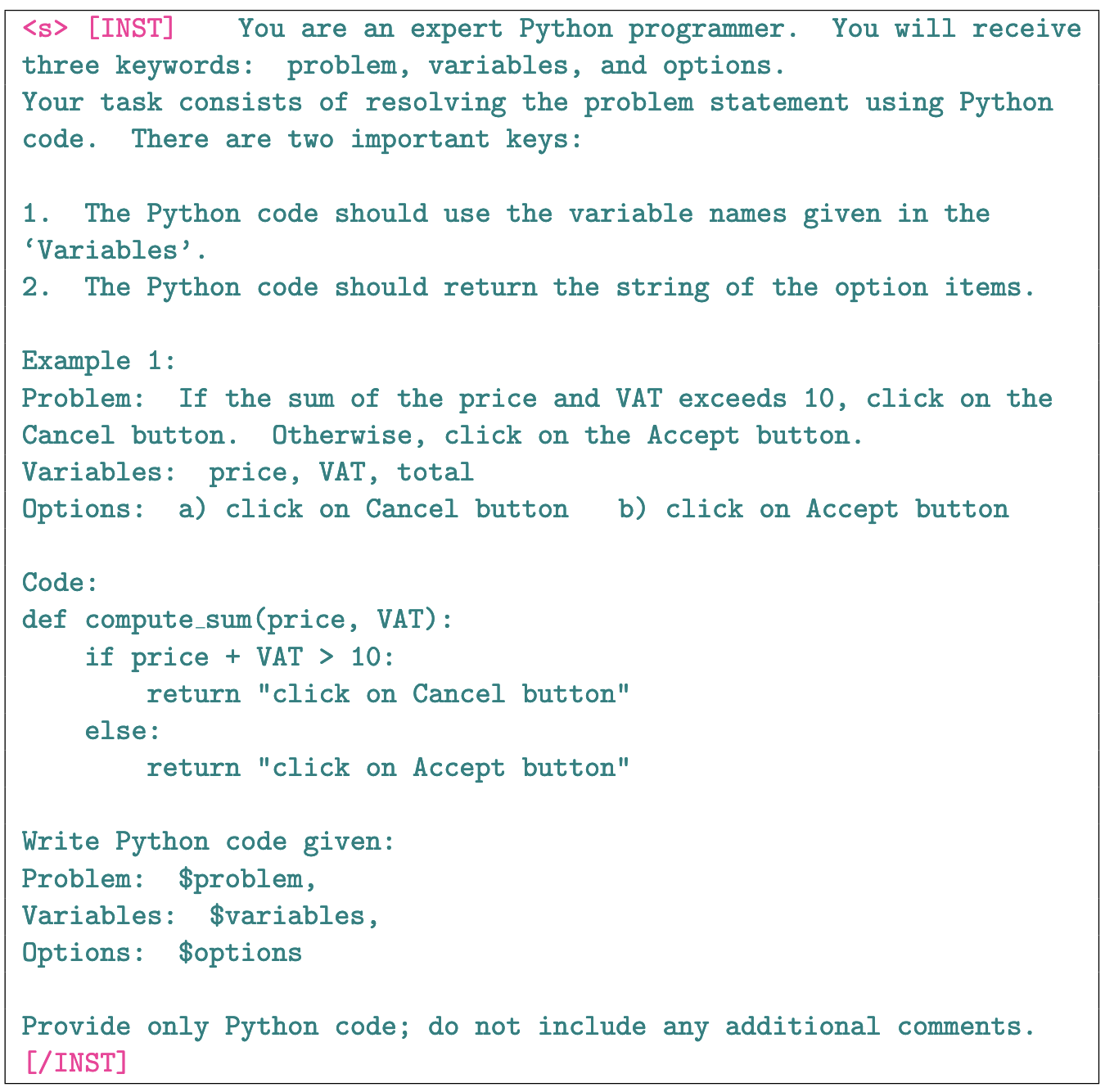}
        \caption{Prompt template for Mistral-7B-Instruct-v0.2 model. \textbf{Note:} Text highlighted in pink represents special tokens, while teal indicates the user's input.}
        \label{tab:prompt_mistral}
\end{figure}

    \textcolor{black}{As discussed in Section~\ref{sec:prompt_eng}, each evaluated model requires a customized template that aligns with its specific tokens for beginning and ending a turn, as well as its role management. For example, the Mistral-7B-Instruct-v0.2 model uses two roles, ``user'' and ``assistant''. In this model, the conversation starts with the user's input, followed by the assistant's response. Table~\ref{tab:prompt_mistral} shows an example of the final template used as input for Mistral-7B-Instruct-v0.2, where the prompt begins with the special tokens ``<s> [INST]'', with the user prompt text highlighted in teal, and ends with ``[/INST]''.}
    
\begin{figure}
        \centering
        \includegraphics[width=\textwidth]{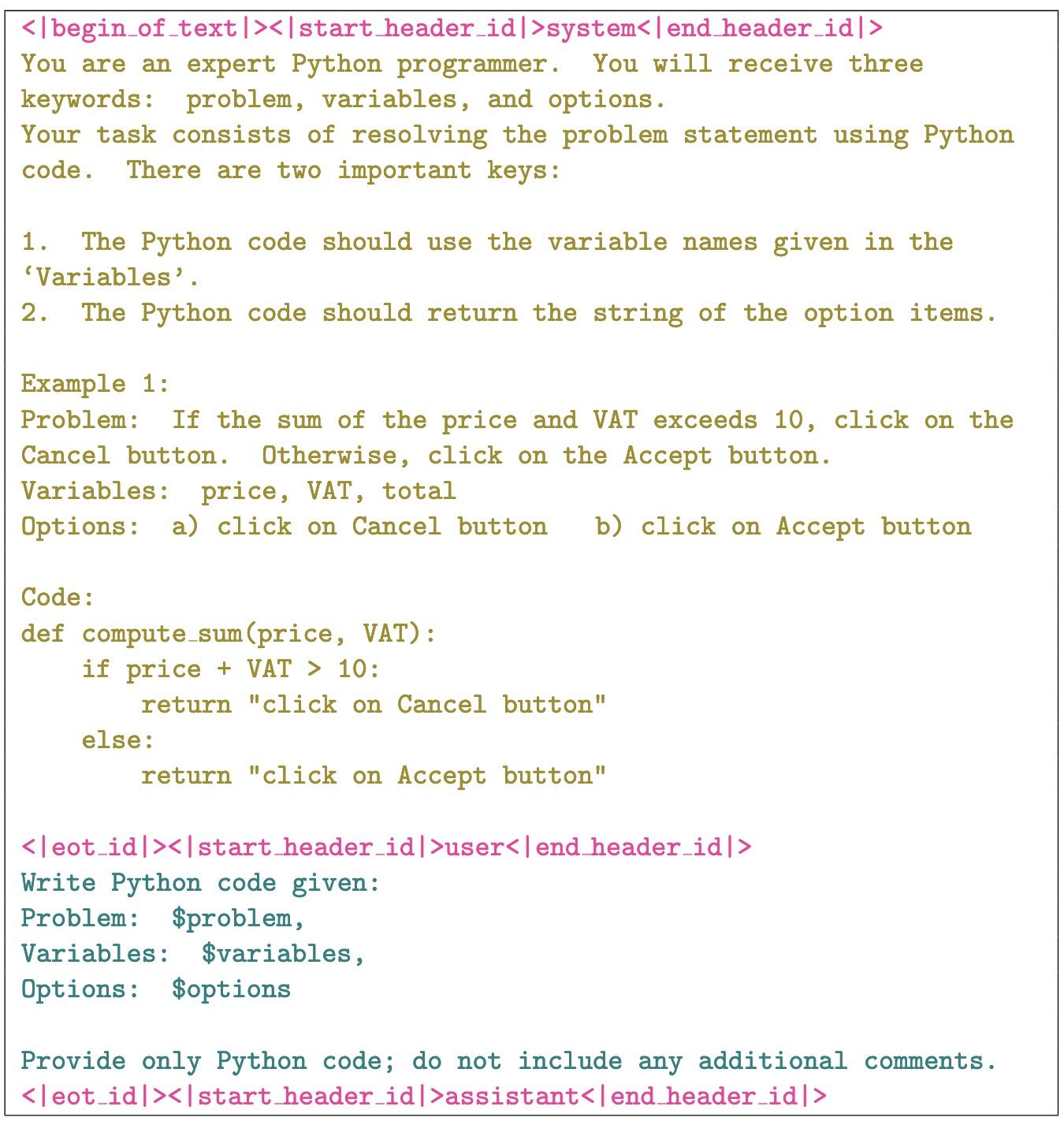}
        \caption{Prompt template for Meta-Llama-3.1-8B-Instruct model. \textbf{Note:} Text highlighted in pink represents special tokens, olive indicates the system's input, and teal represents the user's input.}
        \label{tab:prompt_llama3_1}
\end{figure}

\begin{figure}
        \centering
        \includegraphics[width=\textwidth]{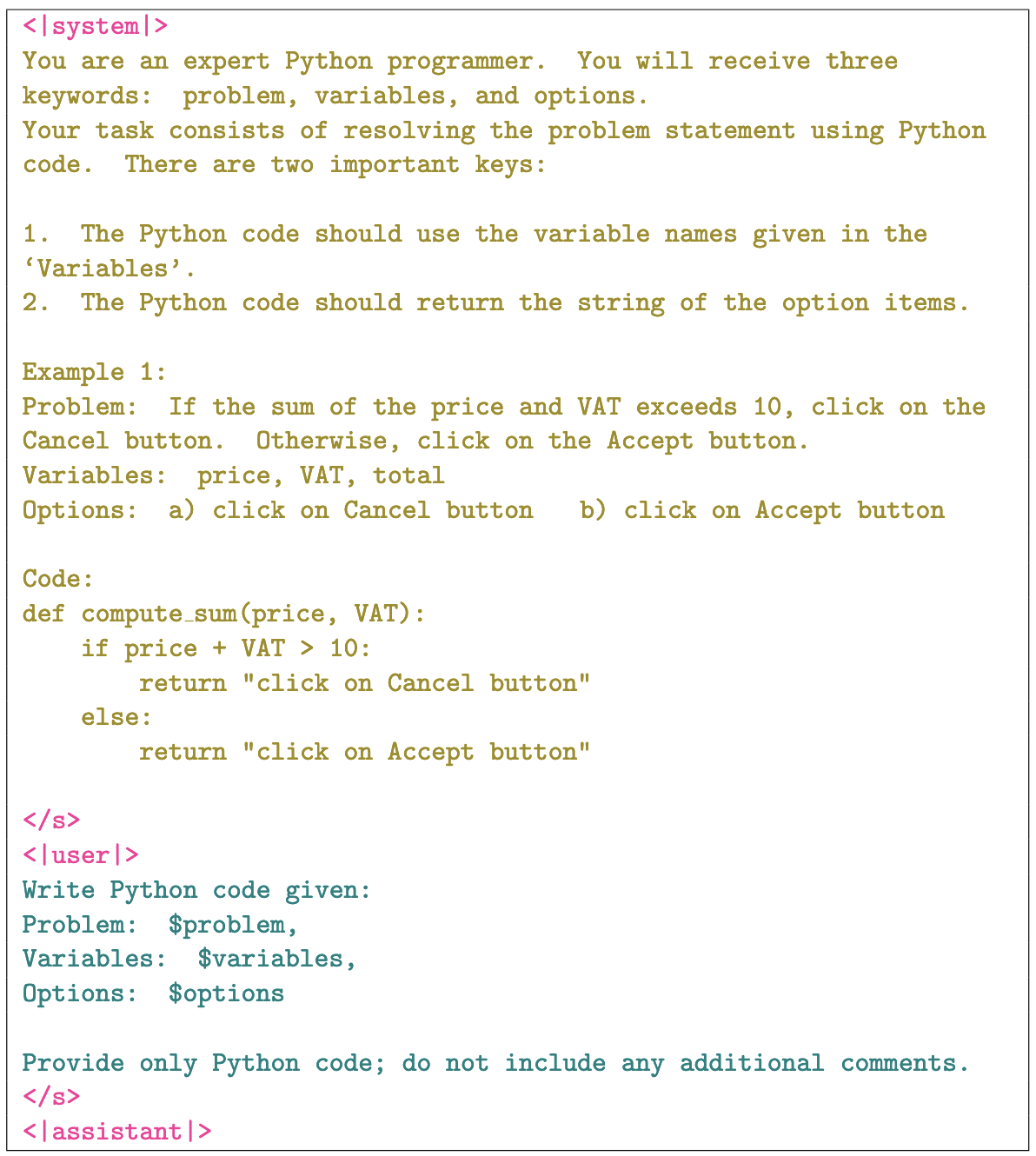}
        \caption{Prompt template for zephyr-7b-beta model. \textbf{Note:} Text highlighted in pink represents special tokens, olive indicates the system's input, and teal represents the user's input.}
        \label{tab:prompt_zephyr}
\end{figure}

\begin{figure}
        \centering
        \includegraphics[width=\textwidth]{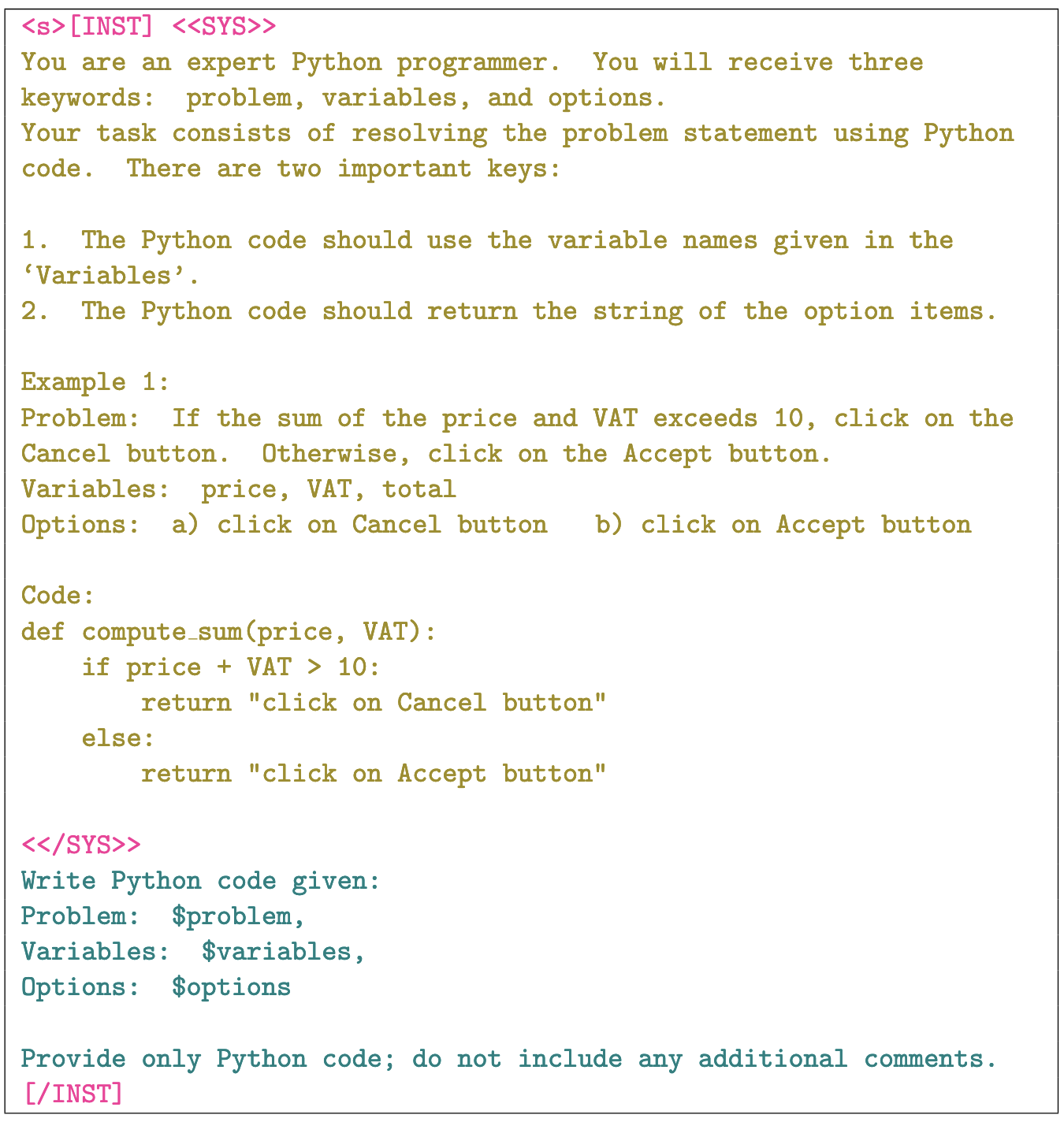}
        \caption{Prompt template for dolphin-2.6-mistral-7b model. \textbf{Note:} Text highlighted in pink represents special tokens, olive indicates the system's input, and teal represents the user's input.}
        \label{tab:prompt_dolphin}
\end{figure}

    \textcolor{black}{In contrast, models like Llama-3.1, Zephyr, and Dolphin use three roles: ``system'', ``user'' and ``assistant''. Unlike the template in Table Table~\ref{tab:prompt_mistral}, these models require the prompt to be split, with part of it (highlighted in olive) assigned to the system role, and the rest (in teal) used as the user prompt. Additionally, each of these models uses different special tokens to mark the beginning and end of roles, highlighted in pink. Refer to Table~\ref{tab:prompt_llama3_1}, Table~\ref{tab:prompt_dolphin} and Table~\ref{tab:prompt_zephyr}  for the complete prompts used with Llama-3.1, Zephyr, and Dolphin, respectively.}

\end{document}